\pdfoutput=1
\documentclass[11pt]{article}
\PassOptionsToPackage{table}{xcolor}
\usepackage{times}
\usepackage{caption}
\usepackage{booktabs}
\usepackage{graphicx}
\usepackage{multirow}
\usepackage{arydshln}
\usepackage{latexsym}
\usepackage{tcolorbox}
\usepackage{microtype}
\usepackage{subcaption}
\usepackage{inconsolata}
\usepackage[preprint]{acl}

\usepackage[T1]{fontenc}
\usepackage[utf8]{inputenc}
\usepackage{amsmath, amsfonts}
\title{TAD-Bench: A Comprehensive Benchmark for Embedding-Based Text Anomaly Detection}

\author{
 \textbf{Yang Cao\textsuperscript{1,2, 3}},
 \textbf{Sikun Yang\textsuperscript{1,2}},
 \textbf{Chen Li\textsuperscript{4}},
 \textbf{Haolong Xiang\textsuperscript{5}},
 \textbf{Lianyong Qi\textsuperscript{6}},
 \textbf{Bo Liu\textsuperscript{7}},
 \textbf{Rongsheng Li\textsuperscript{8}},
 \textbf{Ming Liu\textsuperscript{9}},
\\
 \textsuperscript{1}School of Computing and Information Technology, Great Bay University, China \\
 \textsuperscript{2}Great Bay Institute for Advanced Study, Great Bay University, China \\
 \textsuperscript{3}Tsinghua Shenzhen International Graduate School, Tsinghua University, China \\
 \textsuperscript{4}D3 Center, Osaka University, Japan \\
 \textsuperscript{5}School of Software, Nanjing University of Information Science and Technology, China \\
 \textsuperscript{6}College of Computer Science and Technology
China University of Petroleum (East China), China\\
 \textsuperscript{7}College of Cyberspace Security, Zhengzhou University, China \\
 \textsuperscript{8}School of Computer, Harbin Engineering University, China \\
  \textsuperscript{9}School of IT, Deakin University, Australia \\
\\
}

\begin{document}
\maketitle
\begin{abstract}
Text anomaly detection is crucial for identifying spam, misinformation, and offensive language in natural language processing tasks. Despite the growing adoption of embedding-based methods, their effectiveness and generalizability across diverse application scenarios remain insufficiently explored. To address this, we present TAD-Bench, a comprehensive benchmark designed to systematically evaluate embedding-based approaches for text anomaly detection. TAD-Bench integrates multiple datasets spanning different domains, combining state-of-the-art embeddings from large language models with a variety of anomaly detection algorithms. Through extensive experiments, we analyze the interplay between embeddings and detection methods, uncovering their strengths, weaknesses, and applicability to different tasks. These findings offer new perspectives on building more robust, efficient, and generalizable anomaly detection systems for real-world applications. All the code are available at \url{https://anonymous.4open.science/r/TAD-Bench-B4C6/}.
\end{abstract}

%=======================================================
\section{Introduction}
Anomaly detection (AD) is a critical task in machine learning, widely applied in fraud detection and content moderation to user behavior analysis~\cite{pang2021deep}. Within natural language processing (NLP), anomaly detection has become increasingly relevant for identifying outliers such as harmful content, phishing attempts, and spam reviews. However, while AD tasks in structured data (e.g., tabular, time series, graphs)~\cite{steinbuss2021benchmarking, blazquez2021review, qiao2024deep} have been extensively studied, anomaly detection in the unstructured and high-dimensional domain of text remains underexplored. The inherent complexity of textual data, driven by its diverse syntactic, semantic, and pragmatic structures, presents significant challenges for robust and reliable anomaly detection.

The rise of deep learning and transformer-based models has revolutionized NLP, enabling the development of contextualized embeddings that encode rich semantic and syntactic information. Techniques such as BERT \cite{devlin-etal-2019-bert} and OpenAI's text-embedding models \cite{openai2023new_embeddings} have demonstrated remarkable success across a wide range of NLP tasks, offering dense, high-dimensional representations that effectively capture linguistic nuances. These embeddings have become a cornerstone for many downstream tasks, providing powerful tools for applications such as text classification~\cite{da2023text} and retrieval~\cite{zhu2023large}. Their ability to generalize across tasks and domains positions them as a promising foundation for complex challenges, including anomaly detection.

\begin{figure}[!htbp]
    \centering
    \includegraphics[width=0.6\linewidth]{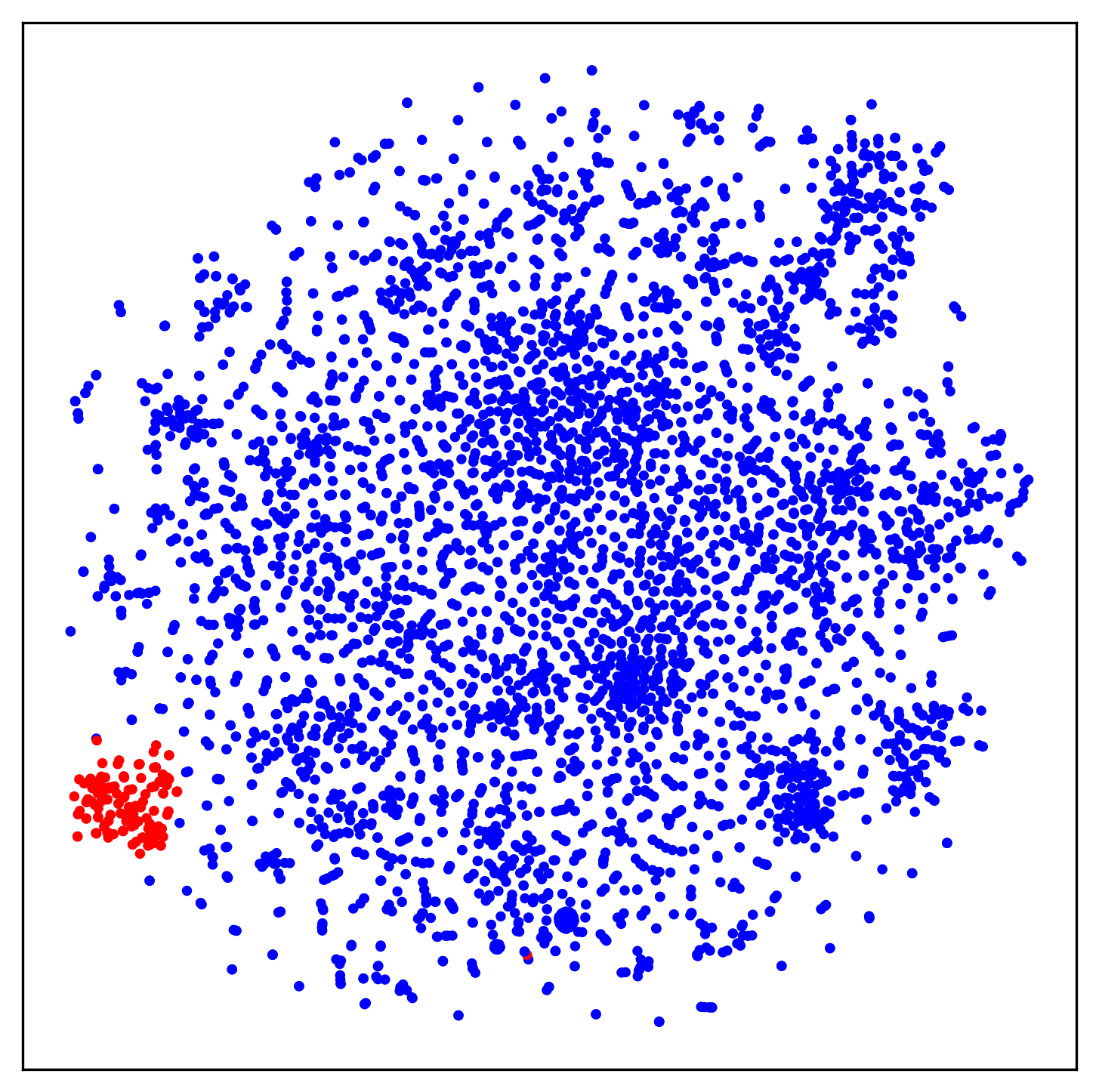}
    \caption{t-SNE visualization of SMS\_Spam dataset's embedding extracted by OpenAI model. Blue and red points are normal and anomaly points, respectively.}
    \label{fig:oe}
\end{figure}

\begin{figure*}[!]
\centering
\includegraphics[width=1\linewidth]{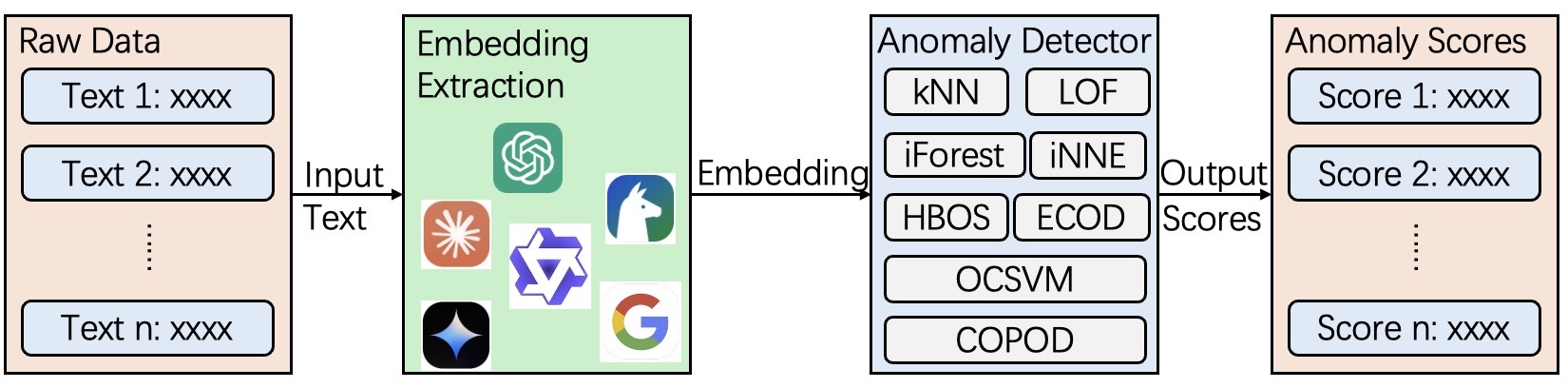}
\caption{Illustration of the embedding-based anomaly detection pipeline, encompassing embedding extraction and anomaly scoring.}
\label{fig:step}
\end{figure*}

In recent years, embedding-based methods have gained significant attention in anomaly detection tasks due to their ability to capture semantic and contextual nuances in data~\cite{wang2024log2graphs}. These methods typically involve two key stages: 1) extracting high-dimensional representations from textual data using pre-trained language models, which encode rich contextual and semantic features. Figure \ref{fig:oe} demonstrates the anomaly distribution of SMS\_Spam dataset embedding extracted from OpenAI model. 2) Applying specialized algorithms to identify anomalies based on these embeddings. The embeddings serve as a compact and expressive feature space, enabling downstream algorithms to efficiently identify deviations or outliers. Figure~\ref{fig:step} shows the steps involved in embedding-based anomaly detection.

Recent efforts, such as AD-NLP~\cite{bejan2023ad}, TAD~\cite{xu2023comparative} and NLP-ADBench~\cite{li2024nlp}, have significantly advanced anomaly detection in NLP. AD-NLP~\cite{bejan2023ad} finds that semantic and stylistic anomalies are easier to detect than those partially dependent on text; TAD~\cite{xu2023comparative} shows the effectiveness of embedding-based methods on multi language applications; NLP-ADBench~\cite{li2024nlp} reveals that no single model performs best across all datasets and high-dimensional embeddings improve detection. 
However, they only use a few embedding models, none of them explore the impact of embeddings quality to anomaly detection performance and tradeoffs between embedding models and anomaly detectors, raising questions about generalization capabilities of embedding-based methods in complex, real-world scenarios. 

Our work aims to move beyond simply filling these gaps, by systematically exploring the following questions: 1) What types of tasks are LLMs (Large Language Models) embeddings paired with anomaly detectors most suitable for, and where do they face limitations?
2) Which embedding methods consistently excel across different anomaly detection tasks?
3) Which anomaly detection algorithms perform robustly across various embeddings and tasks?

In this work, we introduce TAD-Bench, a novel benchmark specifically designed for text anomaly detection. Our objective is to enable a more comprehensive and systematic evaluation of state-of-the-art embeddings, anomaly detection techniques, and their various combinations, offering valuable insights for a broad spectrum of NLP applications. 
% By incorporating a diverse range of embedding models and rigorously evaluating an extensive suite of anomaly detection methods, TAD-Bench facilitates an in-depth understanding of their effectiveness on static datasets, with a strong emphasis on robustness, adaptability, and real-world applicability. 
The main contributions of this work are summarized as follows:
\begin{itemize}
\item Propose TAD-Bench, a benchmark integrating diverse datasets for text anomaly detection across domains such as spam, fake news, and offensive language.
\item Conduct a systematic evaluation of LLM-based embeddings and anomaly detection algorithms, revealing their relative strengths and weaknesses.
\item Provide insights into effective embedding-detector configurations for improving robustness and generalizability in NLP anomaly detection tasks.
\end{itemize}

The key insights of TAD-Bench have been summarized as follows: \textbf{1)} The effectiveness of embedding models varies significantly by task type: they can extract meaningful embeddings on tasks with explicit patterns (e.g. email spam with gibberish text) but struggle with context-dependent anomalies (e.g. offensive language). \textbf{2)} Among different detection algorithms applied to various embeddings, there are significant performance differences, but with default parameters, nearest-neighbor-based methods including kNN and INNE show better robustness. \textbf{3)} Clustering patterns in embedding spaces reveal that spam anomalies form distinct, compact clusters, whereas hate speech and offensive language anomalies are scattered among normal instances, explaining why detection performance varies across domains despite using the same embedding-detector combinations. \textbf{4)} On texts with explicit linguistic patterns like email spam, lightweight embedding models (MINILM) perform comparably to larger models (e.g. OpenAI, Llama), suggesting efficient model selection based on task characteristics.

\section{Related Work}
%=======
\subsection{Text representations}
% Advancements in text representation extraction techniques have been instrumental in driving significant progress in the field of natural language processing.

Early methods like TF-IDF (Term Frequency-Inverse Document Frequency)~\cite{salton1988term} represented text in sparse vector spaces by measuring word importance relative to a corpus. While interpretable and computationally efficient, TF-IDF could not capture semantic relationships between words. Later, dense embeddings such as Word2Vec (Word to Vector) \cite{mikolov2013efficient} and GloVe (Global Vectors for Word Representation)~\cite{pennington2014glove} addressed this limitation by mapping words into continuous vector spaces based on their co-occurrence patterns in large corpora. However, these embeddings were static, assigning the same vector to a word regardless of its context.

To overcome the limitations of static embeddings, contextualized embeddings were introduced, with models like ELMo (Embeddings from Language Models)~\cite{peters-etal-2018-deep} producing word representations that vary based on context. This innovation was further advanced by transformer-based models like BERT (Bidirectional Encoder Representations from Transformers) \cite{devlin-etal-2019-bert}, which used bidirectional attention mechanisms to simultaneously capture left and right context. BERT set new benchmarks in NLP and inspired numerous improvements, including RoBERTa (Robustly Optimized BERT Approach)~\cite{zhuang-etal-2021-robustly} and ALBERT (A Lite BERT)~\cite{lan2020albertlitebertselfsupervised}.

More recently, large language models such as GPT (Generative Pre-trained Transformer~\cite{brown2020language} have significantly advanced the capabilities of embedding methods. These models, trained on massive and diverse datasets, generate highly expressive embeddings that capture both deep semantic relationships and rich generative properties of text. LLMs have exhibited unprecedented performance across a broad spectrum of NLP tasks, solidifying their role as dominant tools for text representation in numerous applications, including anomaly detection, information retrieval, and text generation.

%======
\subsection{Anomaly Detection}
% Existing anomaly detection methods can be broadly categorized into 6 classes: distance, density, isolation, statistical, projection and deep learning-based approaches. Each category offers distinct advantages and is suited for different types of data distributions and anomaly patterns.

Distance-based methods, such as kNN (k-Nearest Neighbors)~\cite{ramaswamy2000efficient}, identify anomalies by measuring the distance of a given data point to its nearest neighbors. Points that are far from their neighbors are considered anomalous. These methods are intuitive and straightforward but suffer from the curse of dimensionality in high-dimensional spaces, where distances lose their discriminative power, reducing their effectiveness.

Density-based methods identify points with significantly lower local density compared to their surroundings as anomaly. LOF (Local Outlier Factor)~\cite{breunig2000lof} measures the local density of a point relative to its neighbors. HBOS (Histogram-Based Outlier Score)~\cite{goldstein2012histogram} estimates densities using histograms for individual features. 

Isolation-based methods assume anomalies are rare and different, iForest (Isolation Forest)~\cite{liu2008isolation, liu2012isolation}, detect anomalies by recursively partitioning the feature space where anomalies require fewer partitions than normal points. Improved techniques, such as iNNE (Isolation-based Nearest Neighbor Ensembles)~\cite{bandaragoda2018isolation}, use hypersphere to partition data space and assigns larger hyperspheres to anomalies, improving robustness in detecting local anomalies.

Probabilistic and statistical methods identify anomalies based on deviations from the data distribution. These approaches assume that normal instances follow a certain statistical pattern, and anomalies appear as outliers that do not conform to this pattern. ECOD (Empirical Cumulative Distribution Function-based Outlier Detection)~\cite{li2022ecod} uses cumulative distribution functions for efficient anomaly scoring, while COPOD (Copula-Based Outlier Detection)~\cite{li2020copod} leverages copulas to model feature dependencies, handling multivariate data effectively. Projection-based methods, such as OCSVM (One-Class SVM)~\cite{scholkopf2001estimating}, separate normal and anomalous data by learning a decision boundary in a high-dimensional feature space. While effective for complex distributions. 

Deep learning-based methods train on normal data to learn representations, identifying anomalies as deviations. Approaches like Deep SVDD (Deep Support Vector Data Description)~\cite{ruff2018deep} and LUNAR (Unifying Local Outlier Detection Methods via Graph Neural Networks)~\cite{goodge2022lunar} capture nonlinear patterns but require substantial data and computational resources.

%=======================================================
\section{Benchmark Settings}
%=========
\subsection{Datasets}
\begin{table}
    \caption{Dataset description. Nor. and Ano. stand for Normal and Anomaly.}
    \label{tab:dataset}
    \centering
    \resizebox{0.5\textwidth}{!}{
    \begin{tabular}{l|cccc}
    \hline
     Dataset    & \# Samples & \# Nor. & \# Ano. & \% Ano.  \\
     \hline
     Email Spam    & 3578            & 3432            & 146           &  4.0805            \\
     SMS Spam       & 4969            & 4825            & 144           &  2.8980            \\
     COVID-Fake     & 1173            & 1120            & 53            &  4.5183            \\
     LIAR2            & 2130            & 2068            & 62            &  2.9108            \\
     OLID     & 641             & 620             & 21            &  3.2761           \\
     Hate Speech     & 4287            & 4163            & 124           &  2.8925                  \\
    \hline
    \end{tabular}
    }
\end{table}

The scarcity of dedicated datasets poses a challenge to the development and evaluation of effective anomaly detection methods in NLP. To address this gap, we curated and transformed 6 existing classification datasets from three common NLP domains: spam detection, fake news detection, and offensive language detection. By incorporating datasets from diverse domains, our benchmark facilitates a comprehensive evaluation of embedding-based anomaly detection methods across various NLP tasks.

Anomalies, as defined in our problem, are inherently rare. However, due to the lack of dedicated datasets for text anomaly detection, we adapted classification datasets by designating specific classes as anomalies and down-sampling them to simulate realistic anomaly rates~\cite{li2024nlp}. For each dataset, the anomaly rate was set to approximately 3\%, reflecting the typical rarity of anomalies in real-world scenarios. 

% Unlike prior studies that structure anomaly detection as a novelty detection problem, where training data is assumed to contain only normal instances and anomalies appear only in the test set (e.g., NLP-ADBench~\cite{li2024nlp}). TAD-Bench does not impose this restriction, all available data is directly used for anomaly detection, without assuming prior knowledge of which instances are normal or anomalous.
% Unlike prior studies that frame anomaly detection as novelty detection, assuming only normal instances in training (e.g., NLP-ADBench~\cite{li2024nlp}), TAD-Bench imposes no such restriction and uses all data directly for anomaly detection.
% Unlike traditional text classification tasks, no additional pre-processing was applied to the source text, as any token, word, or symbol could potentially carry critical information indicative of an anomaly. By preserving the original text in its raw form, we aim to retain all inherent linguistic, structural, and contextual features that contribute to the  the detection of anomalous patterns. Table~\ref{tab:dataset} shows the statistics of the 6 pre-processed datasets used in this benchmark, including \textbf{Email-Spam}~\cite{metsis2006spam}, \textbf{SMS-Spam} \cite{almeida2011contributions}, \textbf{COVID-Fake} \cite{das2021heuristic}, \textbf{LIAR2} \cite{xu2024enhanced}, \textbf{OLID}~\cite{zampierietal2019} and \textbf{Hate-Speech} \cite{davidson2017automated}.

While some studies treat anomaly detection as novelty detection—assuming only normal instances in training (e.g., NLP-ADBench~\cite{li2024nlp}). TAD-Bench removes this constraint and directly utilizes all available data for anomaly detection. Additionally, we retain the original text without extra pre-processing, as any token, word, or symbol may carry critical information indicative of an anomaly. This approach preserves linguistic, structural, and contextual features essential for detecting anomalies. Table~\ref{tab:dataset} presents the statistics of the six pre-processed datasets used in this benchmark, including Email-Spam\cite{metsis2006spam}, SMS-Spam\cite{almeida2011contributions}, COVID-Fake\cite{das2021heuristic}, LIAR2\cite{xu2024enhanced}, OLID\cite{zampierietal2019}, and Hate-Speech\cite{davidson2017automated}.

%========
\subsection{Embedding Models}
To extract high-quality embeddings from the datasets, 8 embedding models were utilized. These include \textbf{BERT} (\emph{bert-base-uncased})~\cite{devlin-etal-2019-bert}, \textbf{MiniLM} (\emph{all-MiniLM-L6-v2})~\cite{wang2020minilm}, \textbf{LLAMA} (\emph{Llama-3.2-1B}), \textbf{stella} (\emph{stella\_en\_400M\_v5})~\cite{zhang2024jasper}, and \textbf{Qwen} (\emph{Qwen2.5-1.5B})~\cite{qwen2,qwen2.5} from the HuggingFace platform, as well as OpenAI-provided models: \textbf{O-ada} (\emph{text-embedding-ada-002}), \textbf{O-small} (\emph{text-embedding-3-small}), and \textbf{O-large} (\emph{text-embedding-3-large})~\cite{openai2023new_embeddings}. All these models are based on the Transformer architecture, which has become the standard for representation learning in NLP tasks. The OpenAI models (O-ada, O-small, O-large) are specifically designed for embedding generation, offering embeddings with varying levels of granularity. On the other hand, LLAMA and Qwen are primarily auto-regressive language models optimized for text generation. In this paper, we repurposed these models for embedding extraction by computing the attention-weighted mean of their last hidden states, ensuring that only valid tokens contribute to the final sentence embeddings.

Notably, LLAMA and Qwen were constrained to a maximum token length of 512 tokens, same as BERT, due to computational resource limitations. Other models, such as MiniLM, Stella, and the OpenAI embeddings, utilized automatic truncation to process longer input sequences. This limitation may restrict LLAMA and Qwen's ability to fully leverage their extended context capabilities, particularly for datasets with longer text instances, such as LIAR2 and Hate-Speech. However, this unified token length ensures a fair comparison of runtime efficiency across models under consistent experimental conditions. It also highlights the trade-offs between computational cost and embedding quality, particularly when resource constraints are a factor in model deployment.

\subsection{Anomaly Detectors}
The embeddings derived from these models were subsequently used as input features for anomaly detection algorithms. To identify anomalous instances, we employed 8 anomaly detection methods sourced from the \textit{PyOD} library\footnote{PyOD: \url{https://pyod.readthedocs.io/en/latest/index.html}} \cite{zhao2019pyod}. These algorithms include \textbf{KNN}, \textbf{LOF}, \textbf{OCSVM}, \textbf{iForest}, \textbf{INNE}, \textbf{ECOD}, \textbf{HBOS} and \textbf{COPOD} . These algorithms were selected to capture diverse anomaly detection paradigms, ensuring robust detection across datasets with varying characteristics, structures, and distributions.

For reproducibility and consistency, we implemented all algorithms using their default hyperparameter settings as specified in their original implementations and research papers. This approach minimizes subjective bias and enables fair comparison across different embedding models. Additionally, we conducted comprehensive grid search for optimal hyperparameter configurations, with search ranges detailed in Table~\ref{tab:hyper}. By evaluating both default and optimized settings across our diverse collection of embedding models and detection algorithms, we provide a thorough assessment of text anomaly detection performance in terms of both computational efficiency and detection effectiveness.

\subsection{Evaluation Criteria and Trials}
Performance was evaluated using the Area Under the Receiver Operating Characteristic Curve (AUROC), a widely adopted metric in anomaly detection tasks for measuring the trade-off between true positive and false positive rates. To ensure the reliability and robustness of the results, each experiment was repeated 5 times, and the average AUROC score was reported.

\section{Experiments}

\subsection{Domain Generalization}
Table \ref{result} summarizes the performance of various anomaly detectors combined with LLM-derived embeddings across different datasets, while Figure 1 highlights their strong performance in specific tasks, particularly spam detection. In both the email spam and SMS spam detection tasks (Figure~\ref{es} and Figure~\ref{ss}, many embedding-detector combinations achieve high AUC scores, with several exceeding 0.8. This strong performance can be attributed to the explicit nature of spam-related features, such as the presence of URLs, nonsensical text, or repetitive patterns. An example Case 1 from the Email Spam dataset is shown below:
\begin{tcolorbox}[colback=gray!20, colframe=black, boxrule=0.1mm, sharp corners, width=\linewidth]
\textbf{Case 1:} \emph{Subject: oxyccontttin no script needeeed your place to ggo too for all ur prreexxxxiscrlpt 10 n pi sx , paaaaain killerzxss noeoo presscippt http : / / hyyydroccodeeeine vicccodinne / vic geeet reeeliefff noowee http : / / offfmeebabyy}
\end{tcolorbox}

\begin{figure}[!htbp]
    \centering
    \includegraphics[width=0.5\linewidth]{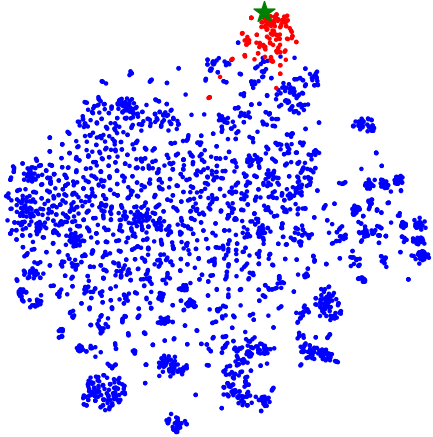}
    \caption{t-SNE demonstration of Case 1 (green star) embedding extracted by O-large.}
    \label{es1}
\end{figure}

Figure \ref{es1} shows that case 1 is located at the edge of the embedding space and deviates from the main data distribution, thus making it easy to be detected.

These features are effectively represented in the semantic spaces created by general-purpose embeddings, enabling anomaly detectors to distinguish spam messages from legitimate ones. Additionally, the relatively small variance in detection performance across embeddings suggests that spam detection primarily relies on surface-level linguistic patterns, which are effectively captured by the embeddings employed in this study.

For fake news detection, the results indicate a more mixed performance across datasets. On the Covid Fake News dataset (Figure~\ref{cfn}, multiple embedding-detector combinations achieve AUC scores close to or exceeding 0.8, suggesting that these methods are capable of identifying subtle stylistic and linguistic differences between fake and real news. These differences may include deviations in tone, phrasing, or structural composition of the text. However, on the LIAR2 dataset (Figure~\ref{liar2}, the AUC scores exhibit much greater variability across different combinations of embeddings and detectors. This variability likely stems from the greater factual complexity of the LIAR2 dataset, where detecting anomalies may require external knowledge or sophisticated reasoning that is not inherently encoded within the embeddings. Despite this variability, the relatively strong performance on the Covid Fake News dataset underscores the potential of embedding-based approaches for fake news detection, particularly when the anomalies are stylistic or linguistic in nature.

In contrast, the performance on hate speech and offensive language detection tasks (Figure~\ref{hs} and Figure~\ref{olid}) is consistently weaker, with AUC scores rarely exceeding 0.6 across embedding-detector combinations. This suggests that the embeddings struggle to capture the nuanced and context-dependent features necessary for these tasks. For instance, hate speech often relies on implicit cues such as sarcasm, cultural references, or subtle forms of hostility, which may not be fully captured by standard embeddings. Similarly, offensive language detection, as observed in the OLID dataset, requires identifying fine-grained differences in tone, intent, and subjectivity, such as distinguishing between neutral, offensive, and sarcastic expressions. These distinctions often depend on broader contextual information, such as the discourse or dialogue in which the language appears.

For example, without additional context, such as the speaker’s intent or the conversational background, the following statement from OLID dataset remains ambiguous whether this statement qualifies as hate speech:
\begin{tcolorbox}[colback=gray!10, colframe=black, boxrule=0.1mm, sharp corners, width=\linewidth]
\textbf{Case 2:} \emph{@USER \#metoo are all racist!}
\end{tcolorbox}

\begin{figure}[!htbp]
    \centering
    \includegraphics[width=0.6\linewidth]{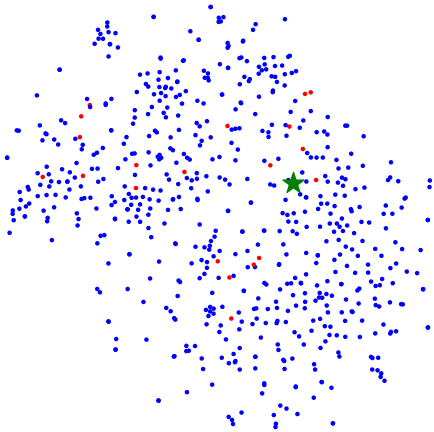}
    \caption{t-SNE demonstration of Case 2 (green star) embedding extracted by O-large.}
    \label{olid1}
\end{figure}

Figure~\ref{olid1} shows that case 2 is mixed in the normal data distribution, making it difficult to be detected.

\begin{table*}[htbp]
\caption{Evaluation across 6 datasets in terms of AUROC.}
\label{evaluation}
\resizebox{1\textwidth}{0.5\textheight}{
\begin{tabular}{ll|cccccc|c}
\hline
\textbf{Embeddings}                        & \textbf{Detectors} & \textbf{Email-Spam} & \textbf{SMS-Spam}         & \textbf{COVID-Fake} & \textbf{LIAR2}  & \textbf{Hate-Speech} & \textbf{OLID}   & \textbf{Average}\\ \hline
\multirow{7}{*}{\textbf{BERT}}             & \textbf{kNN}       & 0.7625              & 0.4484                    & 0.8467              & 0.6594          & \textbf{0.5033}      & 0.5137          & 0.6223 \\
                                           & \textbf{OCSVM}     & 0.7362              & 0.6323                    & 0.7867              & 0.6237          & 0.4866               & 0.4866          & 0.6254\\
                                           & \textbf{IForest}   & 0.7152              & 0.6164                    & 0.7701              & 0.6051          & 0.4925               & 0.4783          & 0.6129\\
                                           & \textbf{LOF}       & 0.6786              & 0.3230                    & \textbf{0.8713}     & \textbf{0.6717} & 0.4632               & 0.4970          & 0.5841\\
                                           & \textbf{ECOD}      & 0.7309              & 0.6235                    & 0.7722              & 0.6175          & 0.4889               & 0.4933          & 0.6211\\
                                           & \textbf{INNE}      & \textbf{0.7732}     & \textbf{0.6497}           & 0.8012              & 0.6362          & 0.4850               & 0.4740          & \textbf{0.6366}\\
                                           & \textbf{HBOS}      & 0.7145              & 0.6251                    & 0.7698              & 0.6190          & 0.4935               & 0.5002          & 0.5317\\
                                           & \textbf{COPOD}     & 0.6454              & 0.5929                    & 0.7714              & 0.6242          & 0.4971               & \textbf{0.5189} & 0.5214\\
                                           % & \textbf{LUNAR}     & 0.7485              & 0.4385                    & 0.7768              & 0.6575          & 0.5014               & 0.5208          & 0.6073\\
                                           % & \textbf{DeepSVDD}  & 0.5858              & 0.4911                    & 0.7229              & 0.5850          & 0.4885               & \textbf{0.5352} & 0.5681\\  
                                           \hline
% \hdashline                                 & \textbf{E-Time}    & 64.95 s             & 10.87 s                   & 4.48 s              & 3.7 s           & 9.79 s               & 1.81 s          \\ \hline
\multirow{7}{*}{\textbf{MINILM}}           & \textbf{kNN}       & 0.9414              & 0.3180                    & \textbf{0.8413}     & \textbf{0.7249} & \textbf{0.5804}      & 0.5063          & \textbf{0.6520}    \\
                                           & \textbf{OCSVM}     & \textbf{0.9626}     & 0.5915        & 0.7843              & 0.6470          & 0.4062               & 0.4520          & 0.6406\\
                                           & \textbf{IForest}   & 0.9078              & 0.5472                    & 0.7455              & 0.5936          & 0.4697               & 0.4531          & 0.6195\\
                                           & \textbf{LOF}       & 0.5587              & 0.5024                    & 0.7433              & 0.6804          & 0.5078               & \textbf{0.5422} & 0.5891\\
                                           & \textbf{ECOD}      & 0.9525              & 0.5934                    & 0.7581              & 0.6532          & 0.3786               & 0.4208          & 0.6261\\
                                           & \textbf{INNE}      & 0.9526              & 0.5737                    & 0.8035              & 0.6601          & 0.4223               & 0.4824          & 0.6491\\
                                           & \textbf{HBOS}      & 0.9478              & 0.6137                    & 0.7441              & 0.6447          & 0.3888               & 0.4316          & 0.5387\\
                                           & \textbf{COPOD}     & 0.9453              & \textbf{0.6317}           & 0.7416              & 0.6695          & 0.3710               & 0.4037          & 0.5375\\
                                           % & \textbf{LUNAR}     & 0.7310              & 0.3298                    & 0.7785              & 0.5895          & 0.5723               & 0.4877          & 0.5815\\
                                           % & \textbf{DeepSVDD}  & 0.6545              & 0.4807                    & 0.6054              & 0.5470          & 0.4660               & 0.5002          & 0.5423 \\   
                                           \hline
% \hdashline                                 & \textbf{E-Time}    & 3.58 s              & 1.48 s                    & 0.64 s              & 0.52 s          & 1.50 s               & 0.45 s          \\ \hline
\multirow{7}{*}{\textbf{O-ada}}       & \textbf{kNN}       & 0.8865              & 0.3212                    & \textbf{0.9094}     & \cellcolor[gray]{0.9}\textbf{0.7921} & \textbf{0.6341} & \textbf{0.5243} & 0.6779\\
                                           & \textbf{OCSVM}     & 0.9310              & 0.8221                    & 0.8143              & 0.7169          & 0.4807               & 0.5048          & 0.7116\\
                                           & \textbf{IForest}   & 0.8872              & 0.7376                    & 0.7432              & 0.6421          & 0.4632               & 0.4891          & 0.6604\\
                                           & \textbf{LOF}       & 0.3808              & 0.5033                    & 0.7316              & 0.7541          & 0.4328               & 0.5376          & 0.5567\\
                                           & \textbf{ECOD}      & 0.9380              & \cellcolor[gray]{0.9}\textbf{0.8822} & 0.8150              & 0.7200          & 0.4610             & 0.4986          & \cellcolor[gray]{0.9}{\textbf{0.7191}} \\
                                           & \textbf{INNE}      & 0.8507              & 0.8031                    & 0.8533              & 0.7378          & 0.4820               & 0.5102          & 0.7062\\
                                           & \textbf{HBOS}      & 0.9433              & 0.8813                    & 0.8164              & 0.7186          & 0.4583               & 0.5098          & 0.6182\\
                                           & \textbf{COPOD}     & \textbf{0.9502}     & 0.8759                    & 0.8153              & 0.7201          & 0.4513               & 0.4811          & 0.6134\\
                                           % & \textbf{LUNAR}     & 0.8671              & 0.3295                    & 0.8944              & 0.7255          & \underline{\textbf{0.6366}} & 0.5298   & 0.6638       \\
                                           % & \textbf{DeepSVDD}  & 0.6347              & 0.5397                    & 0.6041              & 0.4820          & 0.4454               & 0.5289          & 0.5391\\  
                                           \hline
% \hdashline                                 & \textbf{E-Time}    & 154.08 s            & 33.76 s                   & 17.35 s             & 16.67 s         & 35.57 s              & 8.9 s           \\ \hline
\multirow{7}{*}{\textbf{O-small}}     & \textbf{kNN}       & 0.8921              & 0.2290                    & \textbf{0.9400}     & \textbf{0.7756} & \cellcolor[gray]{0.9} \textbf{0.6416}      & \cellcolor[gray]{0.9} \textbf{0.5587} & 0.6728\\
                                           & \textbf{OCSVM}     & 0.9475              & 0.5755                    & 0.8932              & 0.7024          & 0.4577               & 0.5547          & \textbf{0.6885}\\
                                           & \textbf{IForest}   & 0.9058              & 0.6177                    & 0.8085              & 0.5973          & 0.5025               & 0.5580          & 0.6650\\
                                           & \textbf{LOF}       & 0.3863              & 0.5257                    & 0.7809              & 0.7489          & 0.4139               & 0.5581          & 0.5690\\
                                           & \textbf{ECOD}      & 0.9481              & \textbf{0.6301}           & 0.8808              & 0.7022          & 0.4249               & 0.5295          & 0.6859\\
                                           & \textbf{INNE}      & 0.8673              & 0.6080                    & 0.9185              & 0.7198          & 0.4491               & 0.5382          & 0.6835\\
                                           & \textbf{HBOS}      & 0.9522              & 0.6273                    & 0.8719              & 0.7008          & 0.4245               & 0.5157          & 0.5846\\
                                           & \textbf{COPOD}     & \textbf{0.9605}              & 0.5722                    & 0.8664              & 0.6974          & 0.4017               & 0.4963          & 0.5706\\
                                           % & \textbf{LUNAR}     & 0.8657              & 0.2536                    & 0.9201              & 0.7622          & \textbf{0.6454}      & 0.5602          & 0.6679\\
                                           % & \textbf{DeepSVDD}  & 0.6828              & 0.5499                    & 0.6596              & 0.5617          & 0.4931               & 0.4546          & 0.5670 \\  
                                           \hline
% \hdashline                                 & \textbf{E-Time}    & 166.73 s            & 34.16                     & 18.32               & 16.10           & 34.09 s              & 9.0 s           \\ \hline
\multirow{7}{*}{\textbf{O-large}} & \textbf{kNN}       & 0.8292              & 0.1698                    & \cellcolor[gray]{0.9}\textbf{0.9537}     & \textbf{0.7687} & \textbf{0.6291} & \textbf{0.5497} & 0.6500\\
                                           & \textbf{OCSVM}     & 0.9403              & 0.5630                    & 0.8924              & 0.6621          & 0.4260               & 0.4971          & 0.6635\\
                                           & \textbf{IForest}   & 0.8999              & 0.5297                    & 0.8041              & 0.5687          & 0.4516               & 0.5068          & 0.6268\\
                                           & \textbf{LOF}       & 0.4048              & 0.4719                    & 0.8233              & 0.7356          & 0.3833               & 0.5167          & 0.5559\\
                                           & \textbf{ECOD}      & 0.9487              & 0.6422                    & 0.8875              & 0.6540          & 0.3959               & 0.4967          & \textbf{0.6708}\\
                                           & \textbf{INNE}      & 0.8230              & 0.5970                    & 0.9261              & 0.6876          & 0.4197               & 0.5170          & 0.6617 \\
                                           & \textbf{HBOS}      & 0.9538              & 0.6525                    & 0.8849              & 0.6404          & 0.3835               & 0.4989          & 0.5734\\
                                           & \textbf{COPOD}     & \cellcolor[gray]{0.9}\textbf{0.9639}     & \textbf{0.6798}           & 0.8854              &0.6536          & 0.3537              & 0.4980          & 0.5763\\
                                           % & \textbf{LUNAR}     & 0.7670              & 0.2007                    & 0.9287              & 0.7511          & 0.6274               & 0.5242          & 0.6332\\
                                           % & \textbf{DeepSVDD}  & 0.6638              & 0.5742                    & 0.6220              & 0.4740          & 0.4740               & \textbf{0.5414} & 0.5582\\  
                                           \hline
% \hdashline                                 & \textbf{E-Time}    & 206.07 s            & 41.78 s                   & 20.58 s             & 29.97 s         & 44.20 s              & 10.72 s         \\ \hline
\multirow{7}{*}{{\textbf{Llama}}}          & \textbf{kNN}       & 0.8715              & 0.3655                    & \textbf{0.8668}     & 0.7229          & \textbf{0.4991}      & 0.4081            & 0.6223\\
                                           & \textbf{OCSVM}     & 0.9023              & 0.7379                    & 0.8132              & 0.6892          & 0.4774               & 0.4057          & \textbf{0.6710}\\
                                           & \textbf{IForest}   & 0.8962              & 0.7275                    & 0.7833              & 0.6860          & 0.4647               & \textbf{0.4082} & 0.6610\\
                                           & \textbf{LOF}       & 0.6056              & 0.4053                    & 0.8673              & 0.7274          & 0.4376               & 0.3972          & 0.5734\\
                                           & \textbf{ECOD}      & 0.8844              & 0.7573                    & 0.7819              & 0.6989          & 0.4643               & 0.3998          & 0.6644\\
                                           & \textbf{INNE}      & 0.9122              & 0.7065                    & 0.8160              & 0.6935          & 0.4702               & 0.3917          & 0.6650\\
                                           & \textbf{HBOS}      & 0.9017              & 0.7895                    & 0.7758              & 0.7064          & 0.4580               & 0.3898          & 0.5745\\
                                           & \textbf{COPOD}     & \textbf{0.9153}     & \textbf{0.8163}           & 0.7584              & \textbf{0.7291} & 0.4435               & 0.3526          & 0.5736\\
                                           % & \textbf{LUNAR}     & 0.8248              & 0.3756                    & 0.8631              & 0.6799          & \textbf{0.5072}      & 0.4117          & 0.6104\\
                                           % & \textbf{DeepSVDD}  & 0.8007              & 0.6792                    & 0.7100              & 0.6275          & 0.4733               & \textbf{0.4685} & 0.6265 \\   
                                           \hline
% \hdashline                                 & \textbf{E-Time}    & 545.51 s            & 129.16 s                  & 38.15 s             & 28.93 s         & 71.49 s              & 18.02 s         \\ \hline
\multirow{7}{*}{\textbf{stella}}       & \textbf{kNN}       & 0.8654              & 0.3212                    & \textbf{0.9034}     & \textbf{0.6884}     & \textbf{0.4746}      & 0.5016          & 0.6258\\
                                           & \textbf{OCSVM}     & 0.8922              & 0.7165                    & 0.8063              & 0.5103          & 0.3729               & 0.4439          & 0.6237\\
                                           & \textbf{IForest}   & 0.8862              & 0.7377                    & 0.7738              & 0.4999          & 0.3545               & 0.4325          & 0.6141\\
                                           & \textbf{LOF}       & 0.3931              & 0.4733                    & 0.7129              & 0.6549          & 0.4036               & \textbf{0.5285} & 0.5277\\
                                           & \textbf{ECOD}      & 0.9075              & 0.7894                    & 0.8115              & 0.5023          & 0.3421               & 0.4395          & \textbf{0.6321}\\
                                           & \textbf{INNE}      & 0.8271              & 0.6926                    & 0.8366              & 0.5330          & 0.3325               & 0.4532          & 0.6125\\
                                           & \textbf{HBOS}      & 0.9178              & 0.8017                    & 0.8086              & 0.4952          & 0.3355               & 0.4252          & 0.5406\\
                                           & \textbf{COPOD}     & \textbf{0.9300}     & \textbf{0.8589}           & 0.8167              & 0.4936          & 0.3018               & 0.3797          & 0.5401\\
                                           % & \textbf{LUNAR}     & 0.8511              & 0.3293                    & 0.8930              & 0.6451          & \textbf{0.4905}      & 0.5114          & 0.6201\\
                                           % & \textbf{DeepSVDD}  & 0.7437              & 0.5521                    & 0.6290              & 0.5256          & 0.4349               & \textbf{0.5295} & 0.5691\\   
                                           \hline
% \hdashline                                 & \textbf{E-Time}    & 99.75 s             & 19.38 s                   & 12.41 s             & 9.95 s          & 20.18 s              & 5.72 s          \\ \hline
\multirow{7}{*}{\textbf{Qwen}}             & \textbf{kNN}       & 0.8618              & 0.2110                    & \textbf{0.8438}     & 0.6626          & \textbf{0.5163}      & 0.4602              & 0.5926\\
                                           & \textbf{OCSVM}     & 0.8804              & 0.6229                    & 0.7868              & 0.6216          & 0.4916               & \textbf{0.4882} & 0.6486\\
                                           & \textbf{IForest}   & 0.8829              & 0.6195                    & 0.7686              & 0.6155          & 0.4825               & 0.4869          & \textbf{0.6427}\\
                                           & \textbf{LOF}       & 0.6043              & 0.3600                    & 0.8555              & \textbf{0.6894} & 0.4600               & 0.4518          & 0.5702\\
                                           & \textbf{ECOD}      & 0.8678              & 0.6648                    & 0.7680              & 0.6172          & 0.4852               & 0.4773          & 0.6467\\
                                           & \textbf{INNE}      & 0.8839              & 0.5940                    & 0.7833              & 0.6339          & 0.4902               & 0.4693          & 0.6424\\
                                           & \textbf{HBOS}      & 0.8854              & 0.6877                    & 0.7638              & 0.6170          & 0.4847               & 0.4685          & 0.5582\\
                                           & \textbf{COPOD}     & \textbf{0.9044}     & \textbf{0.7393}           & 0.7463              & 0.6291          & 0.4794               & 0.4336          & 0.5617\\
                                           % & \textbf{LUNAR}     & 0.8323              & 0.2388                    & 0.8032              & 0.6643          & 0.5152               & 0.4691          & 0.5872\\
                                           % & \textbf{DeepSVDD}  & 0.8139              & 0.4895                    & 0.6657              & 0.5854          & 0.4778               & 0.4504          & 0.5805\\   
                                           \hline
% \hdashline                                 & \textbf{E-Time}    & 745.85 s            & 129.16 s                  & 58.19 s             & 40.29 s         & 183.24 s             & 20.49 s         \\ \hline
\end{tabular}}
\end{table*}

\begin{figure*}[!htbp]
     \centering
     \begin{subfigure}[b]{0.3\textwidth}
         \centering
         \includegraphics[width=\textwidth]{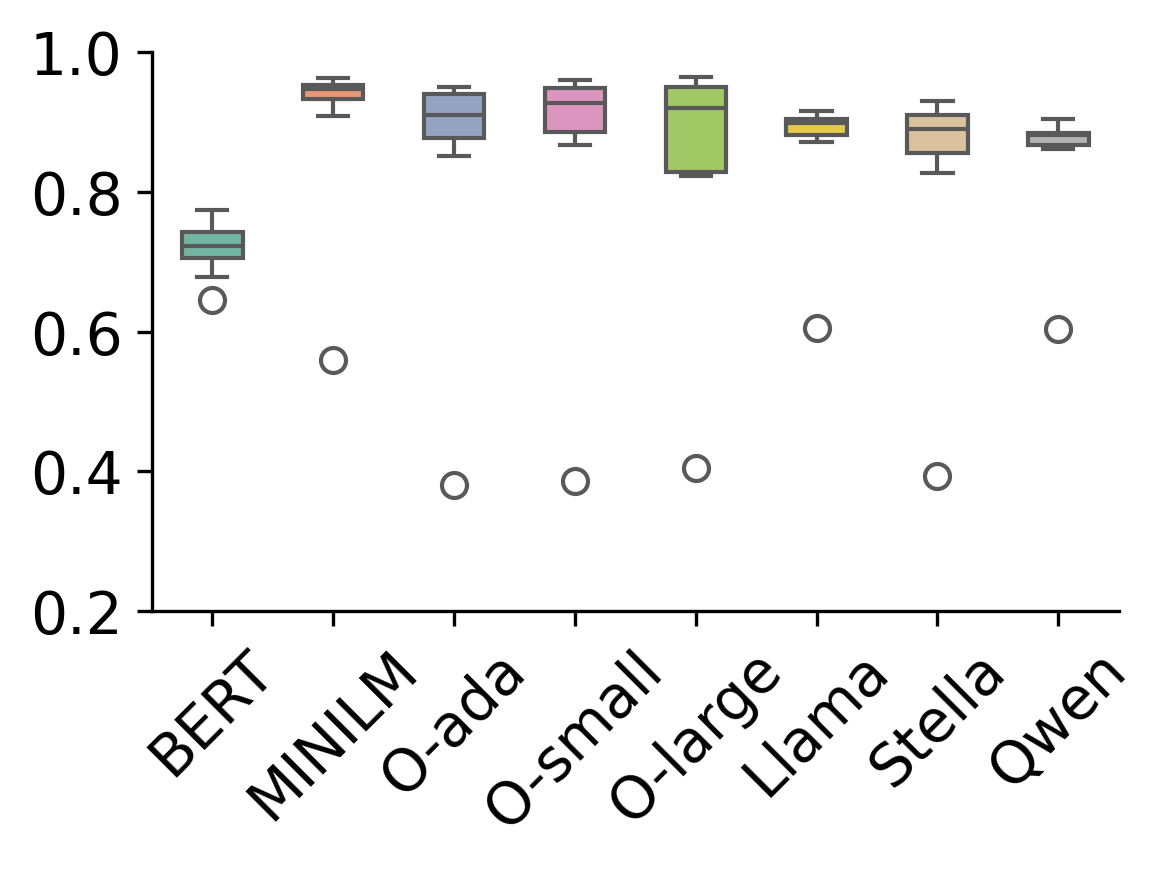}
         \caption{Email Spam}
         \label{es}
     \end{subfigure}
     % \hfill
     \begin{subfigure}[b]{0.3\textwidth}
         \centering
         \includegraphics[width=\textwidth]{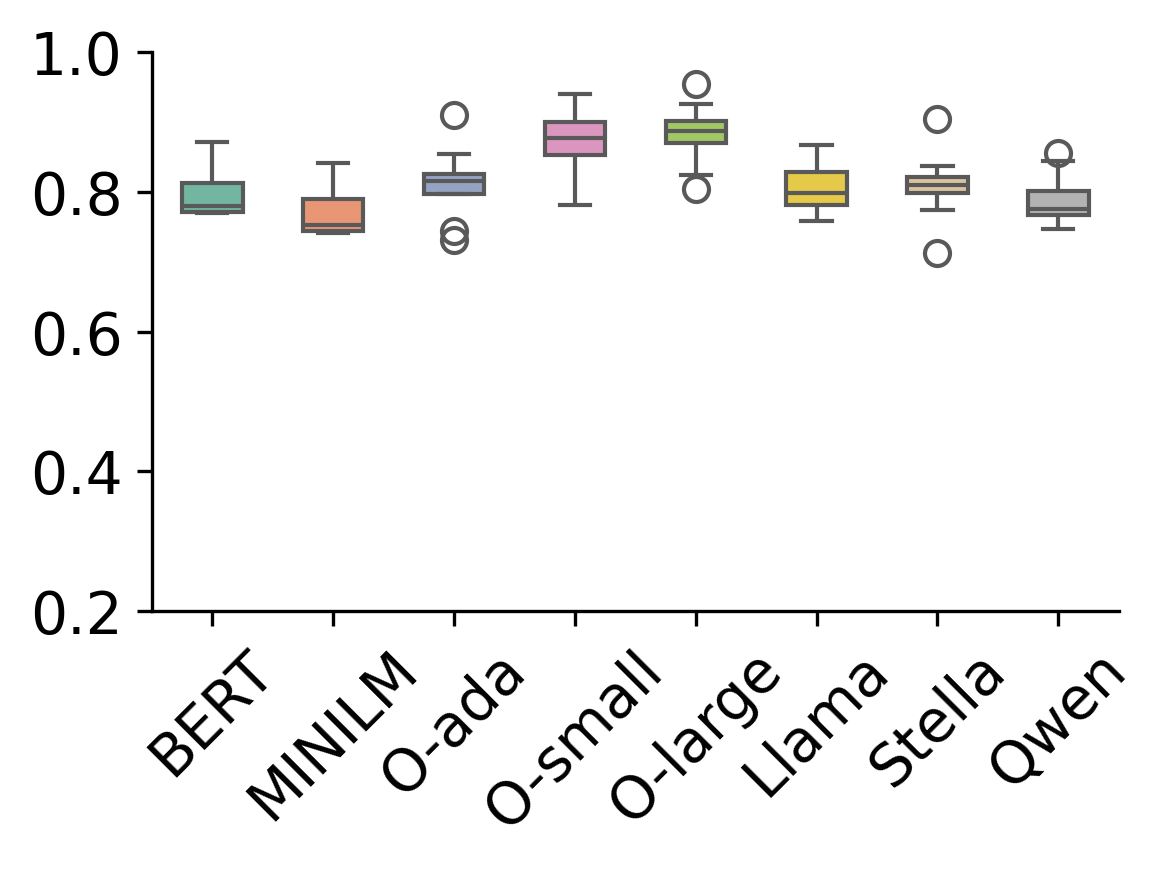}
         \caption{Covid Fake News}
         \label{cfn}
     \end{subfigure}
     % \hfill
     \begin{subfigure}[b]{0.3\textwidth}
         \centering
         \includegraphics[width=\textwidth]{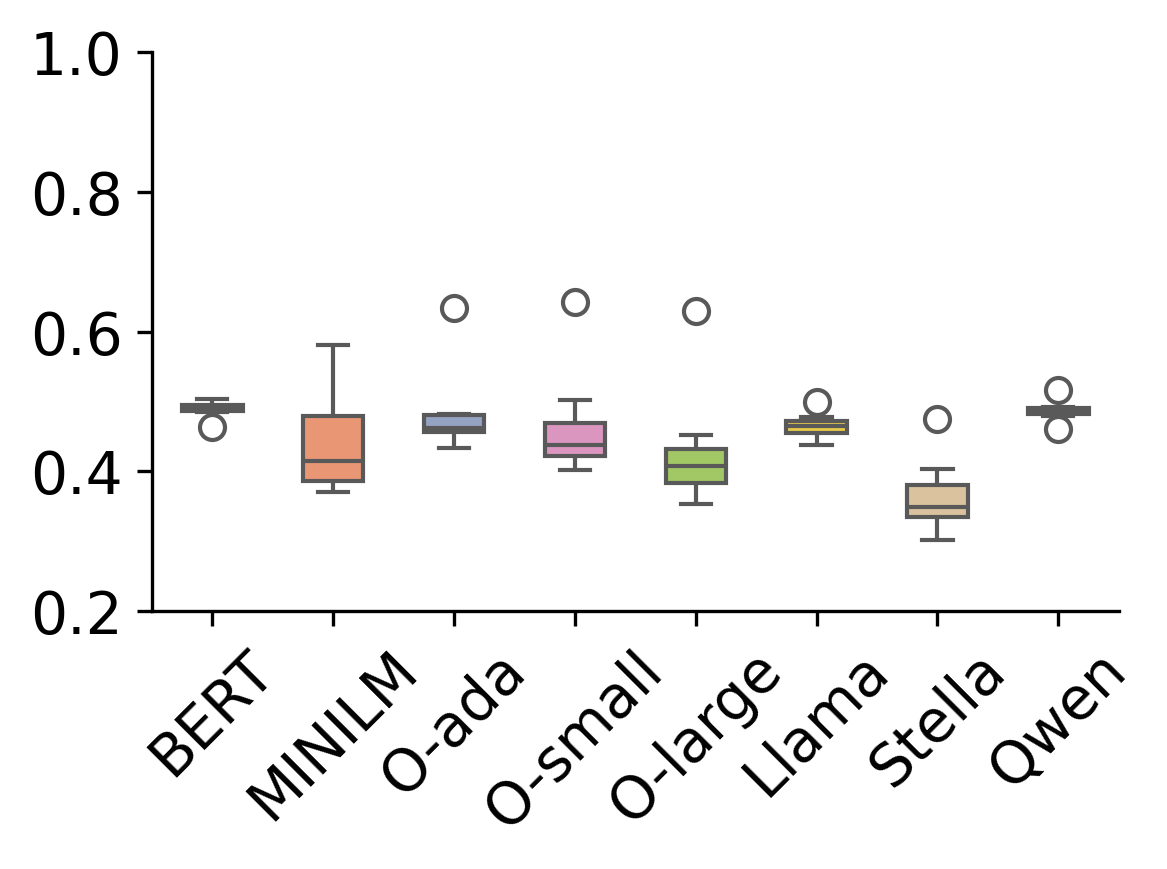}
         \caption{Hate Speech}
         \label{hs}
     \end{subfigure}

    \begin{subfigure}[b]{0.3\textwidth}
         \centering
         \includegraphics[width=\textwidth]{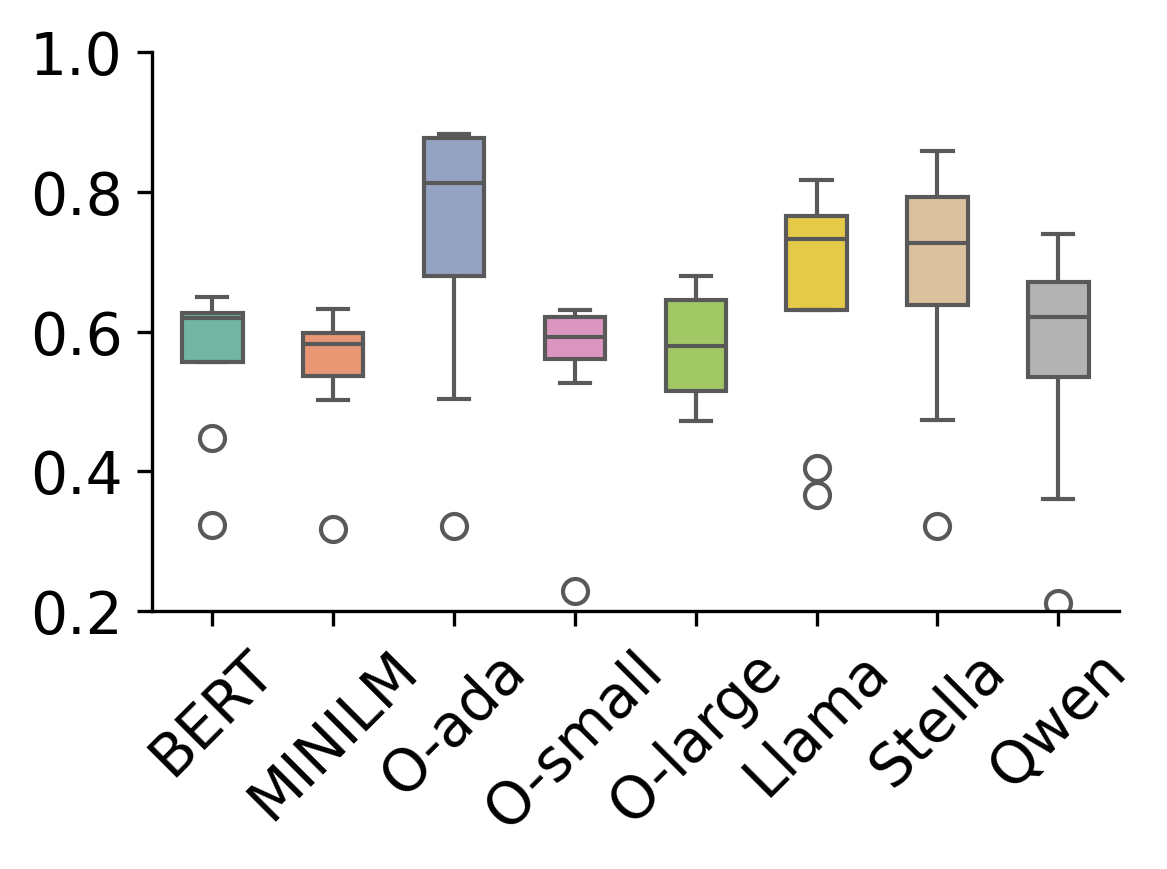}
         \caption{SMS Spam}
         \label{ss}
     \end{subfigure}
     % \hfill
     \begin{subfigure}[b]{0.3\textwidth}
         \centering
         \includegraphics[width=\textwidth]{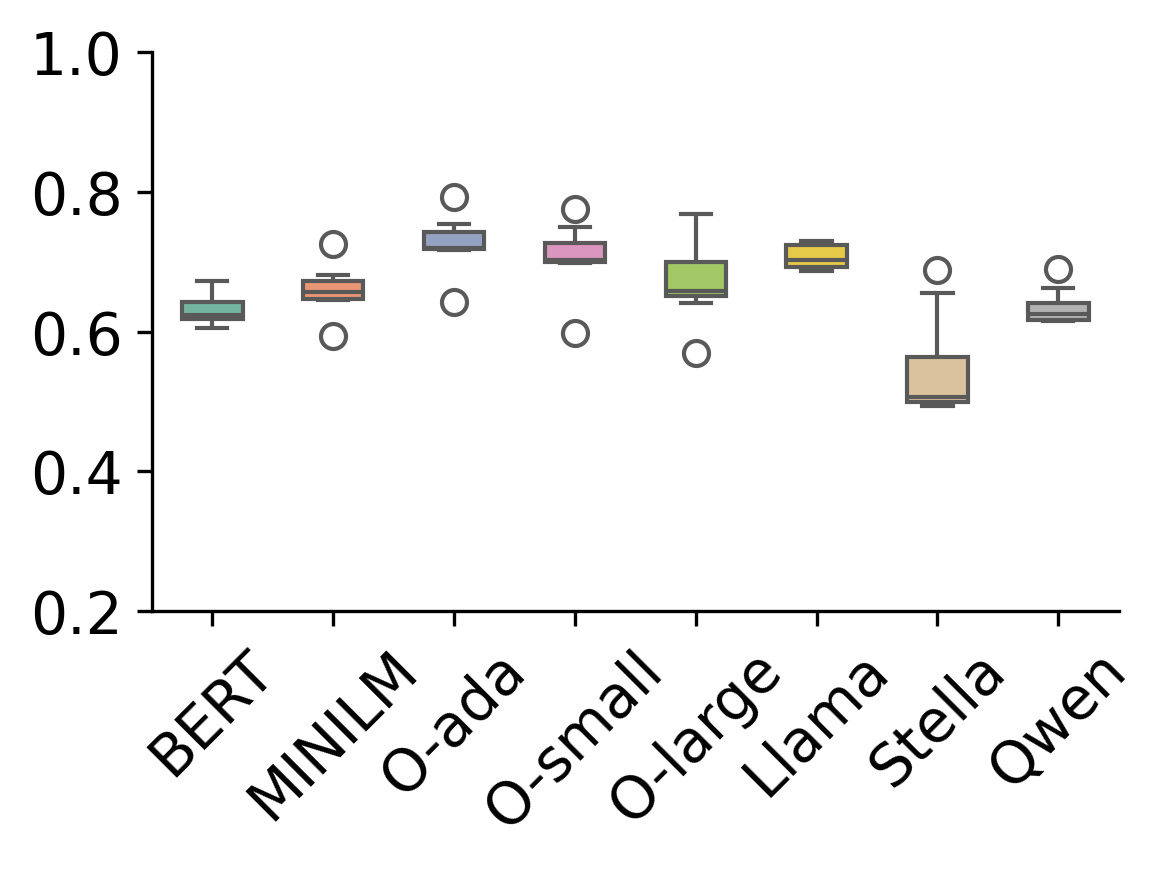}
         \caption{LIAR2}
         \label{liar2}
     \end{subfigure}
     % \hfill
     \begin{subfigure}[b]{0.3\textwidth}
         \centering
         \includegraphics[width=\textwidth]{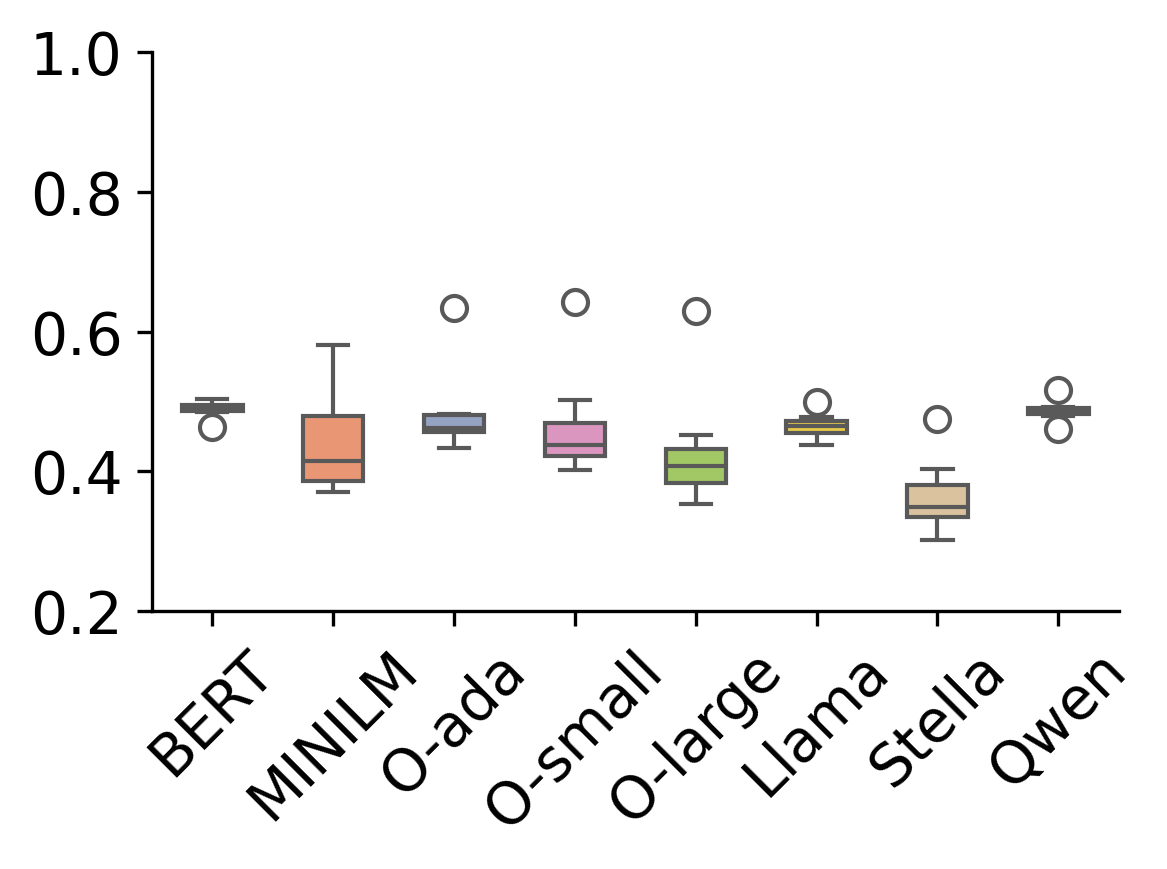}
         \caption{OLID}
         \label{olid}
     \end{subfigure}
        \caption{Boxplot of AUCROC scores for anomaly detectors on different embeddings across 6 datasets.}
        \label{result}
\end{figure*}

%=========

%=========
\subsection{Comparative Effectiveness of Embeddings in Anomaly Detection}\label{compar}
The results in Table~\ref{evaluation} demonstrate the remarkable capabilities of the OpenAI family of embeddings (O-ada, O-small, and O-large), consistently outperforming other embeddings across a variety of anomaly detection tasks. Specifically, O-ada achieves the highest average AUC scores with the ECOD detector (0.8822) on the SMS Spam dataset and with kNN (0.7921) on the LIAR2 dataset. Similarly, O-small demonstrates outstanding performance, achieving the highest AUC scores with kNN on the Hate Speech (0.6416) and OLID (0.5587) datasets. Additionally, O-large secures top AUC scores with COPOD (0.9639) on the Email Spam dataset and with kNN (0.9537) on the COVID Fake News dataset.

In comparison, other embeddings, such as MINILM, exhibit strong performance in specific tasks but lack consistency across more complex datasets. For instance, MINILM achieves exceptional AUC scores of 0.9526 and 0.9626 on the Email Spam datasets when paired with INNE and OCSVM, respectively. However, its performance declines significantly on datasets like OLID and LIAR2, suggesting limitations in capturing deeper contextual or domain-specific cues essential for these tasks. Similarly, embeddings such as stella and Qwen exhibit moderate performance, excelling in a limited subset of tasks but failing to match the versatility of OpenAI embeddings. Their inconsistent performance across datasets indicates that while they may effectively capture certain linguistic patterns, they struggle with tasks requiring a broader understanding of context, intent, or nuanced semantics. 

These observations suggest that OpenAI embeddings, deliver the most robust and consistent performance across a diverse set of tasks. Their ability to effectively capture both explicit textual features (e.g., in spam detection) and nuanced contextual variations (e.g., in Covid Fake News and OLID) highlights their versatility. This underscores their suitability for anomaly detection scenarios that demand both surface-level pattern recognition and deeper linguistic comprehension, making them well-equipped for handling a wide range of text-based anomalies.

% Moreover, the collective performance of the OpenAI family emphasizes the adaptability and efficacy of transformer-based embeddings in addressing anomaly detection challenges across datasets with diverse linguistic and semantic complexities. OpenAI embeddings consistently prove to be dominant and versatile, making them effective across various applications.

\subsection{Performance Across Anomaly Detectors}

\begin{figure}[!htbp]
     \centering
     \begin{subfigure}[b]{0.48\textwidth}
         \centering
         \includegraphics[width=0.95\textwidth]{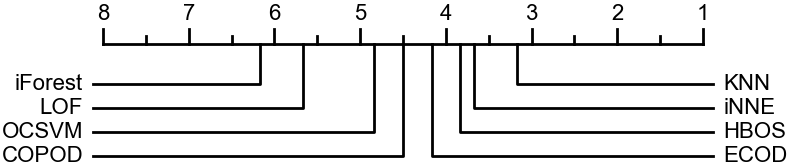}
         \caption{O-ada}
         \label{ada}
     \end{subfigure}

     \begin{subfigure}[b]{0.48\textwidth}
         \centering
         \includegraphics[width=0.95\textwidth]{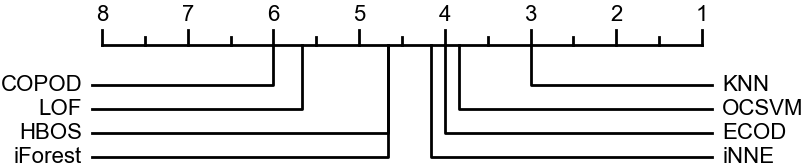}
         \caption{O-small}
         \label{small}
     \end{subfigure}

     \begin{subfigure}[b]{0.48\textwidth}
         \centering
         \includegraphics[width=0.95\textwidth]{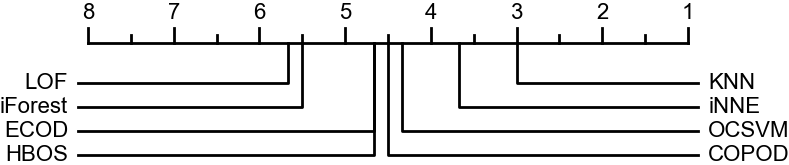}
         \caption{O-large}
         \label{large}
     \end{subfigure}
     \caption{Average rank (lower the better) of 3 differernt OpenAI embeddings-based methods on AUCROC across 6 datasets.}
     \label{rank}
\end{figure}

To evaluate the robustness of anomaly detection algorithms across various embeddings and tasks, we analyze their average rankings using OpenAI embeddings (O-ada, O-small, and O-large) as representative examples (Figure~\ref{rank}). These embeddings were selected based on their strong and consistent performance across datasets, as demonstrated in Section~\ref{compar}. The rankings provide insight into which detection algorithms perform reliably regardless of the embedding or task.

Across all three embeddings, kNN and INNE consistently rank as the top-performing algorithms. This indicates their robustness and adaptability to the semantic structures of LLM-derived embeddings. kNN, in particular, excels due to its ability to effectively model local density variations in feature space, making it well-suited for both explicit-pattern tasks like spam detection and nuanced tasks like fake news and hate speech detection. INNE, with its efficiency and strong generalization capabilities, complements kNN as a reliable alternative in diverse anomaly detection scenarios.

ECOD also ranks highly, consistently appearing among the top three detectors across embeddings. Its lightweight design and ability to estimate density-based anomalies make it a strong candidate for scenarios where computational efficiency is critical. On the other hand, methods like LOF, COPOD, and iForest consistently rank lower, highlighting their limitations in high-dimensional and semantically complex embedding spaces. These methods struggle with noise, data sparsity, and the nuanced patterns encoded in LLM embeddings, which limits their effectiveness across diverse tasks.
% Overall, kNN, INNE, and ECOD perform well, while LOF, COPOD, and iForest struggle with high-dimensional embeddings.

% Overall, kNN, INNE, and ECOD demonstrate robust and consistent performance across embeddings and tasks. In contrast, methods like LOF, COPOD, and iForest show weaker adaptability, likely due to their limitations in handling high-dimensional and semantically rich embeddings.

%=======================================================
\section{Conclusion}
In this paper, we present a comprehensive benchmark for embedding-based text anomaly detection, systematically evaluating the interplay between LLM embeddings and classical anomaly detection algorithms across three diverse domains. Our results reveal both the strengths and limitations of embedding-based anomaly detection methods, demonstrating their effectiveness in tasks with explicit and well-defined patterns while highlighting challenges in capturing implicit, context-dependent anomalies that require broader contextual cues. These findings emphasize the need for more adaptive embeddings and hybrid detection strategies that integrate external knowledge and contextual reasoning.

%=======================================================
\section*{Limitations}
TAD-Bench evaluates anomaly detection across three domains: spam detection, fake news detection, and offensive language detection. While these tasks provide diverse and relevant benchmarks, they do not fully capture the complexity of real-world applications. Strong performance in spam detection highlights the ability of LLM embeddings to capture explicit patterns, while mixed results in fake news detection and poor performance in offensive language detection reveal their limitations in modeling implicit, context-sensitive cues. Expanding to domains like medical, financial, or legal texts that involve unique challenges, and exploring datasets with more implicit anomalies, could better evaluate the adaptability and robustness of these methods.

% In addition, our work uses pre-trained LLM embeddings and default hyperparameters for anomaly detectors, ensuring consistency but potentially underestimating their best-case performance. Fine-tuning LLMs on domain-specific data could improve embedding quality, while systematic hyperparameter optimization might unlock the full potential of anomaly detectors. Future research should explore these directions, leveraging techniques like AutoML to streamline both embedding fine-tuning and parameter tuning, thereby achieving more competitive performance.

Moreover, TAD-Bench focuses solely on embedding-based methods, excluding end-to-end approaches that directly process raw text because due to modularity and efficiency of embedding-based methods, and NLP-ADBench has also shown better performance of embedding-based methods than end-to-end methods. However, end-to-end models like autoencoders or transformer-based methods may capture richer contextual information and handle more complex anomalies. Future work should incorporate end-to-end models and explore hybrid approaches that combine the strengths of both paradigms, providing a more comprehensive evaluation of anomaly detection methods in NLP.

% Furthermore, due to computational resource constraints, we primarily evaluate small-scale LLMs rather than larger, more powerful models. While this allows for a fair comparison across different embedding methods, it may not fully reflect the capabilities of state-of-the-art LLMs in anomaly detection. Future studies could benefit from leveraging larger models with extended context windows and more sophisticated representations to assess their impact on anomaly detection performance.

%=======================================================
\section*{Ethic Statement}

This study adheres to ethical research practices and considerations in the development and evaluation of text anomaly detection methods.  

\textbf{Use of Potentially Offensive Language.} Some examples in this paper may contain offensive, harmful, or misleading language. These examples are used purely for illustrative purposes to demonstrate the challenges of text anomaly detection in real-world scenarios. They do not reflect the opinions, beliefs, or endorsements of the authors.  

\textbf{Data Sources and Usage.} All datasets used in this study are sourced from publicly available research datasets that have been previously used in NLP and anomaly detection research. Proper citations and references to the original datasets are provided in the paper. No private, proprietary, or personally identifiable information was used in this study.  

\textbf{Risks and Responsible Use.} Because anomaly detection models can be misused for purposes such as censorship, surveillance, or unfair content moderation. We strongly emphasize that our benchmark is intended for research and academic purposes only and should be used responsibly with consideration of ethical and societal implications.  

\textbf{Use of AI Assistance} We acknowledge the use of AI-based writing assistants for grammar refinement, spelling correction, and improving the clarity of our manuscript. However, all intellectual contributions, experimental designs, analyses, and conclusions in this paper are solely the work of the authors. 

% #### **Bias and Fairness Considerations**  
% Anomaly detection in text can be **subject to biases**, particularly when dealing with sensitive topics such as **hate speech, misinformation, and spam detection**. We acknowledge that **embedding-based models and anomaly detection algorithms may inherit biases** from their training data. While our benchmark systematically evaluates different models and detection methods, we do not claim that any specific approach is **free from bias** or suitable for deployment in **high-stakes decision-making** without further validation.  

% #### **Reproducibility and Transparency**  
% To ensure transparency and reproducibility, we provide details on dataset processing, model selection, and evaluation **in the main paper and appendix**. We use **default hyperparameters** for anomaly detection methods to avoid cherry-picking results and ensure fair comparisons.  
% By acknowledging these ethical considerations, we aim to promote **responsible AI research and development**, ensuring that anomaly detection methods are evaluated rigorously and applied in ways that minimize harm and maximize fairness.

%=======================================================
\bibliography{custom}

\begin{thebibliography}{44}
\providecommand{\natexlab}[1]{#1}

\bibitem[{Almeida et~al.(2011)Almeida, Hidalgo, and Yamakami}]{almeida2011contributions}
Tiago~A Almeida, Jos{\'e} Mar{\'\i}a~G Hidalgo, and Akebo Yamakami. 2011.
\newblock Contributions to the study of sms spam filtering: new collection and results.
\newblock In \emph{Proceedings of the 11th ACM symposium on Document engineering}, pages 259--262.

\bibitem[{Bandaragoda et~al.(2018)Bandaragoda, Ting, Albrecht, Liu, Zhu, and Wells}]{bandaragoda2018isolation}
Tharindu~R Bandaragoda, Kai~Ming Ting, David Albrecht, Fei~Tony Liu, Ye~Zhu, and Jonathan~R Wells. 2018.
\newblock Isolation-based anomaly detection using nearest-neighbor ensembles.
\newblock \emph{Computational Intelligence}, 34(4):968--998.

\bibitem[{Bejan et~al.(2023)Bejan, Manolache, and Popescu}]{bejan2023ad}
Matei Bejan, Andrei Manolache, and Marius Popescu. 2023.
\newblock Ad-nlp: A benchmark for anomaly detection in natural language processing.
\newblock In \emph{Proceedings of the 2023 Conference on Empirical Methods in Natural Language Processing}, pages 10766--10778.

\bibitem[{Bl{\'a}zquez-Garc{\'\i}a et~al.(2021)Bl{\'a}zquez-Garc{\'\i}a, Conde, Mori, and Lozano}]{blazquez2021review}
Ane Bl{\'a}zquez-Garc{\'\i}a, Angel Conde, Usue Mori, and Jose~A Lozano. 2021.
\newblock A review on outlier/anomaly detection in time series data.
\newblock \emph{ACM computing surveys (CSUR)}, 54(3):1--33.

\bibitem[{Breunig et~al.(2000)Breunig, Kriegel, Ng, and Sander}]{breunig2000lof}
Markus~M Breunig, Hans-Peter Kriegel, Raymond~T Ng, and J{\"o}rg Sander. 2000.
\newblock Lof: identifying density-based local outliers.
\newblock In \emph{Proceedings of the 2000 ACM SIGMOD international conference on Management of data}, pages 93--104.

\bibitem[{Brown et~al.(2020)Brown, Mann, Ryder, Subbiah, Kaplan, Dhariwal, Neelakantan, Shyam, Sastry, Askell et~al.}]{brown2020language}
Tom Brown, Benjamin Mann, Nick Ryder, Melanie Subbiah, Jared~D Kaplan, Prafulla Dhariwal, Arvind Neelakantan, Pranav Shyam, Girish Sastry, Amanda Askell, et~al. 2020.
\newblock Language models are few-shot learners.
\newblock \emph{Advances in neural information processing systems}, 33:1877--1901.

\bibitem[{da~Costa et~al.(2023)da~Costa, Oliveira, and Fileto}]{da2023text}
Liliane~Soares da~Costa, Italo~L Oliveira, and Renato Fileto. 2023.
\newblock Text classification using embeddings: a survey.
\newblock \emph{Knowledge and Information Systems}, 65(7):2761--2803.

\bibitem[{Das et~al.(2021)Das, Basak, and Dutta}]{das2021heuristic}
Sourya~Dipta Das, Ayan Basak, and Saikat Dutta. 2021.
\newblock A heuristic-driven ensemble framework for covid-19 fake news detection.
\newblock In \emph{Combating Online Hostile Posts in Regional Languages during Emergency Situation: First International Workshop, CONSTRAINT 2021, Collocated with AAAI 2021, Virtual Event, February 8, 2021, Revised Selected Papers 1}, pages 164--176. Springer.

\bibitem[{Davidson et~al.(2017)Davidson, Warmsley, Macy, and Weber}]{davidson2017automated}
Thomas Davidson, Dana Warmsley, Michael Macy, and Ingmar Weber. 2017.
\newblock Automated hate speech detection and the problem of offensive language.
\newblock In \emph{Proceedings of the international AAAI conference on web and social media}, volume~11, pages 512--515.

\bibitem[{Devlin et~al.(2019)Devlin, Chang, Lee, and Toutanova}]{devlin-etal-2019-bert}
Jacob Devlin, Ming-Wei Chang, Kenton Lee, and Kristina Toutanova. 2019.
\newblock \href {https://doi.org/10.18653/v1/N19-1423} {{BERT}: Pre-training of deep bidirectional transformers for language understanding}.
\newblock In \emph{Proceedings of the 2019 Conference of the North {A}merican Chapter of the Association for Computational Linguistics: Human Language Technologies, Volume 1 (Long and Short Papers)}, pages 4171--4186, Minneapolis, Minnesota. Association for Computational Linguistics.

\bibitem[{Goldstein and Dengel(2012)}]{goldstein2012histogram}
Markus Goldstein and Andreas Dengel. 2012.
\newblock Histogram-based outlier score (hbos): A fast unsupervised anomaly detection algorithm.
\newblock \emph{KI-2012: poster and demo track}, 1:59--63.

\bibitem[{Goodge et~al.(2022)Goodge, Hooi, Ng, and Ng}]{goodge2022lunar}
Adam Goodge, Bryan Hooi, See-Kiong Ng, and Wee~Siong Ng. 2022.
\newblock Lunar: Unifying local outlier detection methods via graph neural networks.
\newblock In \emph{Proceedings of the AAAI Conference on Artificial Intelligence}, volume~36, pages 6737--6745.

\bibitem[{Lan et~al.(2020)Lan, Chen, Goodman, Gimpel, Sharma, and Soricut}]{lan2020albertlitebertselfsupervised}
Zhenzhong Lan, Mingda Chen, Sebastian Goodman, Kevin Gimpel, Piyush Sharma, and Radu Soricut. 2020.
\newblock \href {https://arxiv.org/abs/1909.11942} {Albert: A lite bert for self-supervised learning of language representations}.
\newblock \emph{Preprint}, arXiv:1909.11942.

\bibitem[{Li et~al.(2024)Li, Li, Xiao, Yang, Nian, Hu, and Zhao}]{li2024nlp}
Yuangang Li, Jiaqi Li, Zhuo Xiao, Tiankai Yang, Yi~Nian, Xiyang Hu, and Yue Zhao. 2024.
\newblock Nlp-adbench: Nlp anomaly detection benchmark.
\newblock \emph{arXiv preprint arXiv:2412.04784}.

\bibitem[{Li et~al.(2020)Li, Zhao, Botta, Ionescu, and Hu}]{li2020copod}
Zheng Li, Yue Zhao, Nicola Botta, Cezar Ionescu, and Xiyang Hu. 2020.
\newblock Copod: copula-based outlier detection.
\newblock In \emph{2020 IEEE international conference on data mining (ICDM)}, pages 1118--1123. IEEE.

\bibitem[{Li et~al.(2022)Li, Zhao, Hu, Botta, Ionescu, and Chen}]{li2022ecod}
Zheng Li, Yue Zhao, Xiyang Hu, Nicola Botta, Cezar Ionescu, and George~H Chen. 2022.
\newblock Ecod: Unsupervised outlier detection using empirical cumulative distribution functions.
\newblock \emph{IEEE Transactions on Knowledge and Data Engineering}, 35(12):12181--12193.

\bibitem[{Liu et~al.(2008)Liu, Ting, and Zhou}]{liu2008isolation}
Fei~Tony Liu, Kai~Ming Ting, and Zhi-Hua Zhou. 2008.
\newblock Isolation forest.
\newblock In \emph{2008 eighth ieee international conference on data mining}, pages 413--422. IEEE.

\bibitem[{Liu et~al.(2012)Liu, Ting, and Zhou}]{liu2012isolation}
Fei~Tony Liu, Kai~Ming Ting, and Zhi-Hua Zhou. 2012.
\newblock Isolation-based anomaly detection.
\newblock \emph{ACM Transactions on Knowledge Discovery from Data (TKDD)}, 6(1):1--39.

\bibitem[{Metsis et~al.(2006)Metsis, Androutsopoulos, and Paliouras}]{metsis2006spam}
Vangelis Metsis, Ion Androutsopoulos, and Georgios Paliouras. 2006.
\newblock Spam filtering with naive bayes-which naive bayes?
\newblock In \emph{CEAS}, volume~17, pages 28--69. Mountain View, CA.

\bibitem[{Mikolov(2013)}]{mikolov2013efficient}
Tomas Mikolov. 2013.
\newblock Efficient estimation of word representations in vector space.
\newblock \emph{arXiv preprint arXiv:1301.3781}, 3781.

\bibitem[{OpenAI(2024)}]{openai2023new_embeddings}
OpenAI. 2024.
\newblock \href {https://openai.com/index/new-embedding-models-and-api-updates/} {New embedding models and api updates}.

\bibitem[{Pang et~al.(2021)Pang, Shen, Cao, and Hengel}]{pang2021deep}
Guansong Pang, Chunhua Shen, Longbing Cao, and Anton Van~Den Hengel. 2021.
\newblock Deep learning for anomaly detection: A review.
\newblock \emph{ACM computing surveys (CSUR)}, 54(2):1--38.

\bibitem[{Pennington et~al.(2014)Pennington, Socher, and Manning}]{pennington2014glove}
Jeffrey Pennington, Richard Socher, and Christopher~D Manning. 2014.
\newblock Glove: Global vectors for word representation.
\newblock In \emph{Proceedings of the 2014 conference on empirical methods in natural language processing (EMNLP)}, pages 1532--1543.

\bibitem[{Peters et~al.(2018)Peters, Neumann, Iyyer, Gardner, Clark, Lee, and Zettlemoyer}]{peters-etal-2018-deep}
Matthew~E. Peters, Mark Neumann, Mohit Iyyer, Matt Gardner, Christopher Clark, Kenton Lee, and Luke Zettlemoyer. 2018.
\newblock \href {https://doi.org/10.18653/v1/N18-1202} {Deep contextualized word representations}.
\newblock In \emph{Proceedings of the 2018 Conference of the North {A}merican Chapter of the Association for Computational Linguistics: Human Language Technologies, Volume 1 (Long Papers)}, pages 2227--2237, New Orleans, Louisiana. Association for Computational Linguistics.

\bibitem[{Pimentel et~al.(2014)Pimentel, Clifton, Clifton, and Tarassenko}]{pimentel2014review}
Marco~AF Pimentel, David~A Clifton, Lei Clifton, and Lionel Tarassenko. 2014.
\newblock A review of novelty detection.
\newblock \emph{Signal processing}, 99:215--249.

\bibitem[{Qiao et~al.(2024)Qiao, Tong, An, King, Aggarwal, and Pang}]{qiao2024deep}
Hezhe Qiao, Hanghang Tong, Bo~An, Irwin King, Charu Aggarwal, and Guansong Pang. 2024.
\newblock Deep graph anomaly detection: A survey and new perspectives.
\newblock \emph{arXiv preprint arXiv:2409.09957}.

\bibitem[{Ramaswamy et~al.(2000)Ramaswamy, Rastogi, and Shim}]{ramaswamy2000efficient}
Sridhar Ramaswamy, Rajeev Rastogi, and Kyuseok Shim. 2000.
\newblock Efficient algorithms for mining outliers from large data sets.
\newblock In \emph{Proceedings of the 2000 ACM SIGMOD international conference on Management of data}, pages 427--438.

\bibitem[{Ruff et~al.(2018)Ruff, Vandermeulen, Goernitz, Deecke, Siddiqui, Binder, M{\"u}ller, and Kloft}]{ruff2018deep}
Lukas Ruff, Robert Vandermeulen, Nico Goernitz, Lucas Deecke, Shoaib~Ahmed Siddiqui, Alexander Binder, Emmanuel M{\"u}ller, and Marius Kloft. 2018.
\newblock Deep one-class classification.
\newblock In \emph{International conference on machine learning}, pages 4393--4402. PMLR.

\bibitem[{Salton and Buckley(1988)}]{salton1988term}
Gerard Salton and Christopher Buckley. 1988.
\newblock Term-weighting approaches in automatic text retrieval.
\newblock \emph{Information processing \& management}, 24(5):513--523.

\bibitem[{Sch{\"o}lkopf et~al.(2001)Sch{\"o}lkopf, Platt, Shawe-Taylor, Smola, and Williamson}]{scholkopf2001estimating}
Bernhard Sch{\"o}lkopf, John~C Platt, John Shawe-Taylor, Alex~J Smola, and Robert~C Williamson. 2001.
\newblock Estimating the support of a high-dimensional distribution.
\newblock \emph{Neural computation}, 13(7):1443--1471.

\bibitem[{Steinbuss and B{\"o}hm(2021)}]{steinbuss2021benchmarking}
Georg Steinbuss and Klemens B{\"o}hm. 2021.
\newblock Benchmarking unsupervised outlier detection with realistic synthetic data.
\newblock \emph{ACM Transactions on Knowledge Discovery from Data (TKDD)}, 15(4):1--20.

\bibitem[{Team(2024)}]{qwen2.5}
Qwen Team. 2024.
\newblock \href {https://qwenlm.github.io/blog/qwen2.5/} {Qwen2.5: A party of foundation models}.

\bibitem[{Wang et~al.(2024)Wang, Xu, and Li}]{wang2024log2graphs}
Caihong Wang, Du~Xu, and Zonghang Li. 2024.
\newblock Log2graphs: An unsupervised framework for log anomaly detection with efficient feature extraction.
\newblock \emph{arXiv preprint arXiv:2409.11890}.

\bibitem[{Wang et~al.(2020)Wang, Wei, Dong, Bao, Yang, and Zhou}]{wang2020minilm}
Wenhui Wang, Furu Wei, Li~Dong, Hangbo Bao, Nan Yang, and Ming Zhou. 2020.
\newblock Minilm: Deep self-attention distillation for task-agnostic compression of pre-trained transformers.
\newblock \emph{Advances in Neural Information Processing Systems}, 33:5776--5788.

\bibitem[{Xu and Kechadi(2024)}]{xu2024enhanced}
Cheng Xu and M-Tahar Kechadi. 2024.
\newblock \href {https://doi.org/10.1109/ACCESS.2024.3418340} {An enhanced fake news detection system with fuzzy deep learning}.
\newblock \emph{IEEE Access}, 12:88006--88021.

\bibitem[{Xu et~al.(2023)Xu, Milleret, and Segond}]{xu2023comparative}
Yizhou Xu, J{\'e}r{\^o}me Milleret, and Fr{\'e}d{\'e}rique Segond. 2023.
\newblock Comparative analysis of anomaly detection algorithms in text data.
\newblock In \emph{Proceedings of the 14th International Conference on Recent Advances in Natural Language Processing}, pages 1234--1245.

\bibitem[{Yang et~al.(2024{\natexlab{a}})Yang, Yang, Hui, Zheng, Yu, Zhou, Li, Li, Liu, Huang, Dong, Wei, Lin, Tang, Wang, Yang, Tu, Zhang, Ma, Xu, Zhou, Bai, He, Lin, Dang, Lu, Chen, Yang, Li, Xue, Ni, Zhang, Wang, Peng, Men, Gao, Lin, Wang, Bai, Tan, Zhu, Li, Liu, Ge, Deng, Zhou, Ren, Zhang, Wei, Ren, Fan, Yao, Zhang, Wan, Chu, Liu, Cui, Zhang, and Fan}]{qwen2}
An~Yang, Baosong Yang, Binyuan Hui, Bo~Zheng, Bowen Yu, Chang Zhou, Chengpeng Li, Chengyuan Li, Dayiheng Liu, Fei Huang, Guanting Dong, Haoran Wei, Huan Lin, Jialong Tang, Jialin Wang, Jian Yang, Jianhong Tu, Jianwei Zhang, Jianxin Ma, Jin Xu, Jingren Zhou, Jinze Bai, Jinzheng He, Junyang Lin, Kai Dang, Keming Lu, Keqin Chen, Kexin Yang, Mei Li, Mingfeng Xue, Na~Ni, Pei Zhang, Peng Wang, Ru~Peng, Rui Men, Ruize Gao, Runji Lin, Shijie Wang, Shuai Bai, Sinan Tan, Tianhang Zhu, Tianhao Li, Tianyu Liu, Wenbin Ge, Xiaodong Deng, Xiaohuan Zhou, Xingzhang Ren, Xinyu Zhang, Xipin Wei, Xuancheng Ren, Yang Fan, Yang Yao, Yichang Zhang, Yu~Wan, Yunfei Chu, Yuqiong Liu, Zeyu Cui, Zhenru Zhang, and Zhihao Fan. 2024{\natexlab{a}}.
\newblock Qwen2 technical report.
\newblock \emph{arXiv preprint arXiv:2407.10671}.

\bibitem[{Yang et~al.(2024{\natexlab{b}})Yang, Nian, Li, Xu, Li, Li, Xiao, Hu, Rossi, Ding et~al.}]{yang2024ad}
Tiankai Yang, Yi~Nian, Shawn Li, Ruiyao Xu, Yuangang Li, Jiaqi Li, Zhuo Xiao, Xiyang Hu, Ryan Rossi, Kaize Ding, et~al. 2024{\natexlab{b}}.
\newblock Ad-llm: Benchmarking large language models for anomaly detection.
\newblock \emph{arXiv preprint arXiv:2412.11142}.

\bibitem[{Zampieri et~al.(2019{\natexlab{a}})Zampieri, Malmasi, Nakov, Rosenthal, Farra, and Kumar}]{zampierietal2019}
Marcos Zampieri, Shervin Malmasi, Preslav Nakov, Sara Rosenthal, Noura Farra, and Ritesh Kumar. 2019{\natexlab{a}}.
\newblock {Predicting the Type and Target of Offensive Posts in Social Media}.
\newblock In \emph{Proceedings of NAACL}.

\bibitem[{Zampieri et~al.(2019{\natexlab{b}})Zampieri, Malmasi, Nakov, Rosenthal, Farra, and Kumar}]{zampieri-etal-2019-predicting}
Marcos Zampieri, Shervin Malmasi, Preslav Nakov, Sara Rosenthal, Noura Farra, and Ritesh Kumar. 2019{\natexlab{b}}.
\newblock \href {https://doi.org/10.18653/v1/N19-1144} {Predicting the type and target of offensive posts in social media}.
\newblock In \emph{Proceedings of the 2019 Conference of the North {A}merican Chapter of the Association for Computational Linguistics: Human Language Technologies, Volume 1 (Long and Short Papers)}, pages 1415--1420, Minneapolis, Minnesota. Association for Computational Linguistics.

\bibitem[{Zhang et~al.(2024)Zhang, Li, Zeng, and Wang}]{zhang2024jasper}
Dun Zhang, Jiacheng Li, Ziyang Zeng, and Fulong Wang. 2024.
\newblock Jasper and stella: distillation of sota embedding models.
\newblock \emph{arXiv preprint arXiv:2412.19048}.

\bibitem[{Zhao et~al.(2019)Zhao, Nasrullah, and Li}]{zhao2019pyod}
Yue Zhao, Zain Nasrullah, and Zheng Li. 2019.
\newblock \href {http://jmlr.org/papers/v20/19-011.html} {Pyod: A python toolbox for scalable outlier detection}.
\newblock \emph{Journal of Machine Learning Research}, 20(96):1--7.

\bibitem[{Zhu et~al.(2023)Zhu, Yuan, Wang, Liu, Liu, Deng, Chen, Liu, Dou, and Wen}]{zhu2023large}
Yutao Zhu, Huaying Yuan, Shuting Wang, Jiongnan Liu, Wenhan Liu, Chenlong Deng, Haonan Chen, Zheng Liu, Zhicheng Dou, and Ji-Rong Wen. 2023.
\newblock Large language models for information retrieval: A survey.
\newblock \emph{arXiv preprint arXiv:2308.07107}.

\bibitem[{Zhuang et~al.(2021)Zhuang, Wayne, Ya, and Jun}]{zhuang-etal-2021-robustly}
Liu Zhuang, Lin Wayne, Shi Ya, and Zhao Jun. 2021.
\newblock \href {https://aclanthology.org/2021.ccl-1.108/} {A robustly optimized {BERT} pre-training approach with post-training}.
\newblock In \emph{Proceedings of the 20th Chinese National Conference on Computational Linguistics}, pages 1218--1227, Huhhot, China. Chinese Information Processing Society of China.

\end{thebibliography}

%=======================================================
\clearpage

% \onecolumn
\appendix

\section{Problem Definitions}
\label{pd}
% In the context of NLP, an anomaly refers to a text instance that exhibits characteristics or patterns that deviate significantly from the majority of the dataset. Such anomalies can manifest in various ways, including rare or niche topics, unusual or complex syntax and semantics, domain-specific jargon, or even intentionally manipulated language, such as spam, fake news, deceptive reviews, or offensive and harmful content. Detecting these anomalies is crucial for numerous real-world applications, such as content moderation, fraud detection, cybersecurity threat analysis, and identifying novel or emerging patterns in large-scale text corpora to enhance decision-making and knowledge discovery.

Let $\mathcal{D} = \{x_1, x_2, \dots, x_N\}$ represent a corpus consisting of $N$ textual instances, where each instance $x_i \in \mathcal{X}$ is represented as a sequence of tokens: 
\begin{equation}
x_i = [t_1, t_2, \dots, t_{L_i}], \nonumber
\end{equation}
where $L_i$ denotes the sequence length of $x_i$. 
% \sikun{The notation $t_j$ is not defined, right!} 
The goal of text anomaly detection is to identify a subset of instances $\mathcal{D}_{\text{anomaly}} \subset \mathcal{D}$, such that $\mathcal{D}_{\text{anomaly}}$ contains samples that deviate significantly from the majority of the dataset $\mathcal{D}_{\text{normal}} = \mathcal{D} \setminus \mathcal{D}_{\text{anomaly}}$.

To achieve this, an anomaly detection algorithm $g$ is applied to the representations of the textual instances to identify potential anomalies. (1) Each text instance $x_i$ is first mapped to a fixed-dimensional vector $\mathbf{z}_i \in \mathbb{R}^d$ using an embedding model $\phi: \mathcal{X} \to \mathbb{R}^d$, such that $\mathbf{z}_i = \phi(x_i)$. (2) The anomaly detection algorithm then assigns an anomaly score $s_i = g(\mathbf{z}_i)$ to each instance, $s_i \in [0, 1]$. 
% \sikun{I am wondering whether we should specify the range of the anomaly score, e.g., nonnegative, or interval-bounded.} 
Based on a predefined threshold $\tau$, an instance $x_i$ is classified as anomalous if:
\[
x_i \in \mathcal{D}_{\text{anomaly}} \iff s_i \geq \tau.
\]

The objective of text anomaly detection is to ensure that $g$ effectively distinguishes between normal and anomalous instances, even in the absence of labeled data, while being robust to the inherent variability and high dimensionality of textual data.

\section{Clarification Between Anomaly and Novelty Detection}

Text Anomaly Detection (TAD), as defined in Section~\ref{pd}, focuses on identifying instances that deviate significantly from the majority of a dataset, regardless of whether anomalies are present during training. While some prior studies (e.g., AD-NLP~\cite{bejan2023ad}, NLP-ADBench~\cite{li2024nlp} and AD-LLM~\cite{yang2024ad}) assume training data contains only normal instances and testing data includes both normal and anomalous samples, this setup aligns more closely with novelty detection~\cite{pimentel2014review}. Novelty detection specifically targets never-before-seen anomalies that are absent from the training phase, often treating anomalies as entirely novel classes.

In contrast, our benchmark evaluates a broader spectrum of anomaly detection scenarios. We do not restrict the training data to purely normal instances, allowing for potential partial supervision or contaminated training sets (e.g., realistic scenarios where anomalies may unintentionally exist in training data). This setup reflects real-world applications where anomaly types are not always fully known a prior, and detection systems must generalize across domains and anomaly types. 
% Our experiments further evaluate robustness to domain shifts and implicit anomalies (e.g., hate speech), which require capturing nuanced contextual patterns rather than merely detecting novel distributions.

% Key distinctions between our framework and novelty detection approaches include:
% \begin{itemize}
%     \item Training Data Assumption: Unlike novelty detection, we do not enforce a strict "normal-only" training regime, enabling adaptability to scenarios where anomalies may leak into training data.
%     \item Anomaly Scope: Our benchmark includes anomalies that may overlap partially with training data (e.g., known spam patterns) and those requiring deeper semantic understanding (e.g., implicit hate speech), whereas novelty detection typically assumes anomalies are entirely unseen.
%     \item Generalization Focus: We prioritize robustness across heterogeneous domains and anomaly types, rather than optimizing for detection of strictly novel classes.
% \end{itemize}

This distinction underscores our goal of advancing generalizable anomaly detection systems for real-world NLP applications, where anomalies may exhibit both explicit and context-dependent patterns.

%=======================================================
\section{Datasets}

%=======================================================
\begin{table*}[!htbp]
\caption{Embedding time of 6 datasets in seconds.}
\label{time}
% \resizebox{1\textwidth}{0.5\textheight}{
\begin{tabular}{c|rrrrrr}
\hline
\textbf{Embeddings}   & \textbf{Email-Spam} & \textbf{SMS-Spam}         & \textbf{COVID-Fake} & \textbf{LIAR2}  & \textbf{Hate-Speech} & \textbf{OLID} \\ \hline
BERT   & 64.95 s             & 10.87 s                   & 4.48 s              & 3.7 s           & 9.79 s               & 1.81 s          \\ 
MINILM & 3.58 s              & 1.48 s                    & 0.64 s              & 0.52 s          & 1.50 s               & 0.45 s          \\ 
O-ada  & 154.08 s            & 33.76 s                   & 17.35 s             & 16.67 s         & 35.57 s              & 8.9 s           \\ 
O-small& 166.73 s            & 34.16                     & 18.32               & 16.10           & 34.09 s              & 9.0 s           \\ 
O-large& 206.07 s            & 41.78 s                   & 20.58 s             & 29.97 s         & 44.20 s              & 10.72 s         \\ 
Llama  & 545.51 s            & 129.16 s                  & 38.15 s             & 28.93 s         & 71.49 s              & 18.02 s         \\
stella & 99.75 s             & 19.38 s                   & 12.41 s             & 9.95 s          & 20.18 s              & 5.72 s          \\
Qwen   & 745.85 s            & 129.16 s                  & 58.19 s             & 40.29 s         & 183.24 s             & 20.49 s         \\
\hline
\end{tabular}
% }
\end{table*}

\textbf{Email-Spam}~\footnote{\url{https://huggingface.co/datasets/kendx/NLP-ADBench/tree/main/datasets/email_spam}} \cite{metsis2006spam} contains 5,171 emails labeled as spam or ham, with spam treated as the anomaly class. We utilized the preprocessed version provided in \cite{li2024nlp}.

\textbf{SMS-Spam} \footnote{\url{https://archive.ics.uci.edu/dataset/228/sms+spam+collection}} \cite{almeida2011contributions} comprises 5,574 SMS messages originally labeled as spam or ham. Spam messages are designated as the anomaly.

\textbf{COVID-Fake} \footnote{\url{https://github.com/diptamath/covid_fake_news?tab=readme-ov-file}} \cite{das2021heuristic} comprises posts collected from social media platforms and fact-checking websites. Real news items were sourced from verified outlets providing accurate COVID-19 information, while fake news was gathered from tweets, posts, and articles containing misinformation about COVID-19. Fake news is treated as the anomaly class.

\textbf{LIAR2} \footnote{\url{https://github.com/chengxuphd/liar2?tab=readme-ov-file}} \cite{xu2024enhanced} consists of approximately 23,000 statements manually labeled by professional fact-checkers for fake news detection tasks. The "True" class, representing accurate statements, is considered the normal class, while the "Pants on Fire" class, representing highly misleading statements, is treated as the anomaly.

\textbf{OLID} \footnote{\url{https://sites.google.com/site/offensevalsharedtask/olid}} \cite{zampieri-etal-2019-predicting}~\cite{zampierietal2019} contains 14,200 annotated English tweets, categorized using a three-level annotation model. For this benchmark, only the Level A (Offensive Language Detection) annotations are used, where tweets labeled as offensive are considered as anomalies, and non-offensive tweets are considered as normal.

\textbf{Hate-Speech} \footnote{\url{https://github.com/t-davidson/hate-speech-and-offensive-language}} \cite{davidson2017automated} contains tweets annotated by CrowdFlower users. The tweet content is used as data, with "hate speech" treated as anomalies.

\section{Evaluation with Hyperparameter Search}

Experiments with hyperparameter search reveal significant insights into the performance dynamics of anomaly detection algorithms when paired with various embedding models. The hyperparameters for LOF and HBOS are searched in $\{5, 10, 20, 40\}$. For iForest and iNNE, sample points are searched in $\{2, 4, 8, 16, 32, 64, 128, 256\}$ and with defalt $t=200$. $\gamma$ of OCSVM are searched in $\{0.05, 0.1, 0.5, 1\}$ and with RBF kernel. Both COPOD and ECOD don't have hyperparameters.

When comparing Table~\ref{tab:hyper} with the default parameters results in Table 2, we observe that kNN emerges as a particularly strong performer across multiple embedding models, especially with the OpenAI family of embeddings. This suggests that distance-based approaches effectively leverage the semantic information captured by these models, particularly in tasks like spam and fake news detection. 
% The consistent performance of kNN across diverse datasets indicates its versatility in handling different anomaly types and distributions.

INNE demonstrates the most balanced and robust performance profile when considering average scores across all datasets and embedding combinations. Its isolation-based approach with hypersphere partitioning appears particularly well-suited to the complex topological structure of embedding spaces, allowing it to identify local anomalies that other methods might miss. The performance improvement of INNE after hyperparameter optimization is especially notable with embedding models like Llama and stella, suggesting a strong complementarity between isolation-based algorithms and these embedding architectures.

% LOF exhibit dramatic performance improvements after optimization, particularly evident in the COVID-Fake and LIAR2 datasets. This substantial shift from their poor default performance highlights how critical parameter tuning can be for certain algorithm classes. In contrast, statistical methods such as ECOD and COPOD show relatively minimal gains from parameter optimization, indicating their inherent stability across different configurations. OCSVM presents perhaps the most inconsistent pattern, with performance degradation in several embedding-dataset combinations after optimization, pointing to fundamental challenges in applying boundary-based methods to the high-dimensional spaces characteristic of text embeddings.

\begin{table*}[]
\caption{Evaluation across 6 datasets in terms of AU-ROC under parameter search.}
\label{tab:hyper}
\resizebox{\textwidth}{0.5\textheight}{%
\begin{tabular}{ll|llllll|l}
\hline
\textbf{Embeddings}               & \textbf{Detectors} & \textbf{Email-Spam} & \textbf{SMS-Spam} & \textbf{COVID-Fake} & \textbf{LIAR2}  & \textbf{Hate-Speech} & \textbf{OLID}   & \textbf{Average} \\ \hline
\multirow{8}{*}{\textbf{BERT}}    & \textbf{kNN}       & 0.7707              & 0.5663            & 0.8467              & 0.6594          & 0.5033               & 0.5164          & 0.6438           \\
                                  & \textbf{OCSVM}     & 0.5862              & 0.5322            & 0.4868              & 0.5059          & \textbf{0.5154}      & 0.4982          & 0.5208           \\
                                  & \textbf{IForest}   & 0.7187              & 0.6307            & 0.7794              & 0.6178          & 0.4957               & 0.5315          & 0.6290           \\
                                  & \textbf{LOF}       & 0.7014              & 0.3531            & \textbf{0.8707}     & \textbf{0.676}  & 0.4636               & 0.5018          & 0.5944           \\
                                  & \textbf{ECOD}      & 0.7309              & 0.6235            & 0.7722              & 0.6175          & 0.4889               & 0.4933          & 0.6211           \\
                                  & \textbf{INNE}      & \textbf{0.7748}     & \textbf{0.6551}   & 0.8381              & 0.6572          & 0.4897               & 0.5078          & \textbf{0.6538}  \\
                                  & \textbf{HBOS}      & 0.717               & 0.6252            & 0.7711              & 0.6209          & 0.4935               & 0.5065          & 0.6224           \\
                                  & \textbf{COPOD}     & 0.6454              & 0.5929            & 0.7714              & 0.6242          & 0.4971               & \textbf{0.5189} & 0.6083           \\ \hline
\multirow{8}{*}{\textbf{MINILM}}  & \textbf{kNN}       & \cellcolor[gray]{0.9}\textbf{0.9669}     & 0.3537            & \textbf{0.8413}     & \textbf{0.7249} & \textbf{0.5804}      & 0.5063          & 0.6623           \\
                                  & \textbf{OCSVM}     & 0.6929              & 0.582             & 0.5241              & 0.5549          & 0.5538               & \textbf{0.5988} & 0.5844           \\
                                  & \textbf{IForest}   & 0.9112              & 0.5655            & 0.7373              & 0.5982          & 0.4582               & 0.4872          & 0.6263           \\
                                  & \textbf{LOF}       & 0.6896              & 0.5507            & 0.8095              & 0.6759          & 0.5286               & 0.5689          & 0.6372           \\
                                  & \textbf{ECOD}      & 0.9525              & 0.5934            & 0.7581              & 0.6532          & 0.3786               & 0.4208          & 0.6261           \\
                                  & \textbf{INNE}      & 0.9603              & \textbf{0.6531}   & 0.8226              & 0.6754          & 0.5019               & 0.5381          & \textbf{0.6919}  \\
                                  & \textbf{HBOS}      & 0.9489              & 0.6177            & 0.7447              & 0.6614          & 0.3937               & 0.4333          & 0.6333           \\
                                  & \textbf{COPOD}     & 0.9453              & 0.6317            & 0.7416              & 0.6695          & 0.371                & 0.4037          & 0.6271           \\ \hline
\multirow{8}{*}{\textbf{O-ada}}   & \textbf{kNN}       & \textbf{0.9506}     & 0.3535            & \textbf{0.9094}     & \cellcolor[gray]{0.9}\textbf{0.7921} & \textbf{0.6341}      & 0.5243          & 0.6940           \\
                                  & \textbf{OCSVM}     & 0.7535              & 0.5633            & 0.56                & 0.4513          & 0.5049               & 0.6474          & 0.5801           \\
                                  & \textbf{IForest}   & 0.8826              & 0.7168            & 0.7493              & 0.6268          & 0.5137               & 0.5194          & 0.6681           \\
                                  & \textbf{LOF}       & 0.4203              & 0.536             & 0.798               & 0.7568          & 0.4683               & \textbf{0.5368} & 0.5860           \\
                                  & \textbf{ECOD}      & 0.938               & 0.8822            & 0.815               & 0.72            & 0.461                & 0.4986          & 0.7191           \\
                                  & \textbf{INNE}      & 0.9469              & \cellcolor[gray]{0.9}\textbf{0.8905}   & 0.8633              & 0.7558          & 0.5236               & 0.5319          & \cellcolor[gray]{0.9}\textbf{0.7520}  \\
                                  & \textbf{HBOS}      & 0.9438              & 0.8851            & 0.8194              & 0.7212          & 0.4692               & 0.5124          & 0.7252           \\
                                  & \textbf{COPOD}     & 0.9502              & 0.8759            & 0.8153              & 0.7201          & 0.4513               & 0.4811          & 0.7157           \\ \hline
\multirow{8}{*}{\textbf{O-small}} & \textbf{kNN}       & \textbf{0.9637}     & 0.229             & \textbf{0.94}       & \textbf{0.7872} & \cellcolor[gray]{0.9}\textbf{0.6416}      & 0.5587          & 0.6867           \\
                                  & \textbf{OCSVM}     & 0.5716              & 0.5253            & 0.5165              & 0.4664          & 0.5505               & \cellcolor[gray]{0.9}\textbf{0.6059} & 0.5394           \\
                                  & \textbf{IForest}   & 0.9021              & 0.6007            & 0.7871              & 0.6086          & 0.4805               & 0.5388          & 0.6530           \\
                                  & \textbf{LOF}       & 0.4642              & 0.5215            & 0.8596              & 0.7491          & 0.4509               & 0.5602          & 0.6009           \\
                                  & \textbf{ECOD}      & 0.9481              & 0.6301            & 0.8808              & 0.7022          & 0.4249               & 0.5295          & 0.6859           \\
                                  & \textbf{INNE}      & 0.9572              & \textbf{0.6471}   & 0.9295              & 0.7374          & 0.5234               & 0.5884          & \textbf{0.7305}  \\
                                  & \textbf{HBOS}      & 0.953               & 0.6284            & 0.8738              & 0.7009          & 0.4245               & 0.5301          & 0.6851           \\
                                  & \textbf{COPOD}     & 0.9605              & 0.5722            & 0.8664              & 0.6974          & 0.4017               & 0.4963          & 0.6658           \\ \hline
\multirow{8}{*}{\textbf{O-large}} & \textbf{kNN}       & 0.9121              & 0.1698            & \cellcolor[gray]{0.9}\textbf{0.9537}     & \textbf{0.7687} & \textbf{0.6291}      & \textbf{0.5497} & 0.6639           \\
                                  & \textbf{OCSVM}     & 0.7794              & 0.5353            & 0.5124              & 0.4302          & 0.5086               & 0.5938          & 0.5600           \\
                                  & \textbf{IForest}   & 0.9142              & 0.5603            & 0.8089              & 0.5887          & 0.507                & 0.554           & 0.6555           \\
                                  & \textbf{LOF}       & 0.5009              & 0.4748            & 0.8736              & 0.7435          & 0.4296               & 0.5429          & 0.5942           \\
                                  & \textbf{ECOD}      & 0.9487              & 0.6422            & 0.8875              & 0.654           & 0.3959               & 0.4967          & 0.6708           \\
                                  & \textbf{INNE}      & 0.9537              & 0.6765            & 0.9295              & 0.7125          & 0.4899               & 0.519           & \textbf{0.7135}  \\
                                  & \textbf{HBOS}      & 0.9552              & 0.6562            & 0.8866              & 0.6458          & 0.3882               & 0.5108          & 0.6738           \\
                                  & \textbf{COPOD}     & \textbf{0.9639}     & \textbf{0.6798}   & 0.8854              & 0.6536          & 0.3537               & 0.498           & 0.6724           \\ \hline
\multirow{8}{*}{\textbf{Llama}}   & \textbf{kNN}       & \textbf{0.9323}     & 0.5389            & 0.8668              & 0.7241          & 0.4991               & 0.4081          & 0.6616           \\
                                  & \textbf{OCSVM}     & 0.8929              & 0.5431            & 0.5785              & 0.5045          & \textbf{0.5224}      & \textbf{0.529}  & 0.5951           \\
                                  & \textbf{IForest}   & 0.8971              & 0.7509            & 0.7815              & 0.6833          & 0.4691               & 0.4267          & 0.6681           \\
                                  & \textbf{LOF}       & 0.7188              & 0.461             & \textbf{0.8704}     & 0.7286          & 0.4564               & 0.5093          & 0.6241           \\
                                  & \textbf{ECOD}      & 0.8844              & 0.7573            & 0.7819              & 0.6989          & 0.4643               & 0.3998          & 0.6644           \\
                                  & \textbf{INNE}      & 0.918               & 0.8083            & 0.8666              & 0.7074          & 0.4778               & 0.4929          & \textbf{0.7118}  \\
                                  & \textbf{HBOS}      & 0.9029              & 0.7908            & 0.7782              & 0.7064          & 0.4606               & 0.3961          & 0.6725           \\
                                  & \textbf{COPOD}     & 0.9153              & \textbf{0.8163}   & 0.7584              & \textbf{0.7291} & 0.4435               & 0.3526          & 0.6692           \\ \hline
\multirow{8}{*}{\textbf{stella}}  & \textbf{kNN}       & \textbf{0.935}      & 0.3451            & \textbf{0.9034}     & \textbf{0.6884} & 0.4746               & 0.5016          & 0.6414           \\
                                  & \textbf{OCSVM}     & 0.8655              & 0.5818            & 0.5701              & 0.5071          & \textbf{0.5516}      & 0.5084          & 0.5974           \\
                                  & \textbf{IForest}   & 0.8884              & 0.7398            & 0.7927              & 0.517           & 0.4284               & 0.4716          & 0.6397           \\
                                  & \textbf{LOF}       & 0.4722              & 0.5159            & 0.7749              & 0.6575          & 0.4341               & \textbf{0.5598} & 0.5691           \\
                                  & \textbf{ECOD}      & 0.9075              & 0.7894            & 0.8115              & 0.5023          & 0.3421               & 0.4395          & 0.6321           \\
                                  & \textbf{INNE}      & 0.927               & 0.8139            & 0.8466              & 0.6266          & 0.4118               & 0.5243          & \textbf{0.6917}  \\
                                  & \textbf{HBOS}      & 0.9178              & 0.8038            & 0.8161              & 0.4952          & 0.3391               & 0.4276          & 0.6333           \\
                                  & \textbf{COPOD}     & 0.93                & \textbf{0.8589}   & 0.8167              & 0.4936          & 0.3018               & 0.3797          & 0.6301           \\ \hline
\multirow{8}{*}{\textbf{Qwen}}    & \textbf{kNN}       & \textbf{0.9171}     & 0.2671            & 0.8438              & 0.6626          & 0.4991               & 0.4602          & 0.6083           \\
                                  & \textbf{OCSVM}     & 0.8304              & 0.5431            & 0.5279              & 0.5105          & \textbf{0.5224}      & \textbf{0.5712} & 0.5843           \\
                                  & \textbf{IForest}   & 0.8758              & 0.6552            & 0.7598              & 0.6169          & 0.4691               & 0.4948          & 0.6453           \\
                                  & \textbf{LOF}       & 0.6915              & 0.4299            & 0.8575              & \textbf{0.6972} & 0.4564               & 0.4894          & 0.6037           \\
                                  & \textbf{ECOD}      & 0.8678              & 0.6648            & 0.768               & 0.6172          & 0.4643               & 0.4773          & 0.6432           \\
                                  & \textbf{INNE}      & 0.8989              & 0.7057            & \textbf{0.8639}     & 0.6733          & 0.4778               & 0.518           & \textbf{0.6896}  \\
                                  & \textbf{HBOS}      & 0.8877              & 0.6886            & 0.7644              & 0.6187          & 0.4606               & 0.4741          & 0.6490           \\
                                  & \textbf{COPOD}     & 0.9044              & \textbf{0.7393}   & 0.7463              & 0.6291          & 0.4435               & 0.4336          & 0.6494           \\ \hline
\end{tabular}%
}
\end{table*}

\section{Embedding Models}

\begin{table}[!htbp]
    \caption{Embedding Models Overview. M and B are for million and billion, respectively.}
    \label{tab:embed}
    \centering
    \resizebox{0.5\textwidth}{!}{
    \begin{tabular}{l|ccc}
    \hline
     Models         & Max Tokens & \# Dimensions  & \# Parameters\\
     \hline
     BERT           & 512        & 768            & 110 M       \\
     MINILM         & 512        & 384            & 22.7 M      \\
     O-ada     & 8191       & 1536           & -       \\
     O-small   & 8191       & 1536           & -       \\
     O-large   & 8191       & 3072           & -       \\
     LLAMA      & 4096       & 2048           & 1.24 B         \\
     stella         & 2048       & 1024           & 435 M           \\
     Qwen        & 8192       & 1536           & 1.54 B       \\
    \hline
    \end{tabular}
    }
\end{table}

To effectively represent textual data, we use various pre-trained embedding models that transform text into dense vector representations. Table~\ref{tab:embed} summarizes the embedding models employed in this paper. These embeddings serve as feature inputs for anomaly detection models, enabling them to capture semantic similarities and deviations in text. We selected a diverse set of embedding models, balancing between model size, token length limits, and computational efficiency. The models used in this study are:

\begin{itemize}
    \item BERT~\footnote{\url{https://huggingface.co/google-bert/bert-base-uncased}} (\emph{bert-base-uncased})
    \item MINILM~\footnote{\url{https://huggingface.co/sentence-transformers/all-MiniLM-L6-v2}} (\emph{all-MiniLM-L6-v2})
    \item O-ada~\footnote{\url{https://platform.openai.com/docs/guides/embeddings/}} (\emph{text-embedding-ada-002})
    \item O-small~\footnote{\url{https://platform.openai.com/docs/guides/embeddings/}} (\emph{text-embedding-3-small})
    \item O-large~\footnote{\url{https://platform.openai.com/docs/guides/embeddings/}} (\emph{text-embedding-3-large})
    \item LLAMA~\footnote{\url{https://huggingface.co/meta-llama/Llama-3.2-1B}} (\emph{Llama-3.2-1B})
    \item stella~\footnote{\url{https://huggingface.co/NovaSearch/stella\_en\_400M\_v5}} (\emph{stella\_en\_400M\_v5})
    \item Qwen~\footnote{\url{https://huggingface.co/Qwen/Qwen2.5-1.5B}} (\emph{Qwen2.5-1.5B})
\end{itemize}

Beyond model size and token limits, computational efficiency is a key factor in selecting embedding models, particularly for real-world applications where inference speed is critical. Table~\ref{time} presents the embedding time (in seconds) required to process six datasets using each embedding model.

From the Table~\ref{time}, we observe a significant variation in embedding extraction time. MINILM is the fastest across all datasets, taking only a few seconds, making it ideal for applications requiring real-time embedding generation. BERT offers a moderate trade-off, with embedding times significantly lower than larger models but higher than MINILM. OpenAI’s embeddings (O-ada, O-small, O-large) are relatively slow, likely due to their high-dimensional output and extended token support. Llama and Qwen models require the most computation, with Qwen taking up to 745.85 seconds on the Email-Spam dataset, reflecting the high computational cost of large autoregressive models.

%=======================================================
\section{Comparative Analysis of Anomaly Detection Algorithms}

Anomaly detection algorithms vary in their underlying assumptions, computational efficiency, and effectiveness across different types of data distributions. In this section, we provide a comparative analysis of the eight anomaly detection methods used in this study: kNN, OCSVM, iForest, LOF, HBOS, ECOD, INNE and COPOD. 

Distance-based methods, such as kNN, define anomalies based on their relative distance to surrounding points. kNN anomaly detection computes the distance between a data point and its $k$th nearest neighbor, with larger distances indicating potential anomalies. This method is conceptually simple and effective in low-dimensional spaces with clear separation between normal and anomalous points. However, its primary drawback is the curse of dimensionality, where distance metrics lose discriminative power as dimensionality increases. Additionally, kNN is computationally expensive, with a worst-case complexity of $O(n^2)$, making it impractical for large datasets without optimizations such as approximate nearest neighbor search.

Density-based approaches assume that anomalies reside in low-density regions relative to normal points. LOF estimates the local density of a point by comparing it with the densities of its neighbors. It is highly effective in detecting anomalies in datasets with non-uniform density distributions, where global models may fail. However, LOF is computationally expensive complexity $O(n^2)$ in the worst case and sensitive to the choice of neighborhood size, requiring careful hyperparameter tuning.

A more efficient density estimation approach is HBOS, which models feature distributions independently using histograms. This makes it computationally extremely fast $O(n)$ and scalable to large datasets. However, HBOS assumes feature independence, limiting its effectiveness when strong feature correlations exist. In such cases, its effectiveness diminishes as it fails to capture intricate relationships between features, potentially leading to suboptimal anomaly detection performance.

Isolation-based approaches, such as iForest, take a different perspective by recursively partitioning the feature space. Since anomalies are typically isolated with fewer splits, iForest identifies them based on the depth required to isolate each point. iForest is computationally efficient $O(nlogn)$ and performs well in high-dimensional spaces compared to distance-based methods, but it is struggle with local anomalies. An extension of iForest, INNE, replaces axis-aligned splits with hypersphere-based partitions. This enhances robustness in detecting anomalies in complex distributions, particularly local anomalies. 

Statistical approaches model the underlying distribution of data and identify anomalies as points that significantly deviate from expected behavior. ECOD estimates anomaly scores based on the empirical cumulative distribution function (ECDF) for each feature independently. It is parameter-free and computationally efficient $O(n)$, making it highly scalable. However, like HBOS, ECOD assumes feature independence, which can limit its effectiveness in multivariate settings. COPOD improves upon ECOD by leveraging copula functions to model dependencies between features, making it more effective for detecting anomalies in correlated data. However, this comes at the cost of increased computational complexity, making COPOD less scalable for very large datasets.

%=================================================================

\begin{table*}[!htbp]
	\centering
	\caption{t-SNE visualization of embeddings from 8 models across 6 datasets. Blue points represent normal instances, while red points denote anomalies.}
    \resizebox{\textwidth}{!}{
	\begin{tabular}{ccccccc}
		\toprule
		\textbf{Embeddings}   & \textbf{Email-Spam} & \textbf{SMS-Spam}         & \textbf{COVID-Fake} & \textbf{LIAR2}  & \textbf{Hate-Speech} & \textbf{OLID}                    \\
		\midrule
		\textbf{BERT}       & \includegraphics[width=0.15\textwidth]{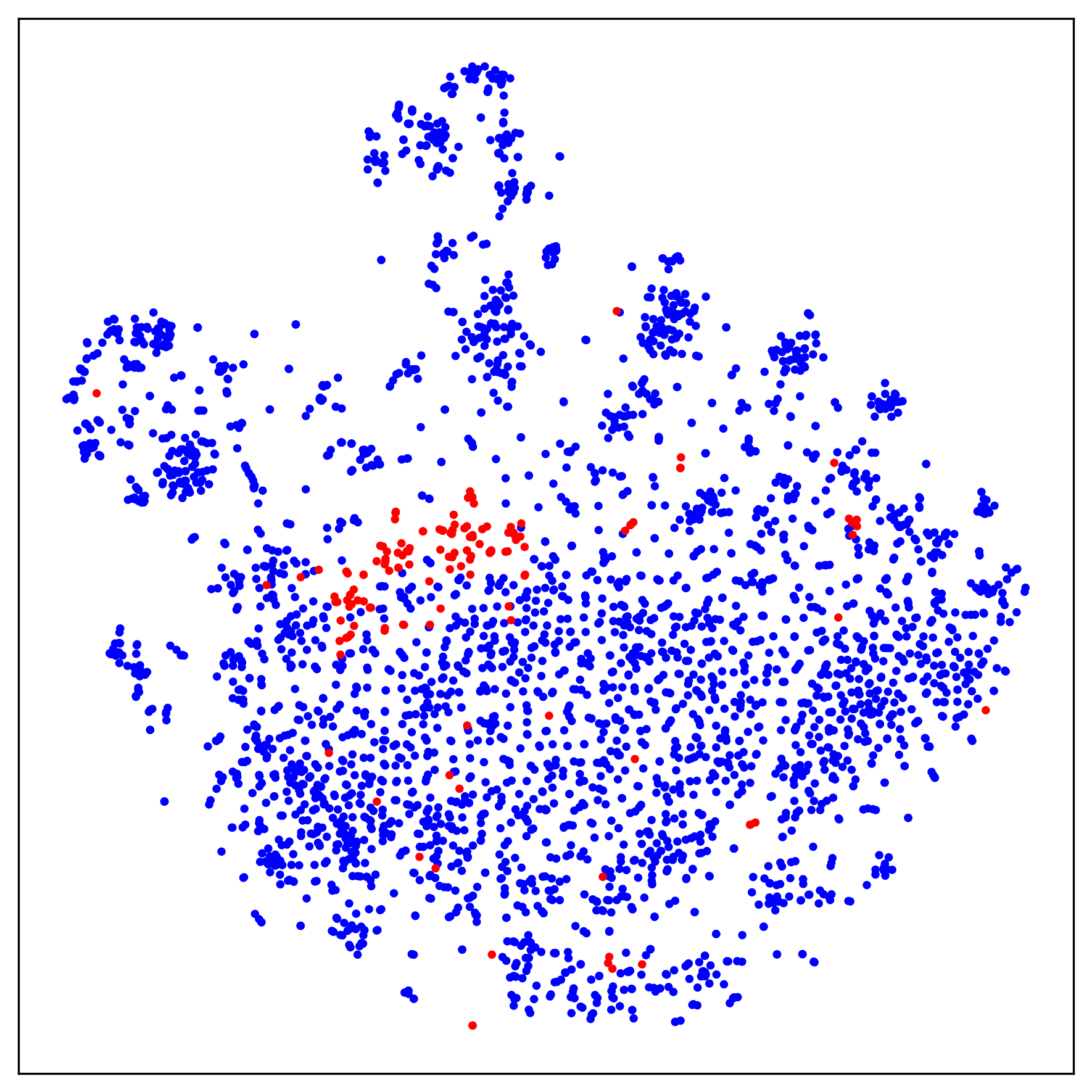} &
	\includegraphics[width=0.15\textwidth]{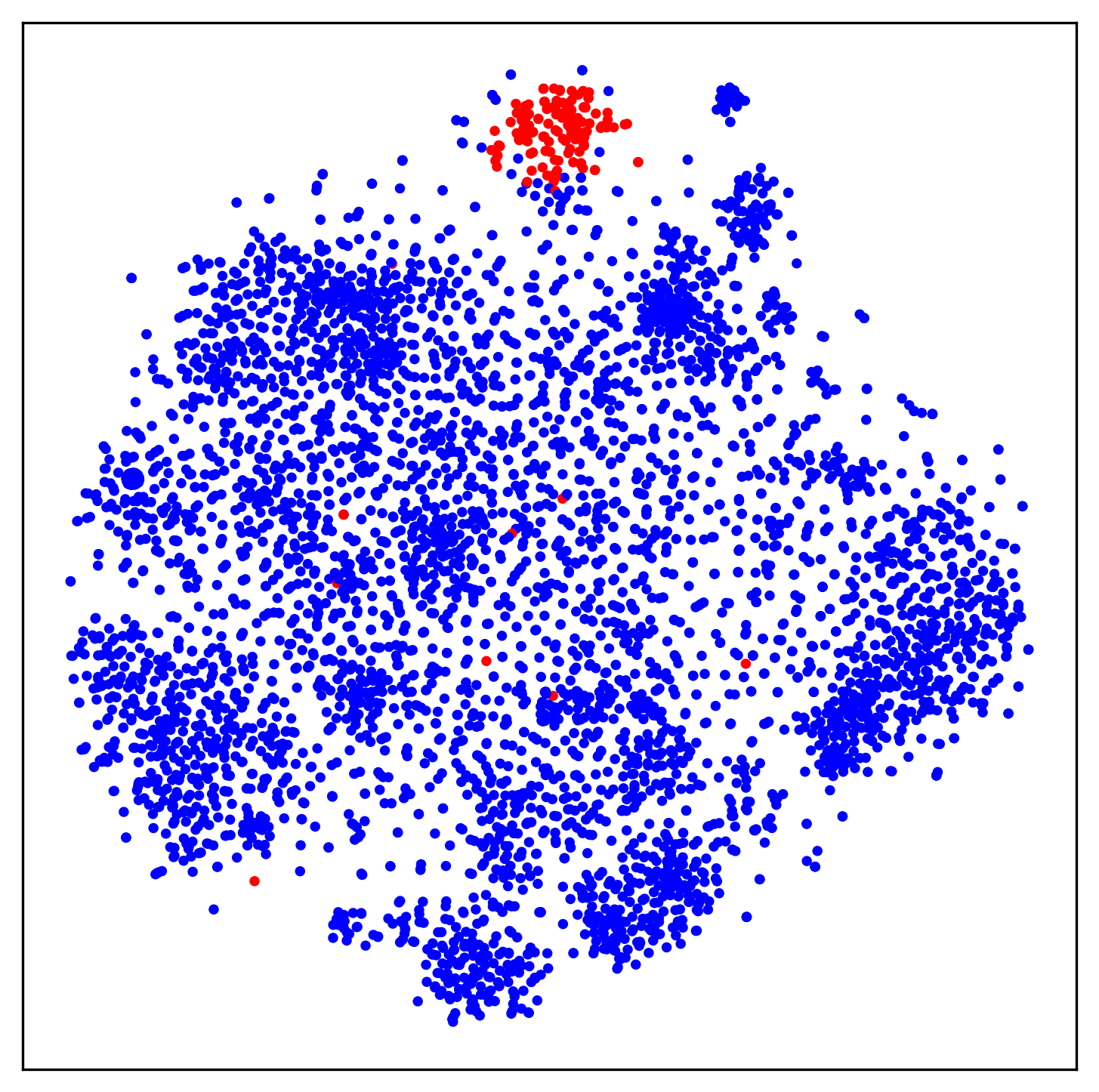} &
		\includegraphics[width=0.15\textwidth]{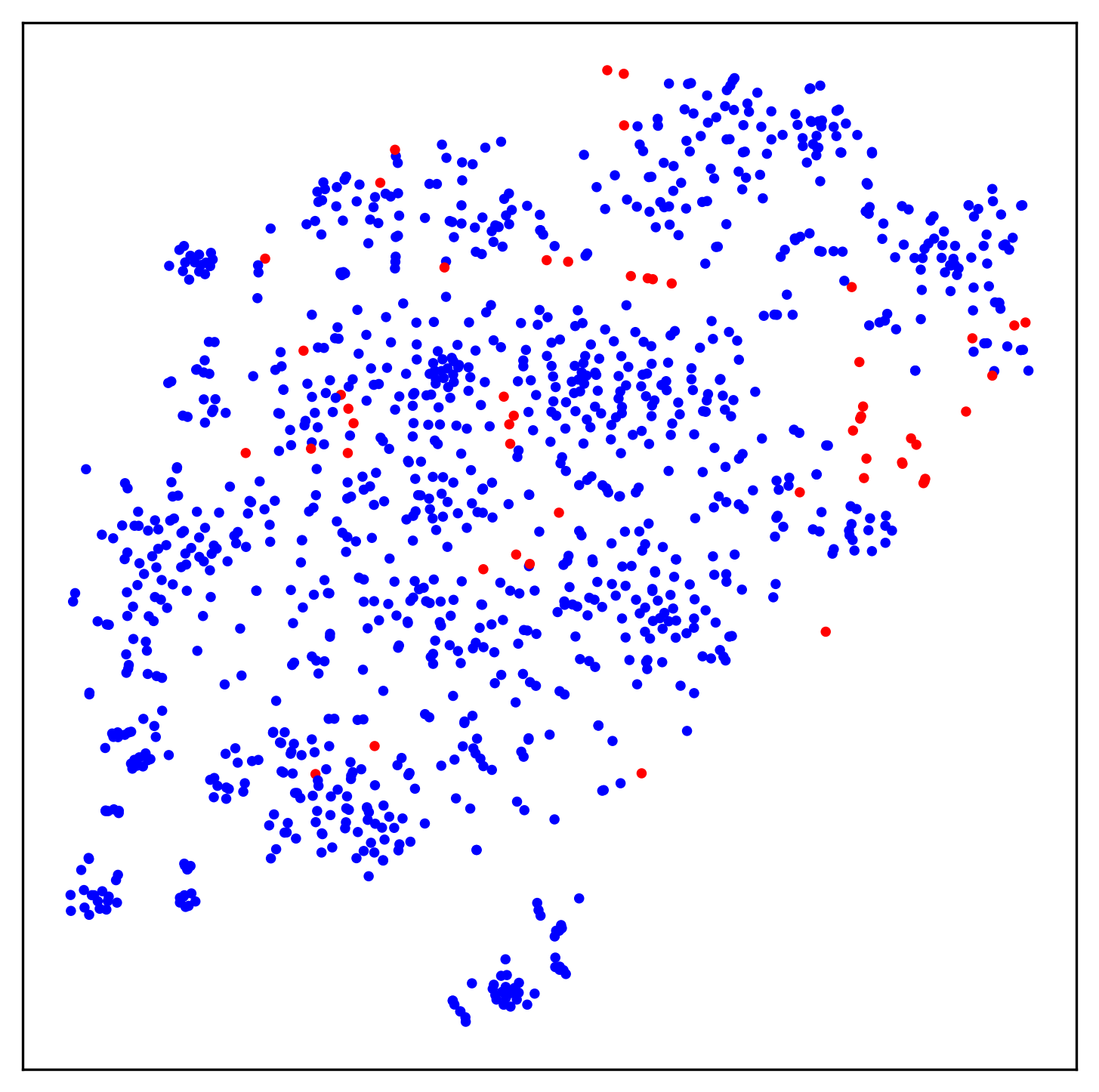} &
        \includegraphics[width=0.15\textwidth]{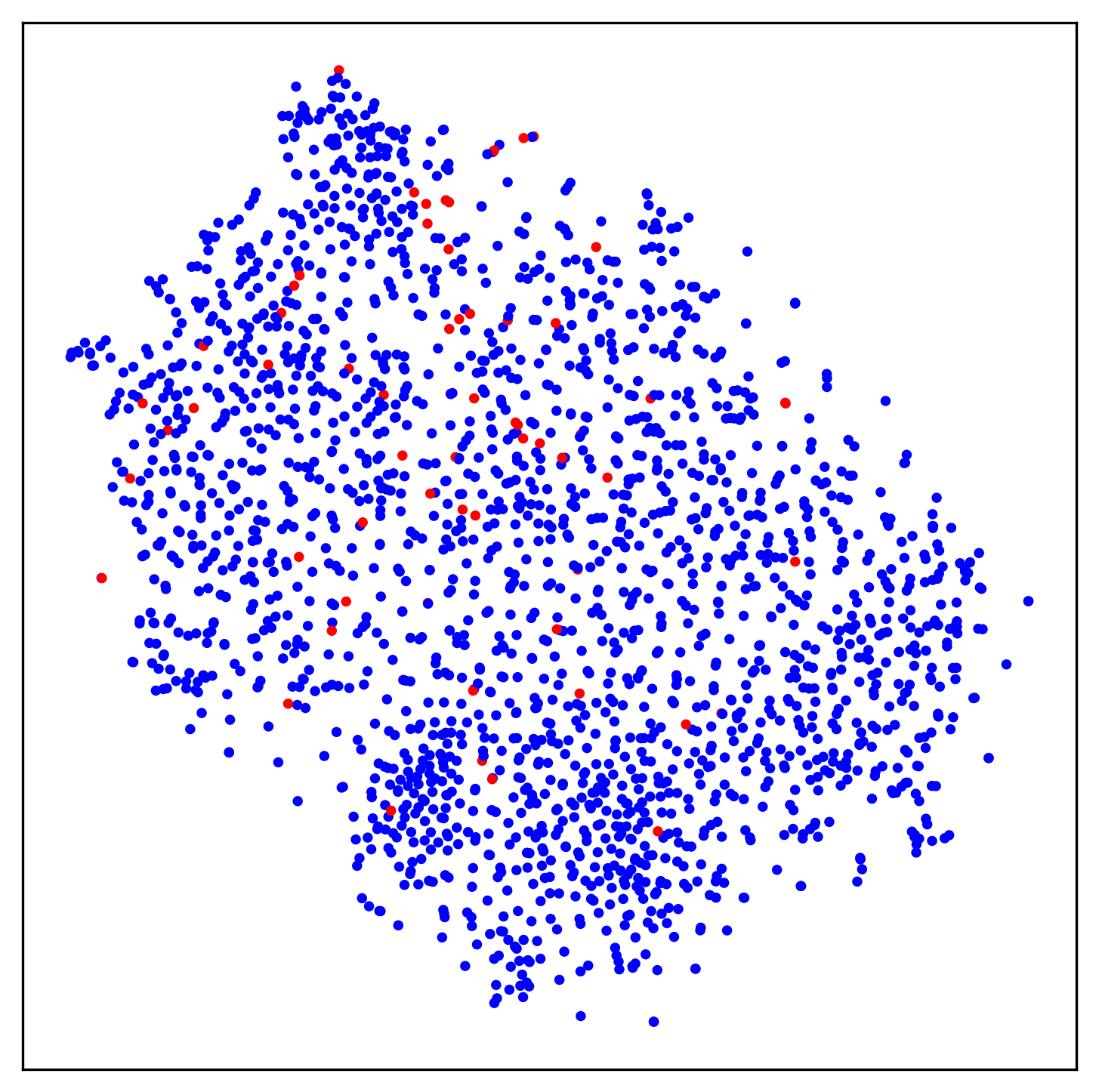} &
        \includegraphics[width=0.15\textwidth]{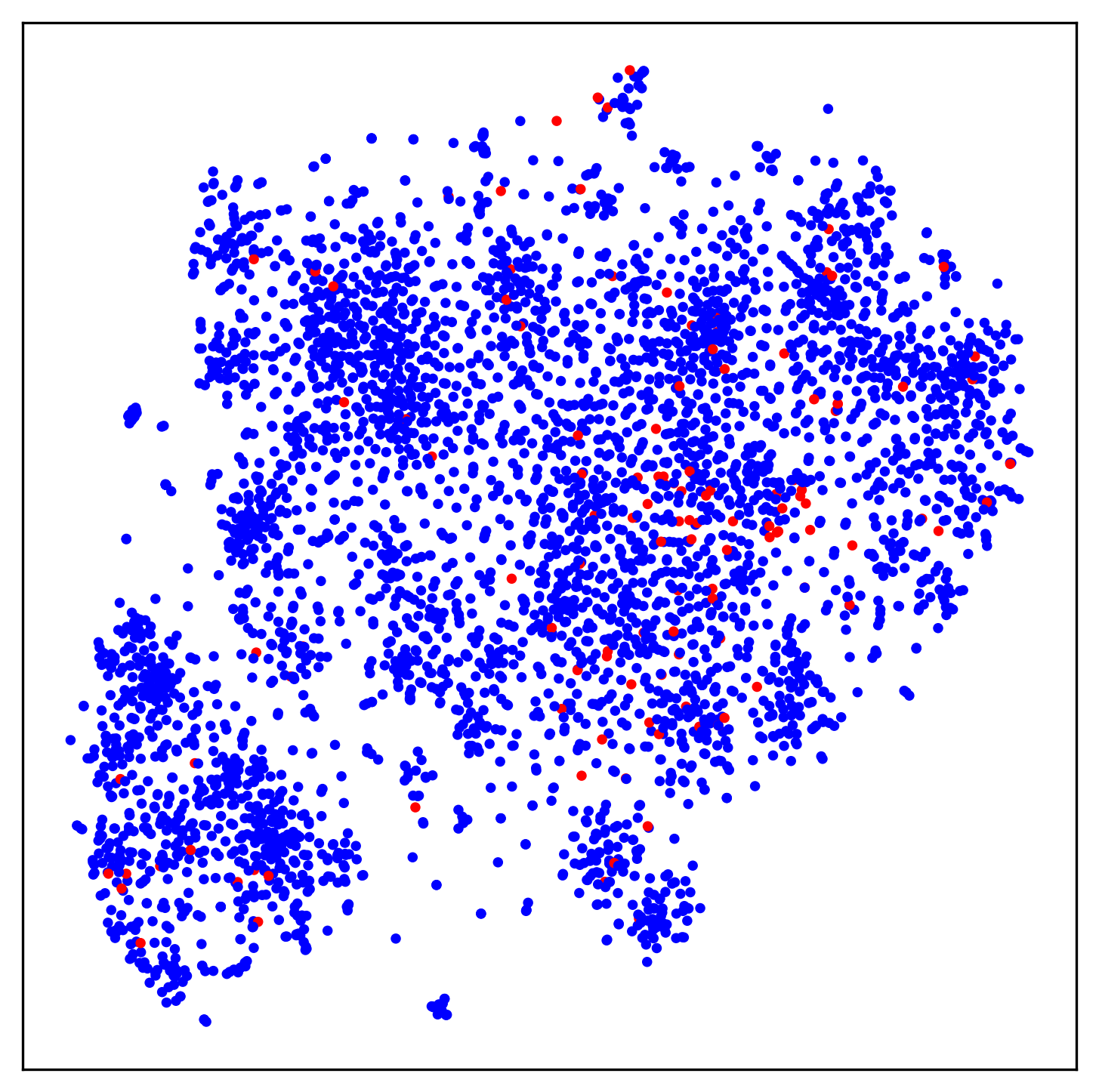} &
        \includegraphics[width=0.15\textwidth]{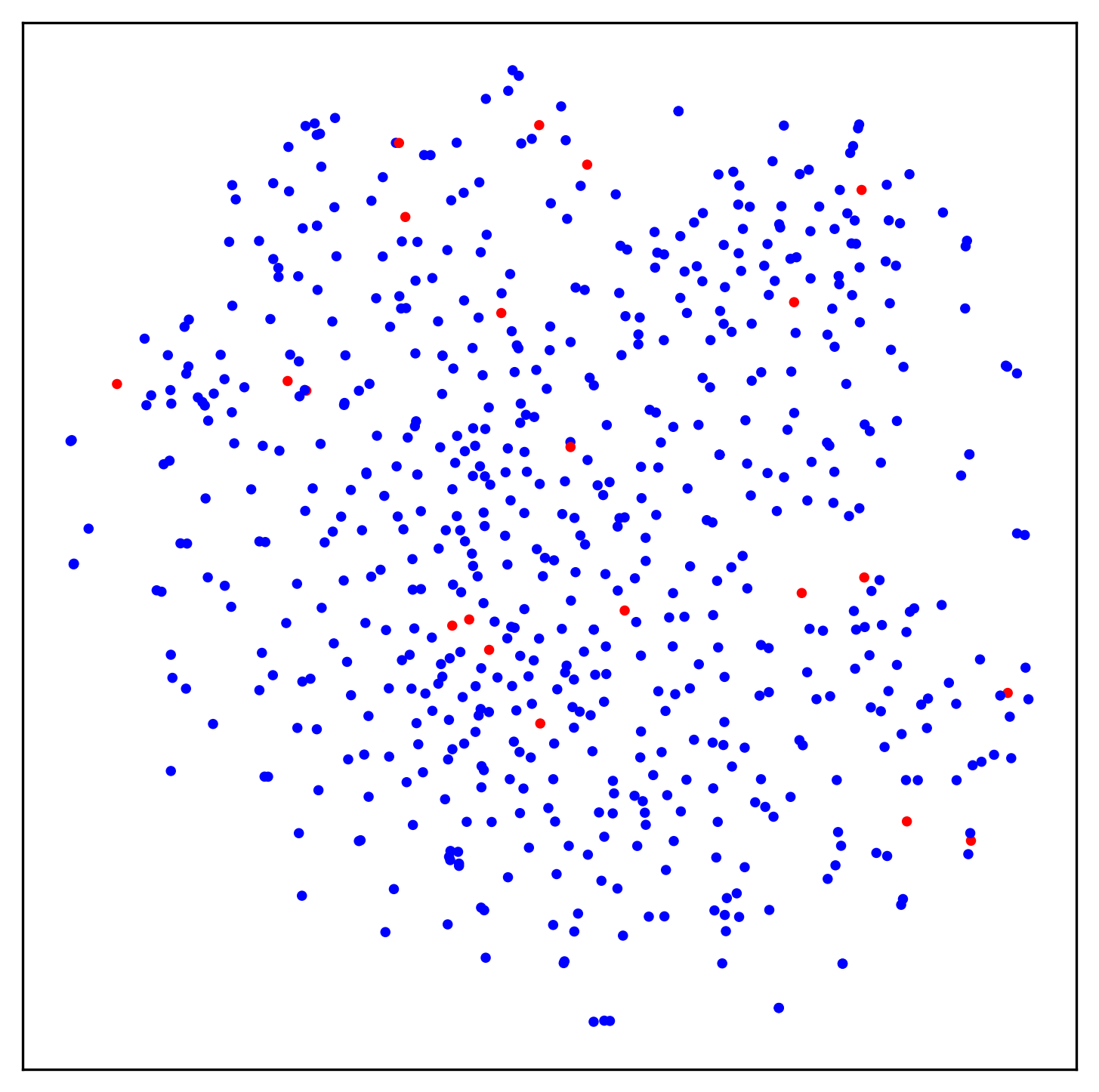}\\
        \midrule
		\textbf{MINILM}       & \includegraphics[width=0.15\textwidth]{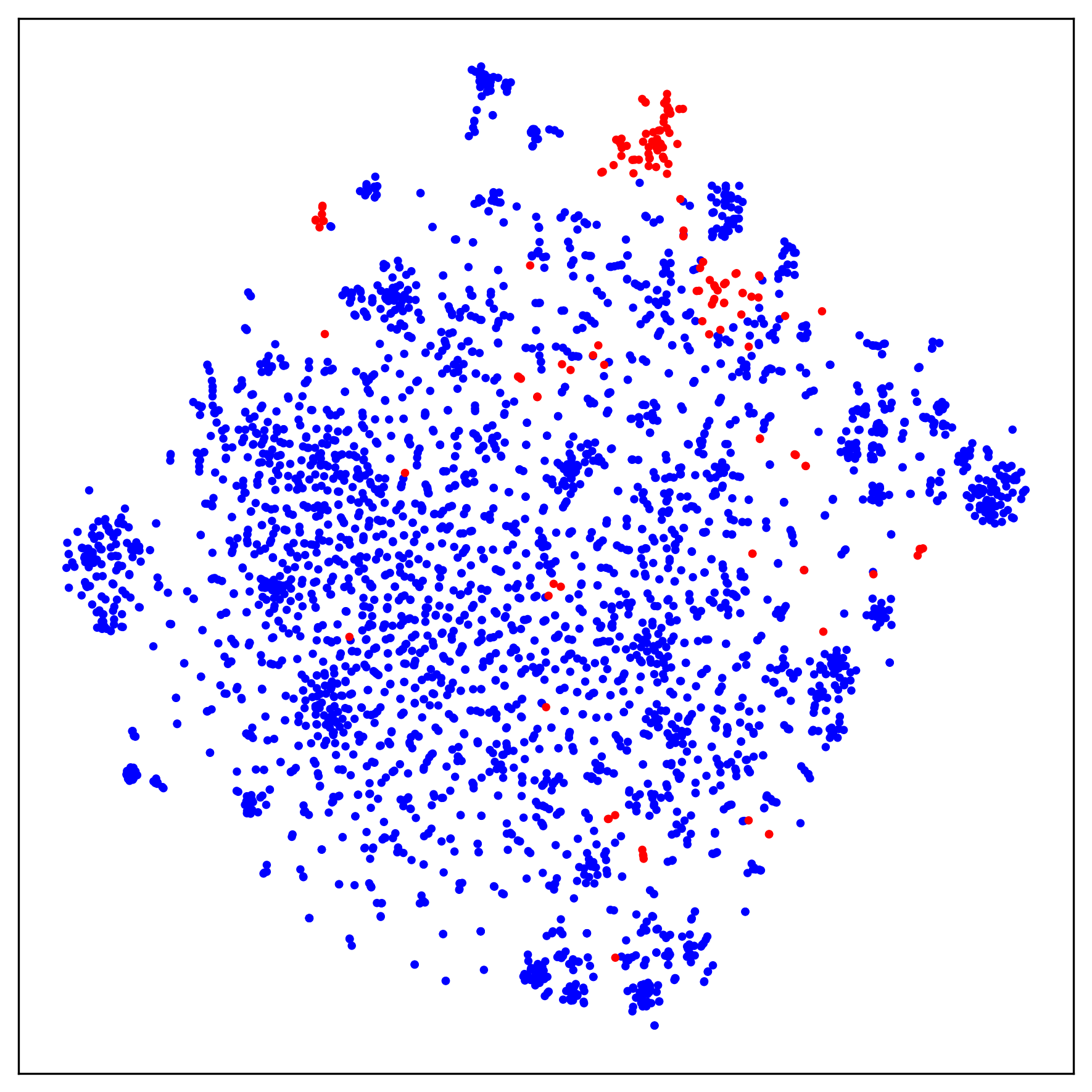} &
	\includegraphics[width=0.15\textwidth]{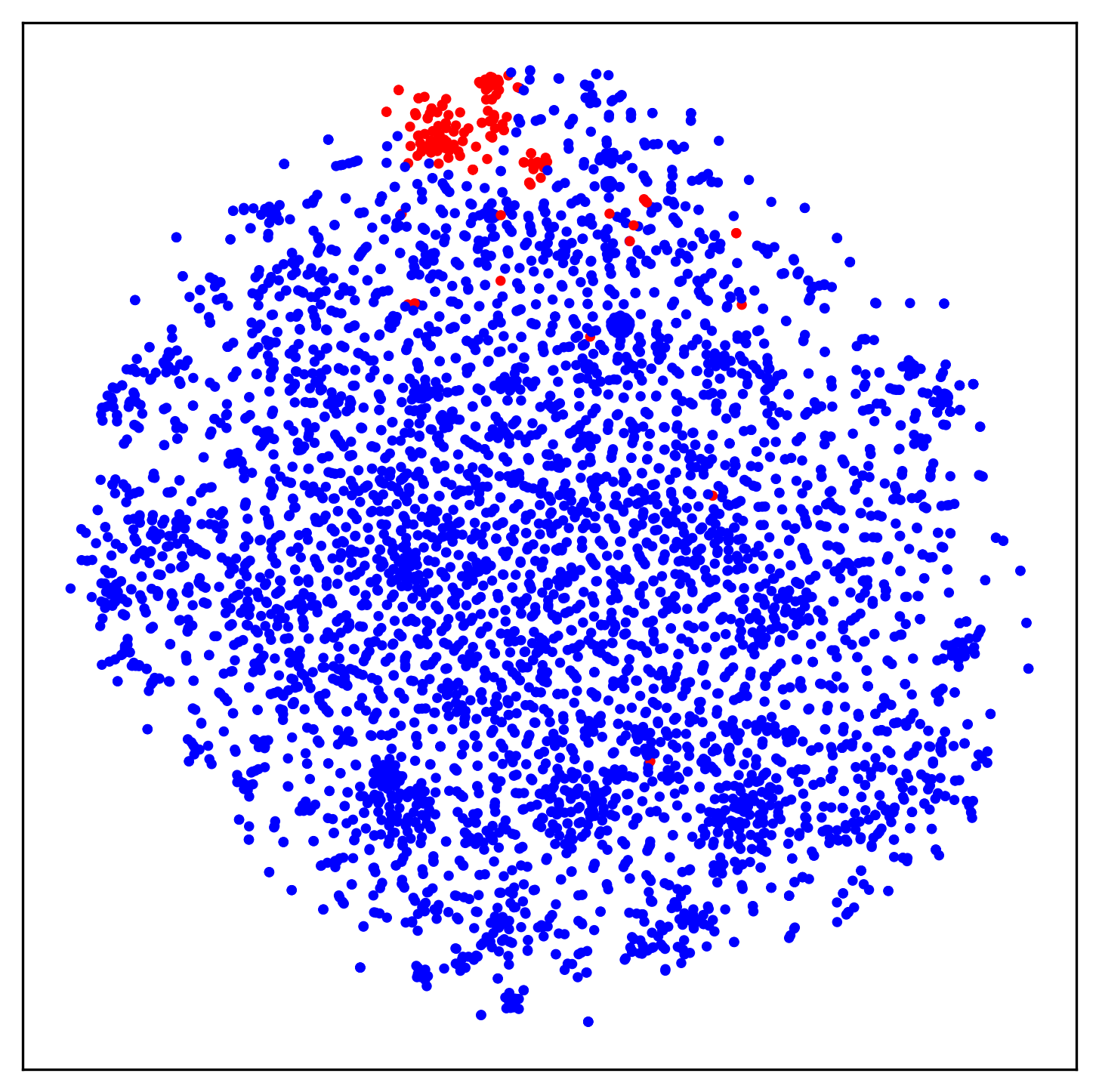} &
		\includegraphics[width=0.15\textwidth]{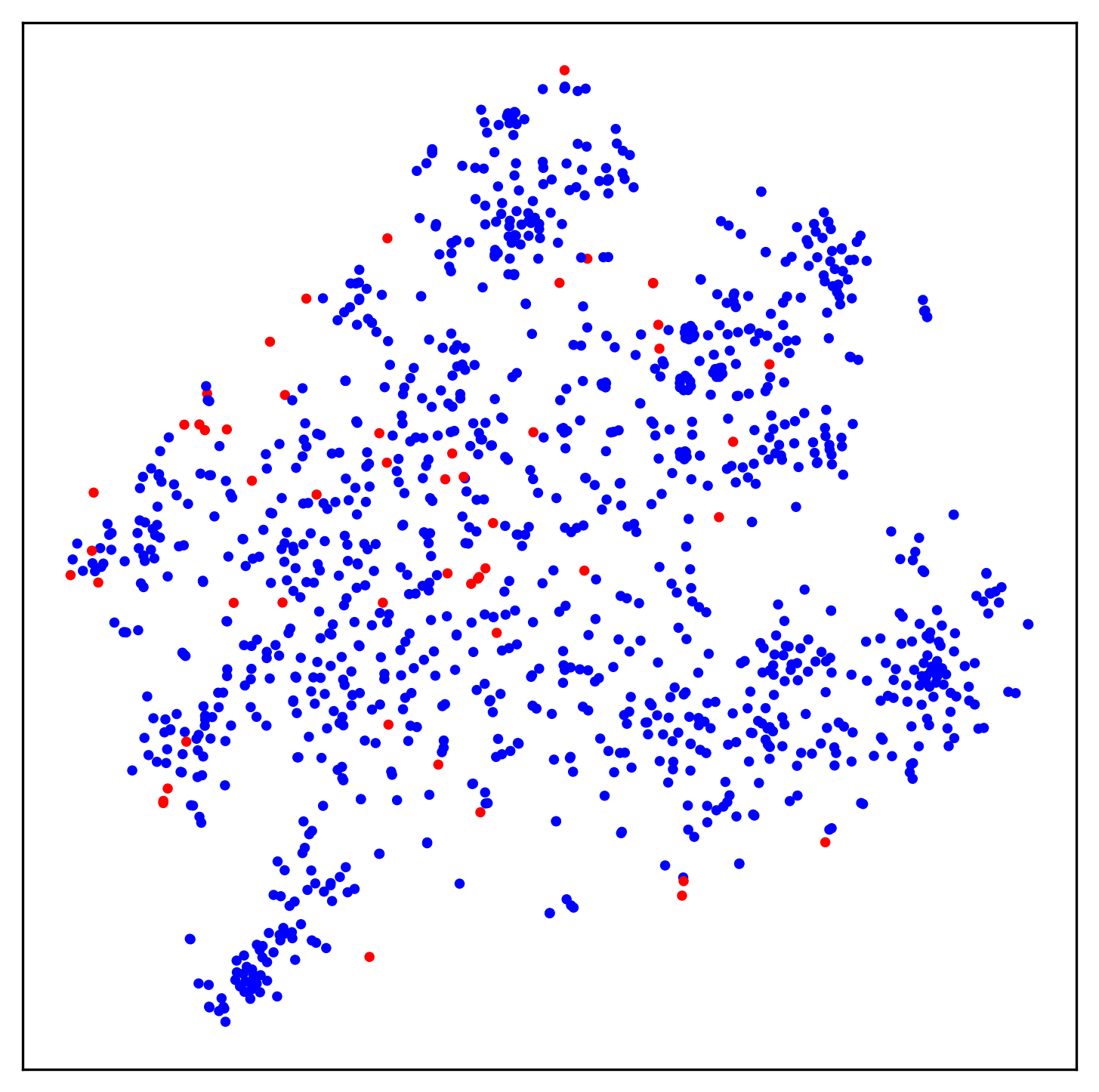} &
        \includegraphics[width=0.15\textwidth]{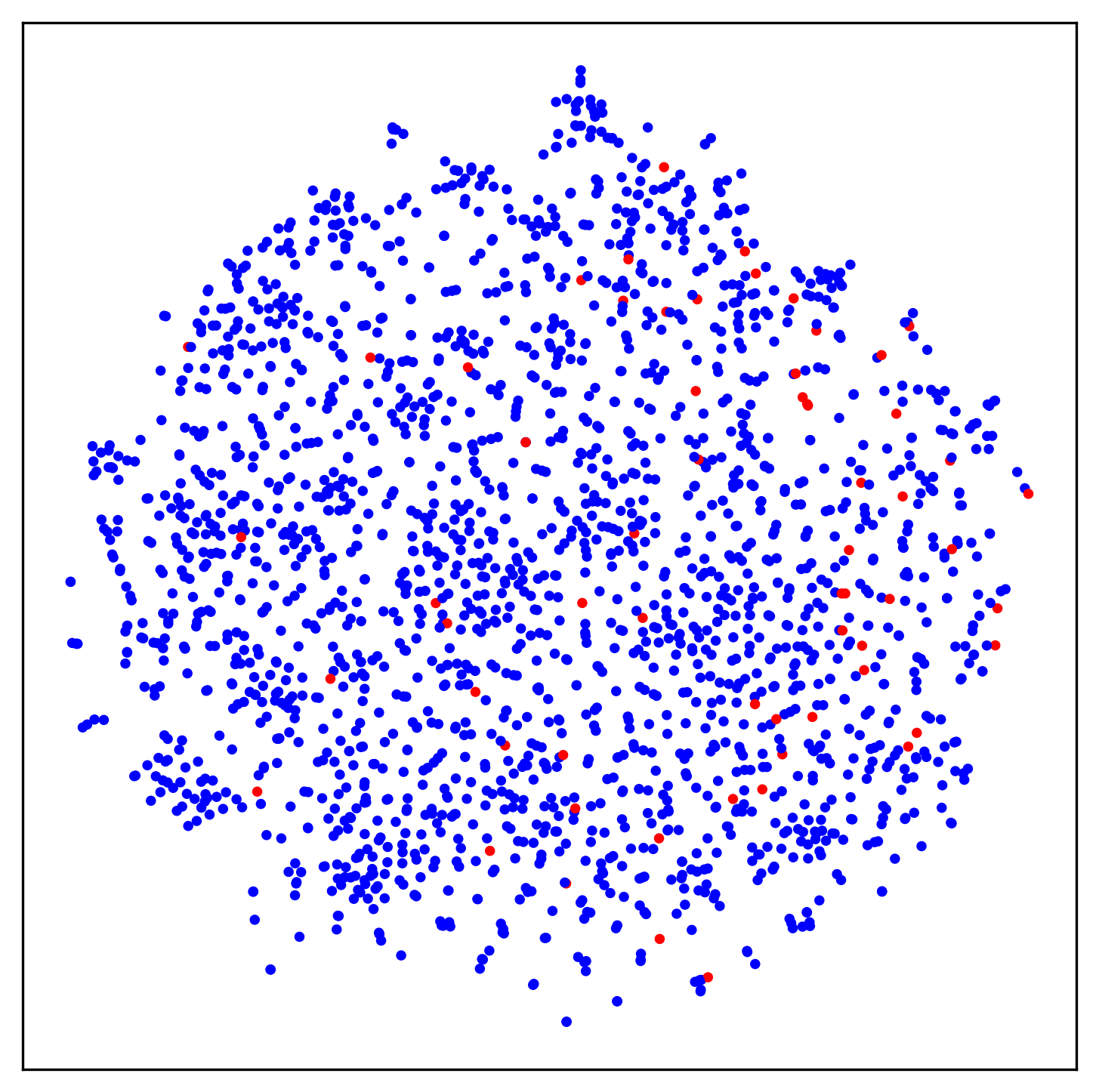} &
        \includegraphics[width=0.15\textwidth]{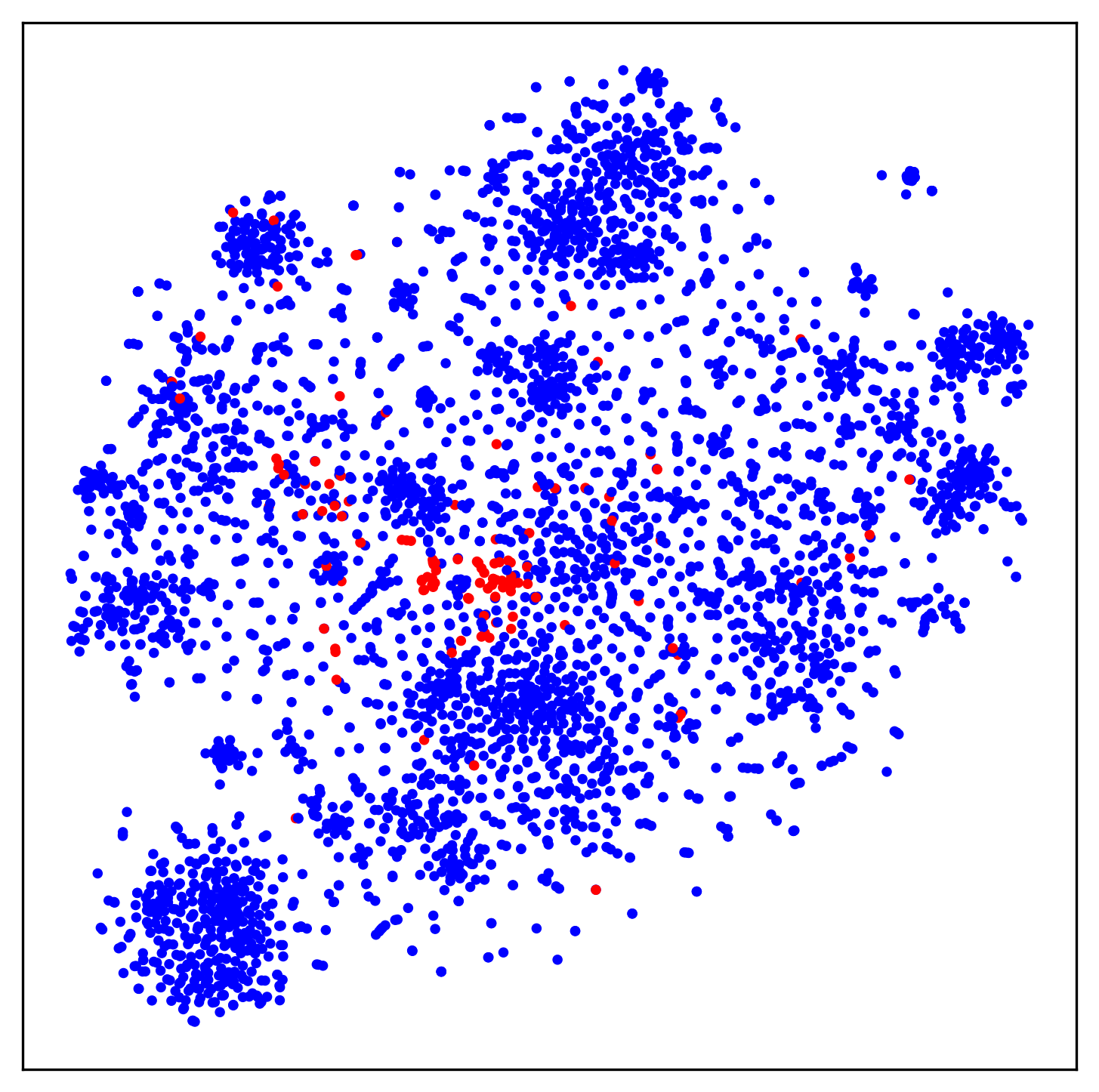} &
        \includegraphics[width=0.15\textwidth]{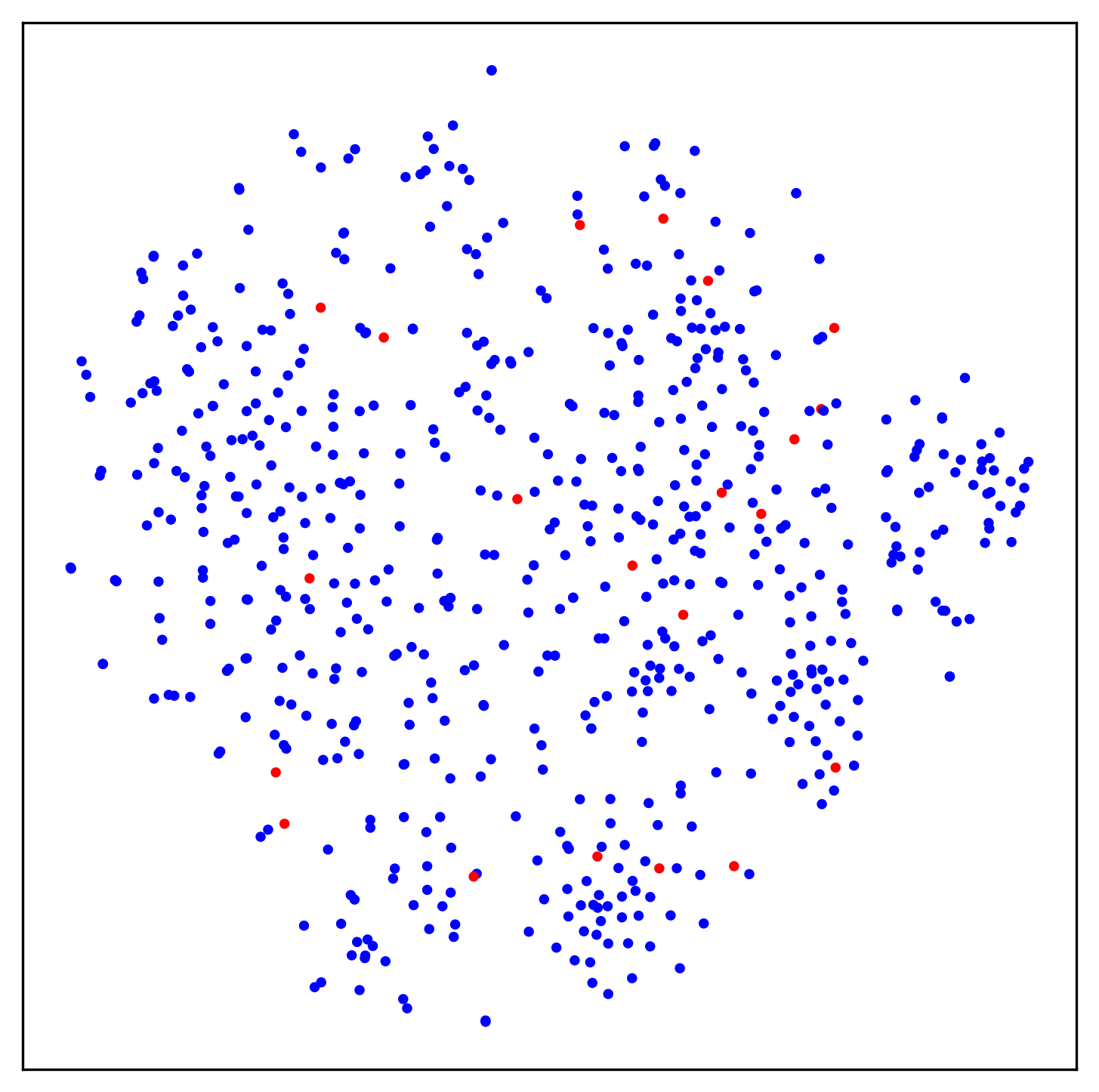}\\
        \midrule
		\textbf{O-ada}       & \includegraphics[width=0.15\textwidth]{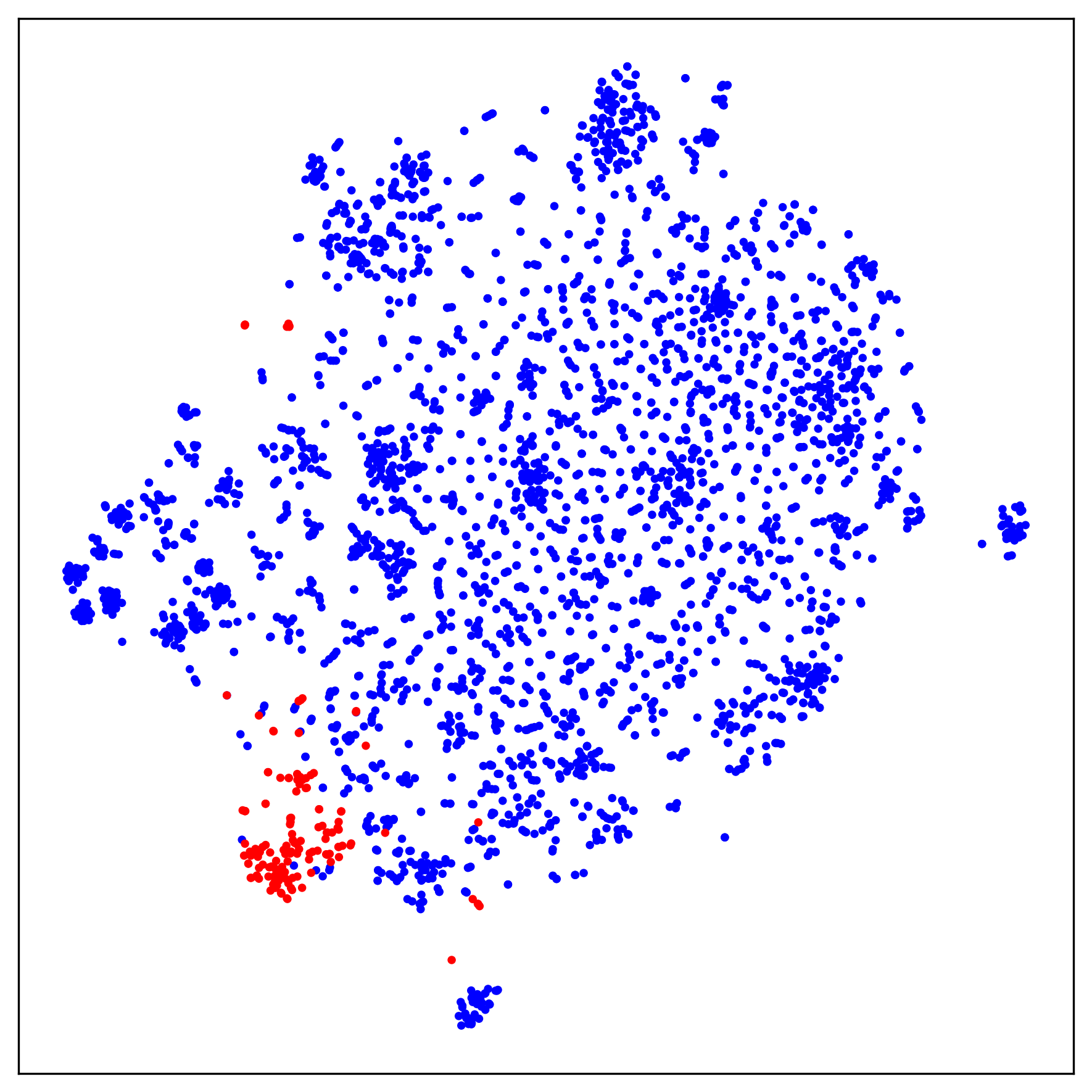} &
	\includegraphics[width=0.15\textwidth]{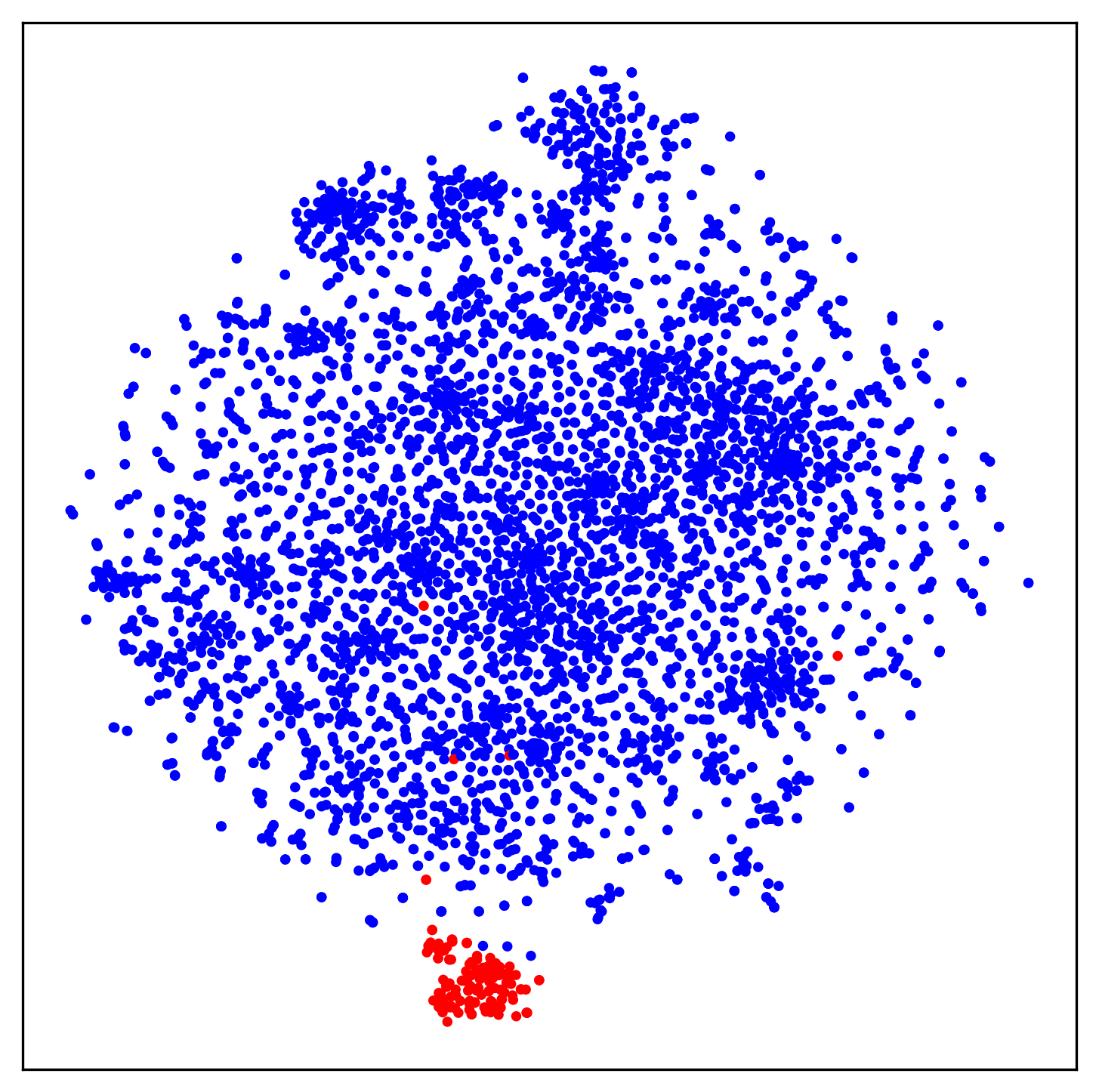} &
		\includegraphics[width=0.15\textwidth]{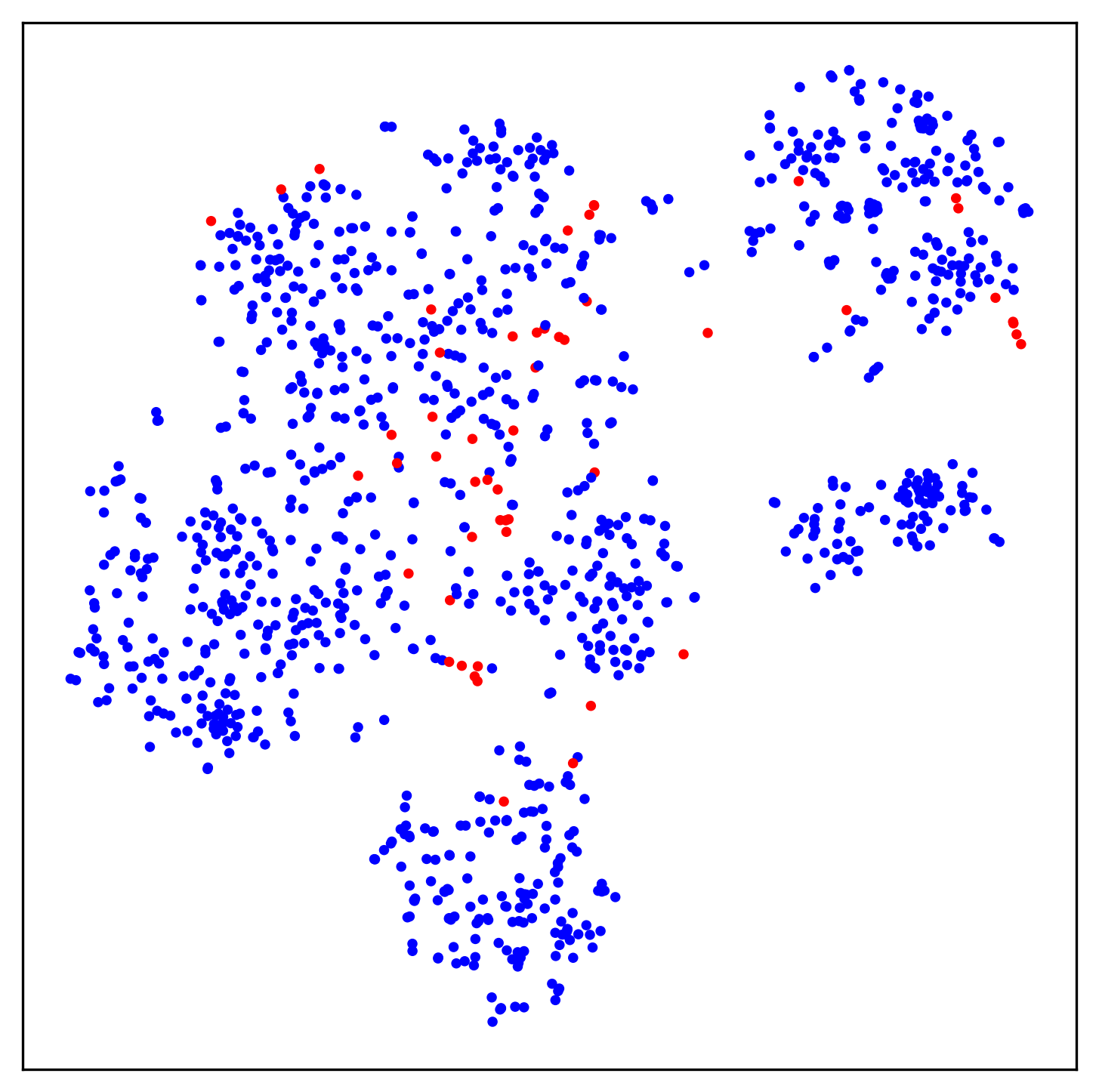} &
        \includegraphics[width=0.15\textwidth]{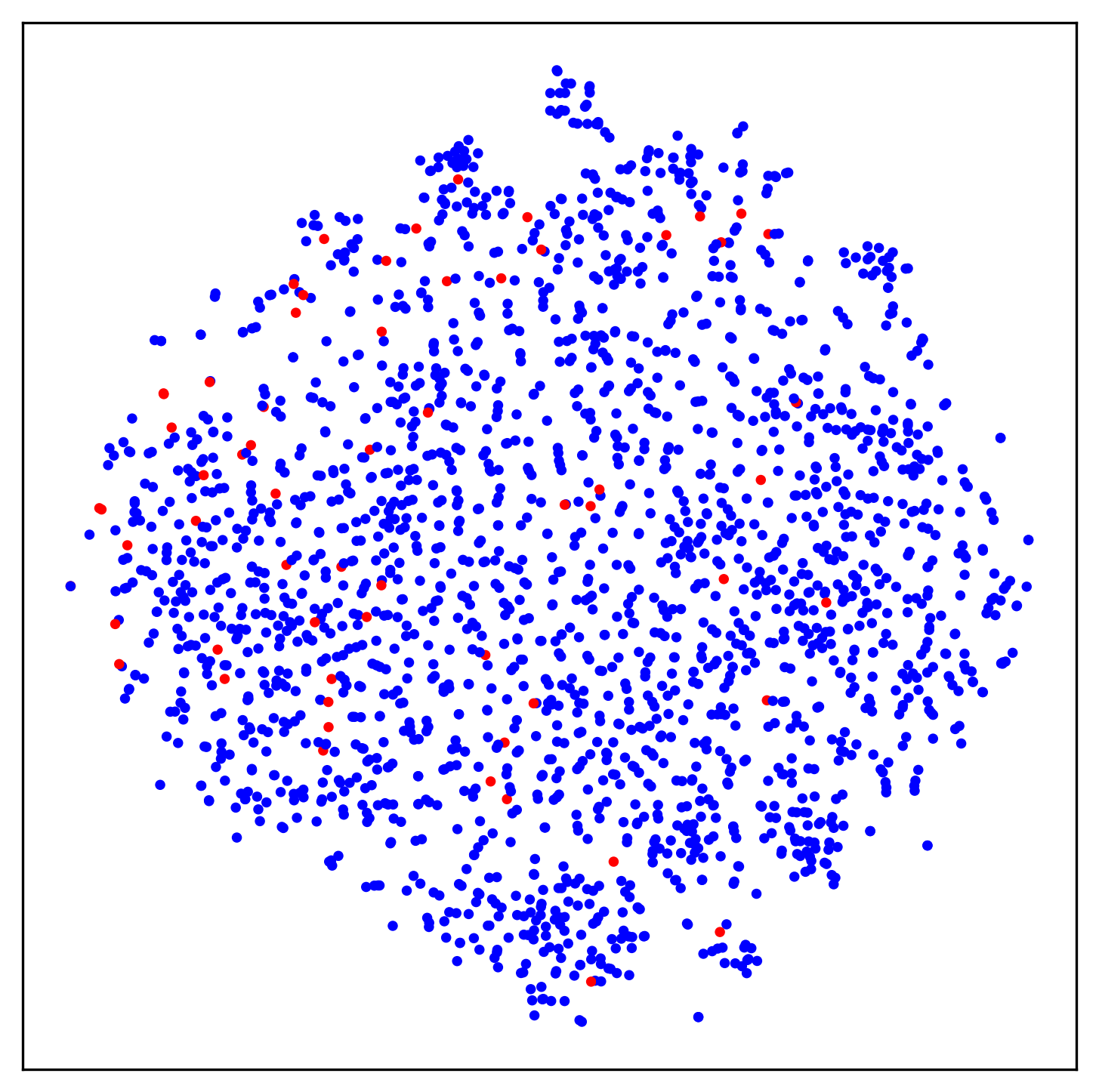} &
        \includegraphics[width=0.15\textwidth]{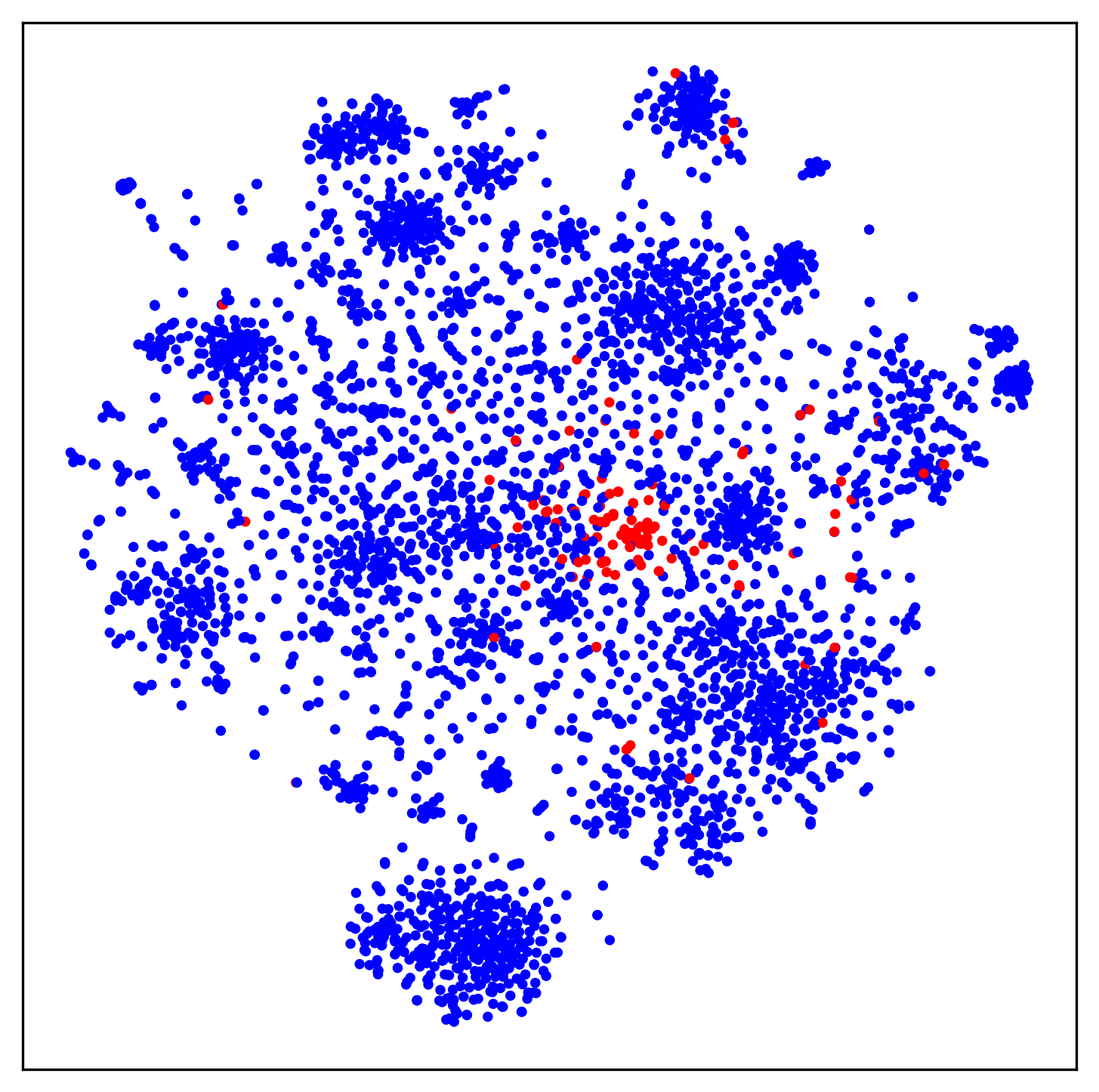} &
        \includegraphics[width=0.15\textwidth]{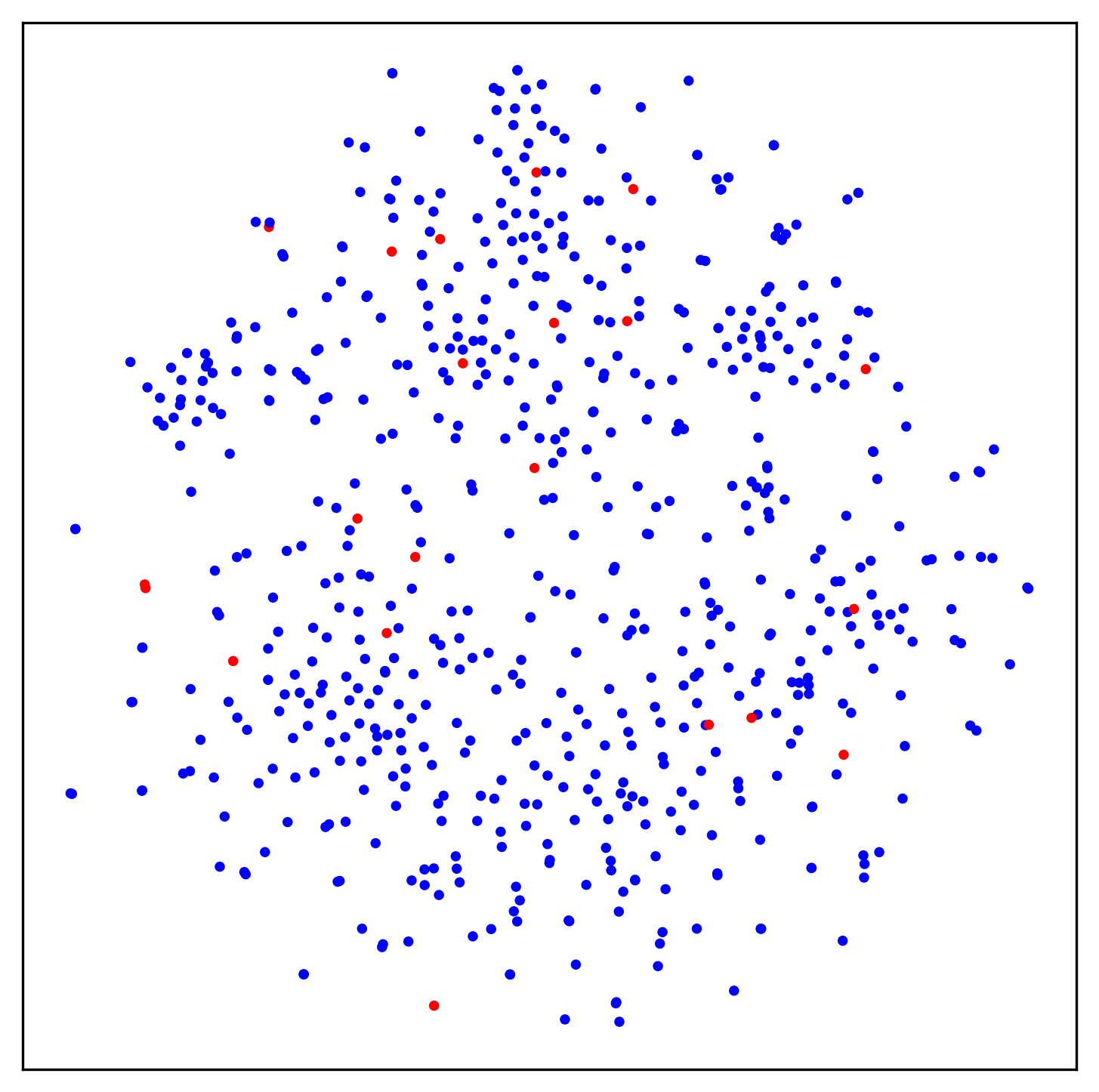}\\
        \midrule
		\textbf{O-small}       & \includegraphics[width=0.15\textwidth]{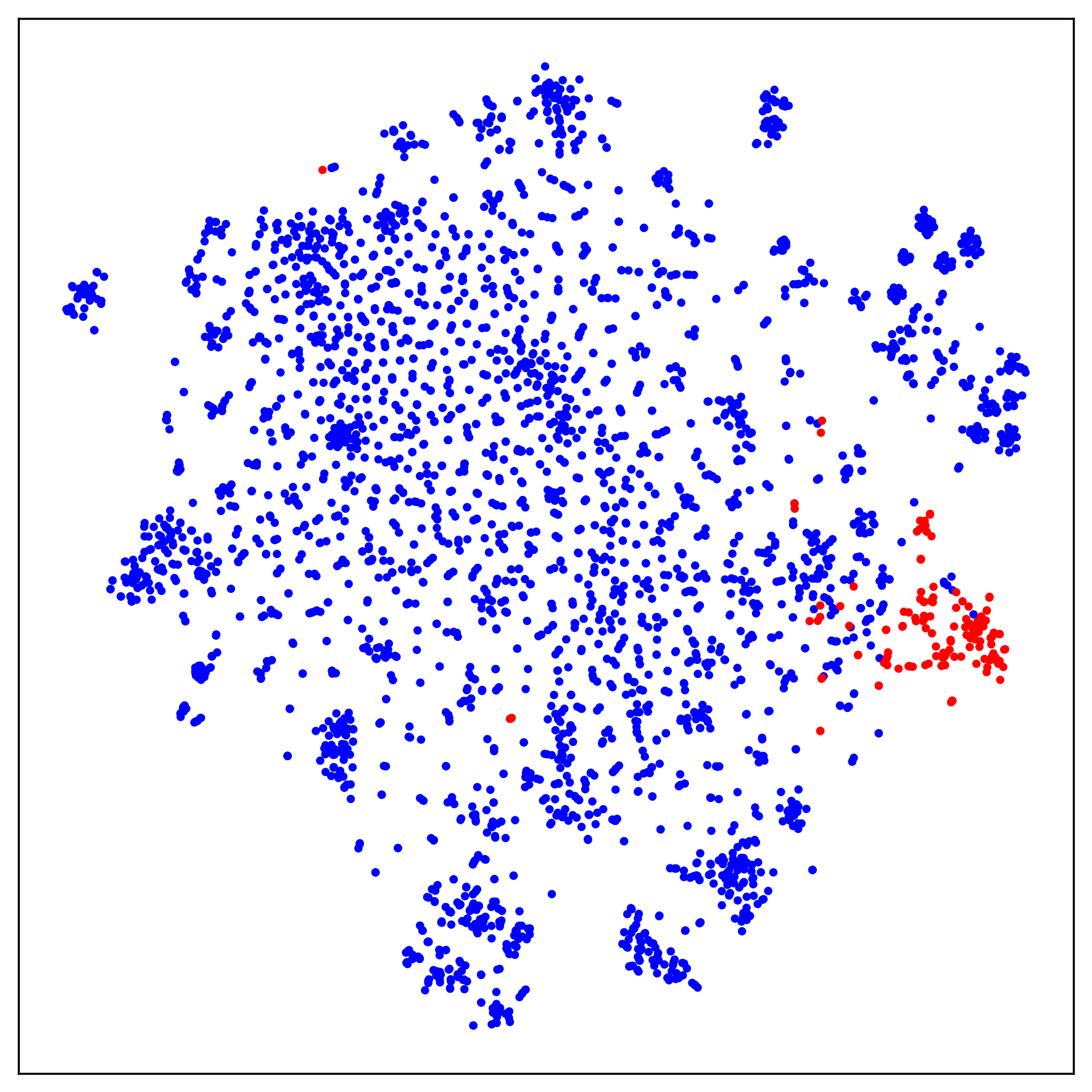} &
	\includegraphics[width=0.15\textwidth]{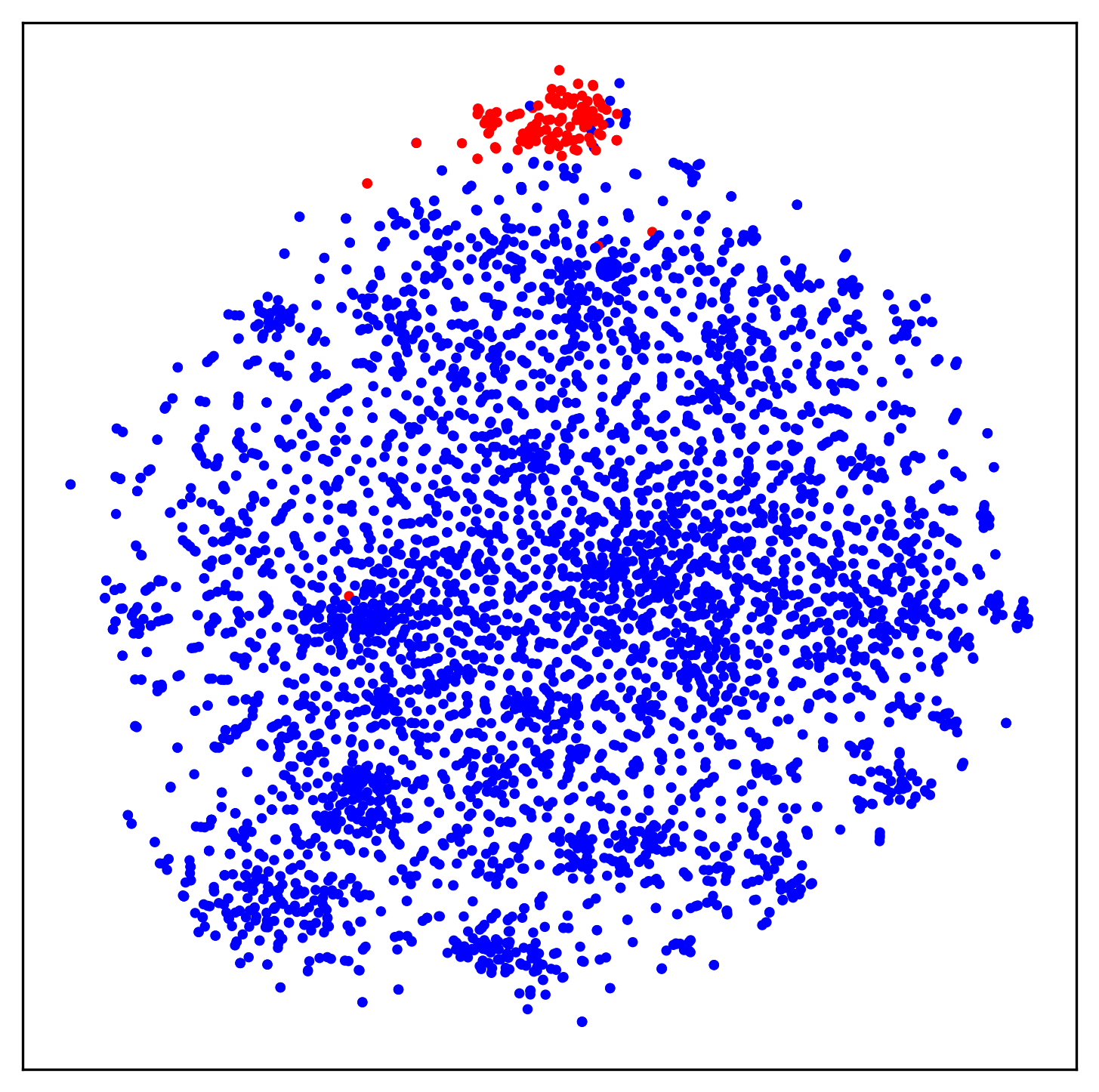} &
		\includegraphics[width=0.15\textwidth]{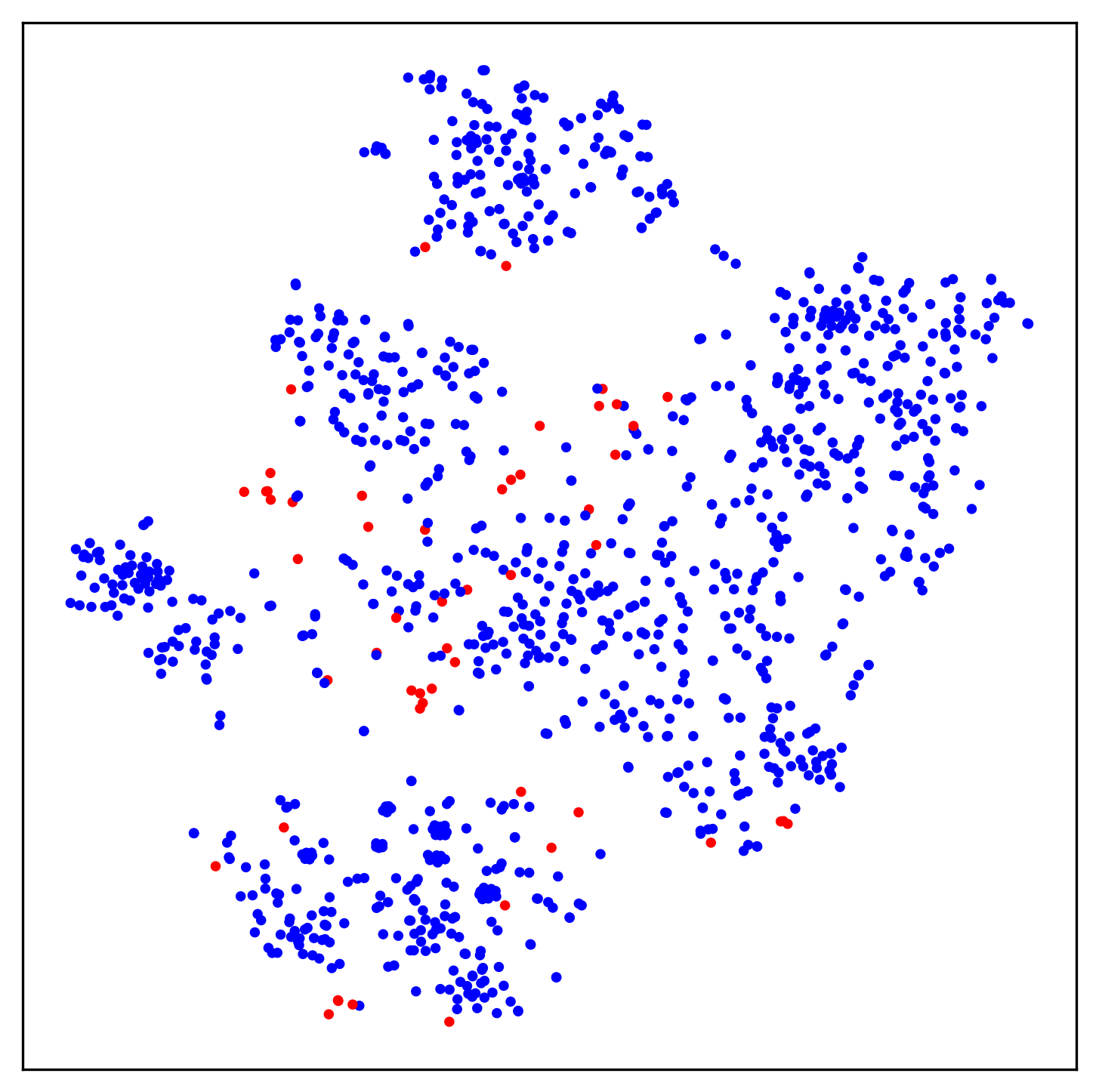} &
        \includegraphics[width=0.15\textwidth]{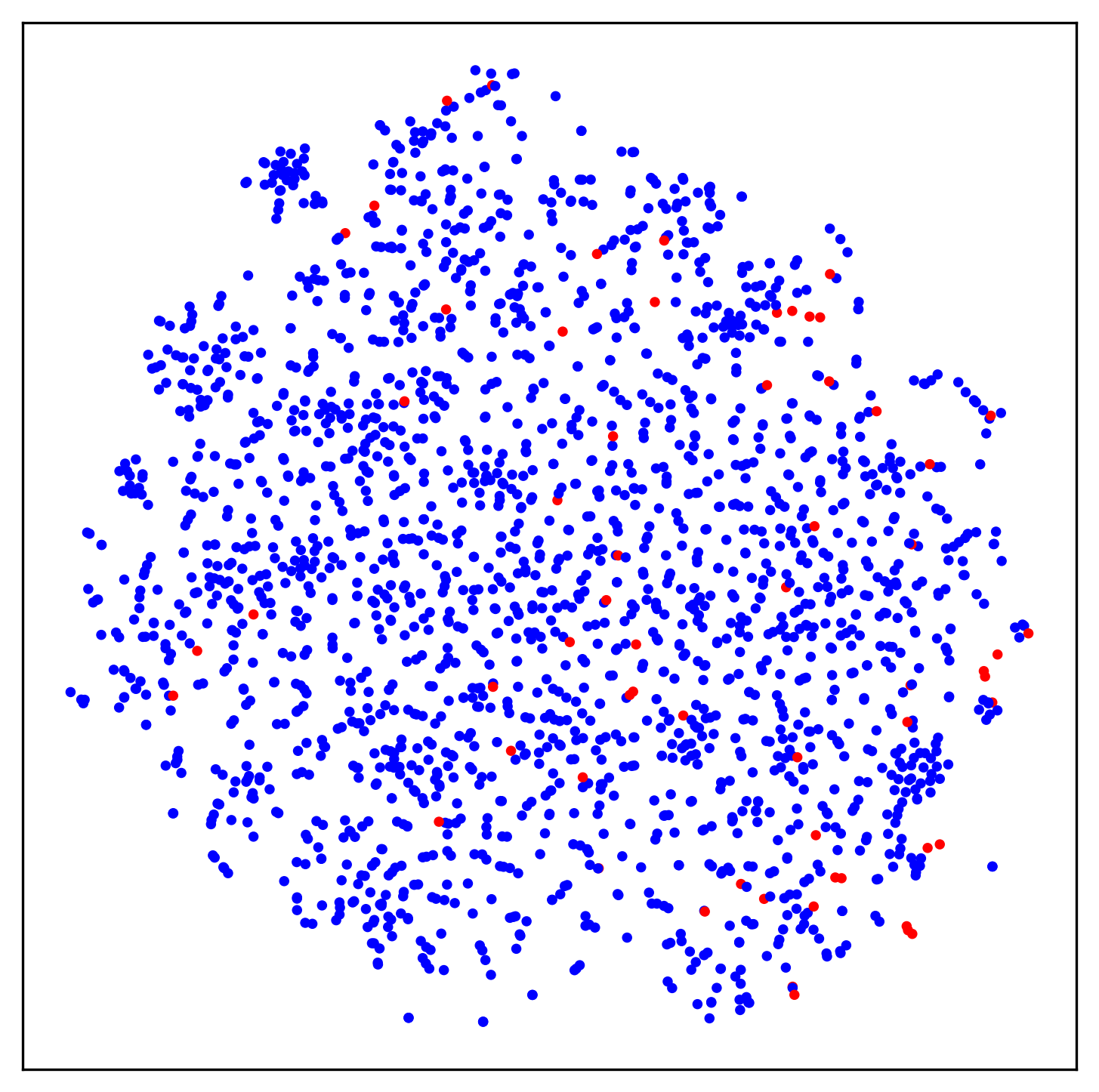} &
        \includegraphics[width=0.15\textwidth]{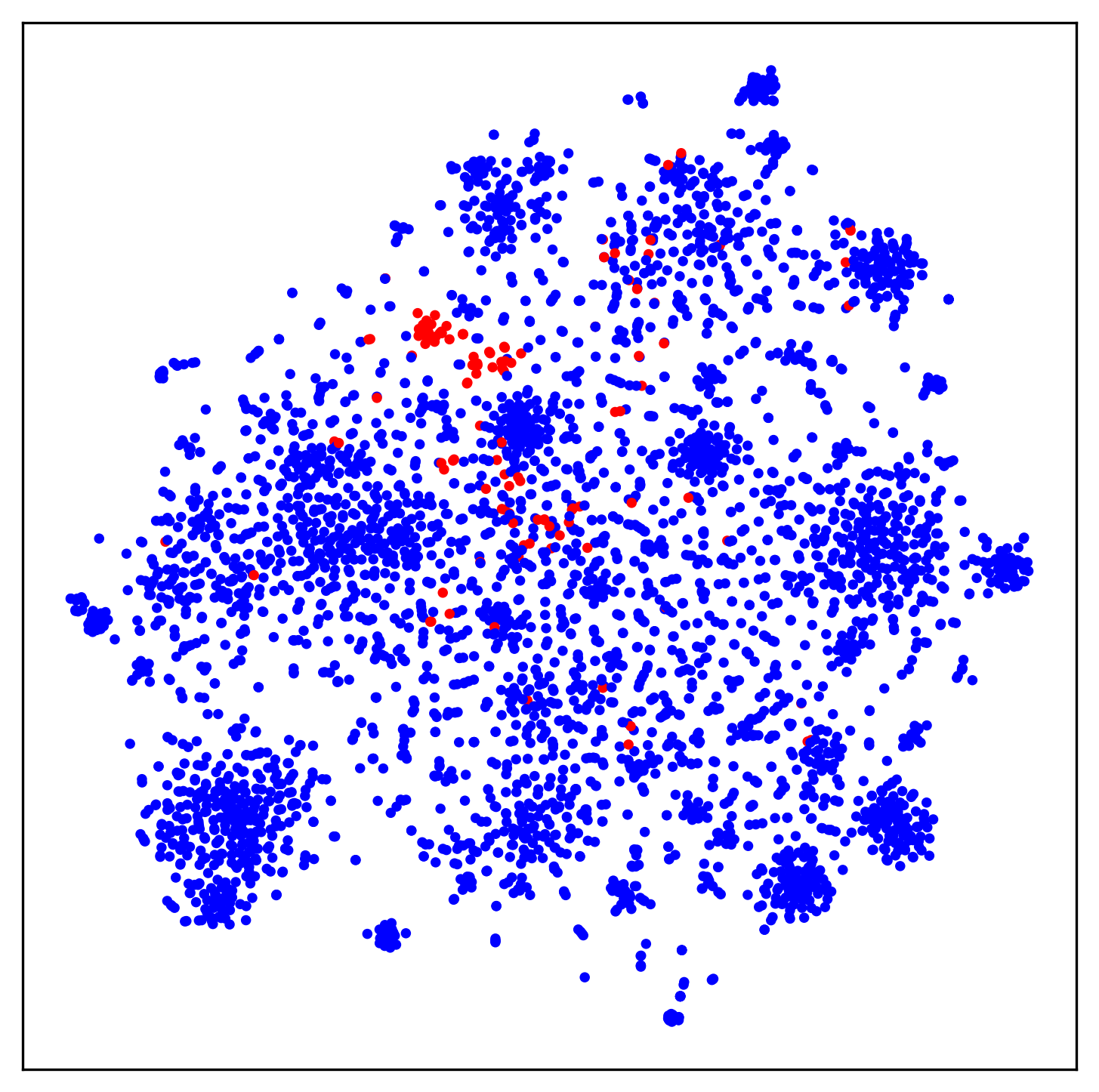} &
        \includegraphics[width=0.15\textwidth]{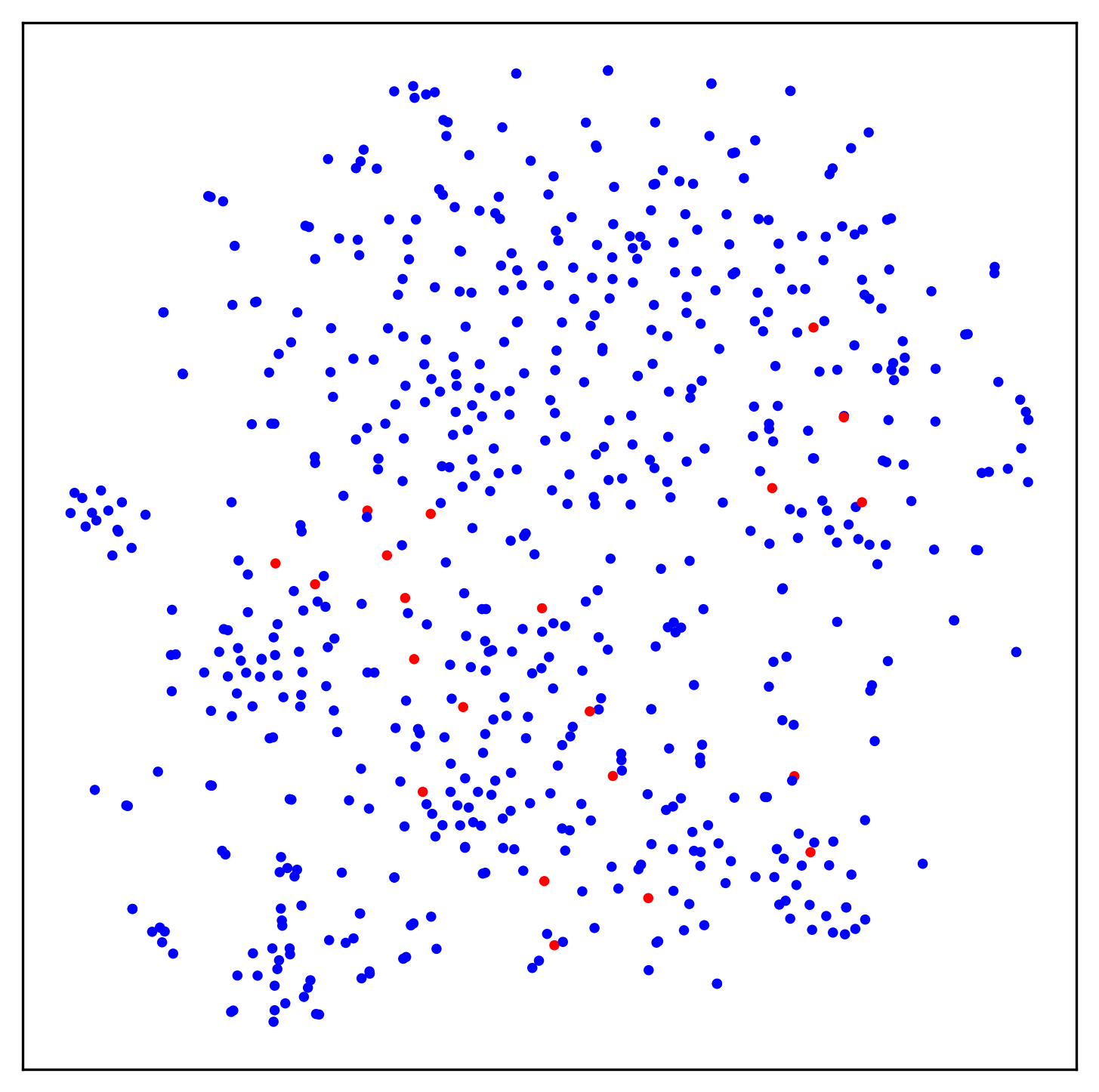}\\
        \midrule
		\textbf{O-large}       & \includegraphics[width=0.15\textwidth]{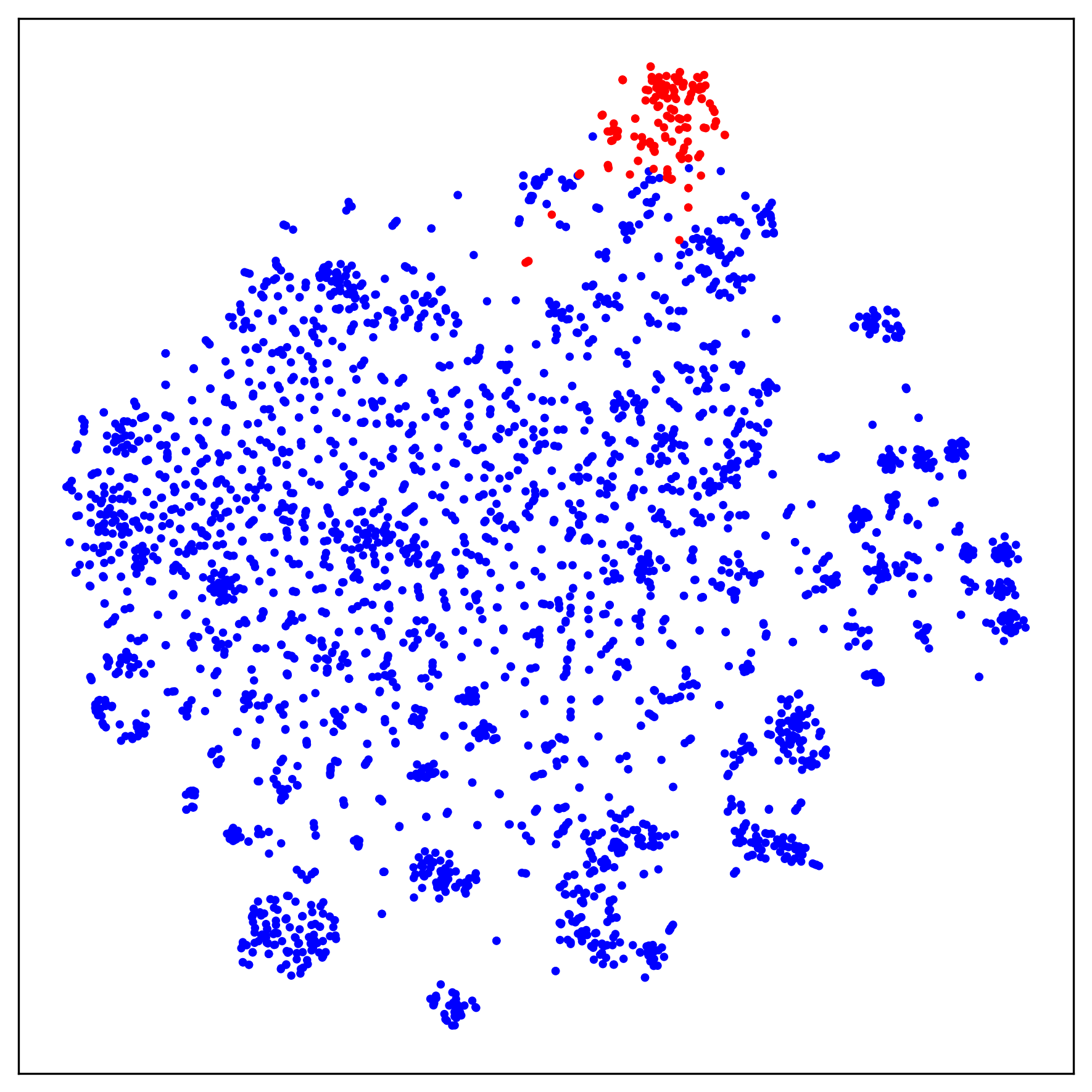} &
	\includegraphics[width=0.15\textwidth]{figures/ebd/openai_large_smsspam.png} &
		\includegraphics[width=0.15\textwidth]{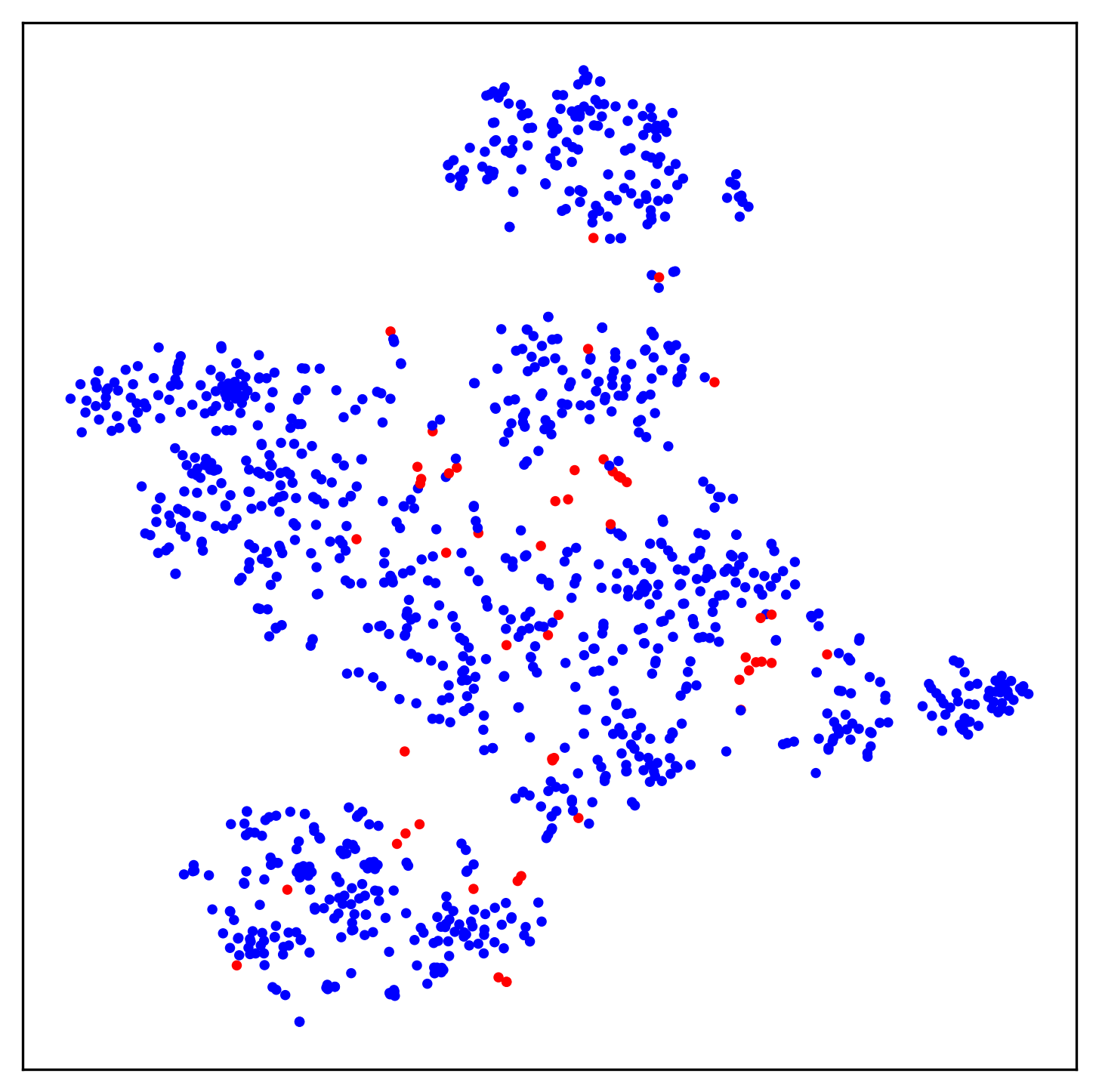} &
        \includegraphics[width=0.15\textwidth]{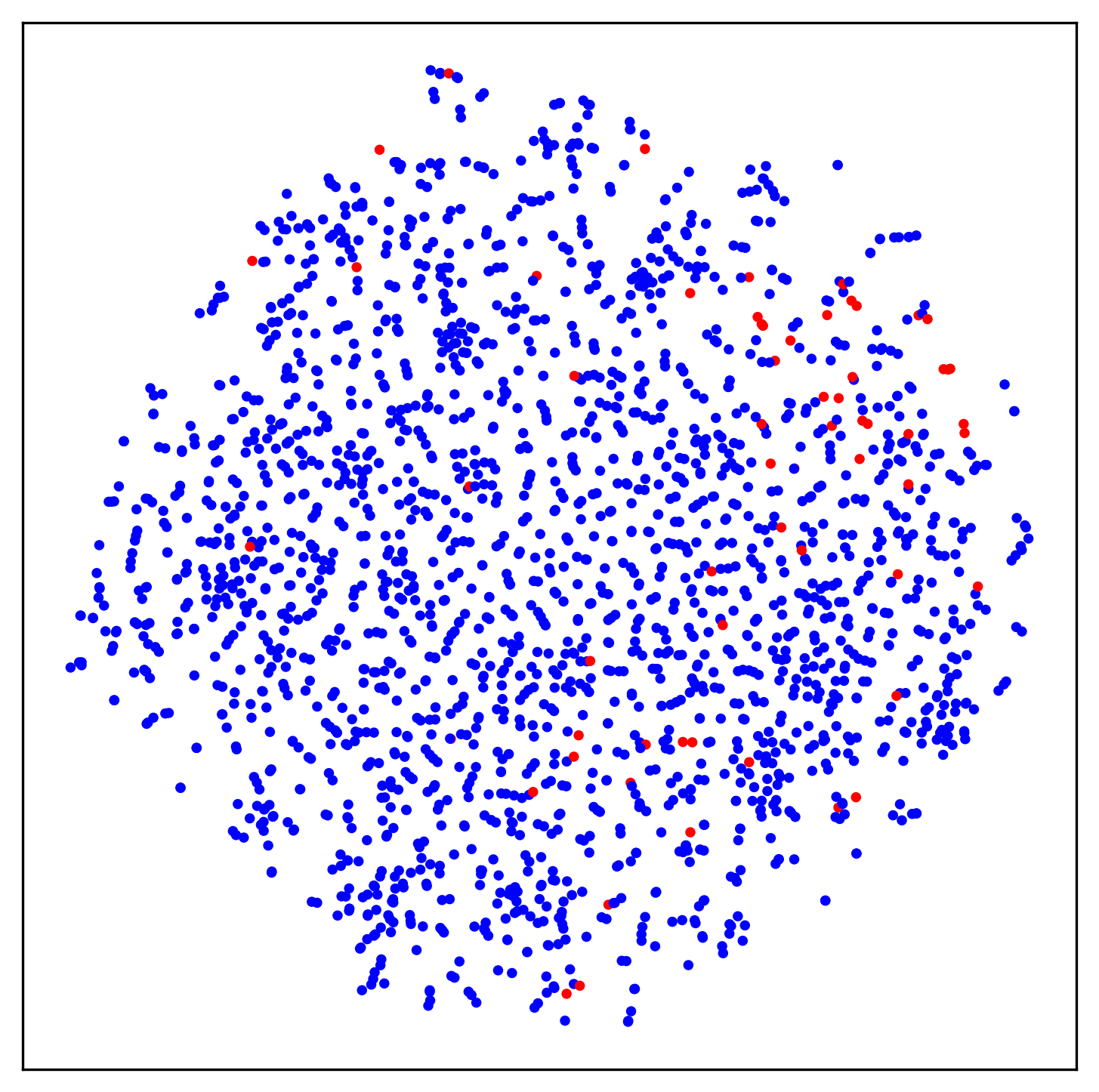} &
        \includegraphics[width=0.15\textwidth]{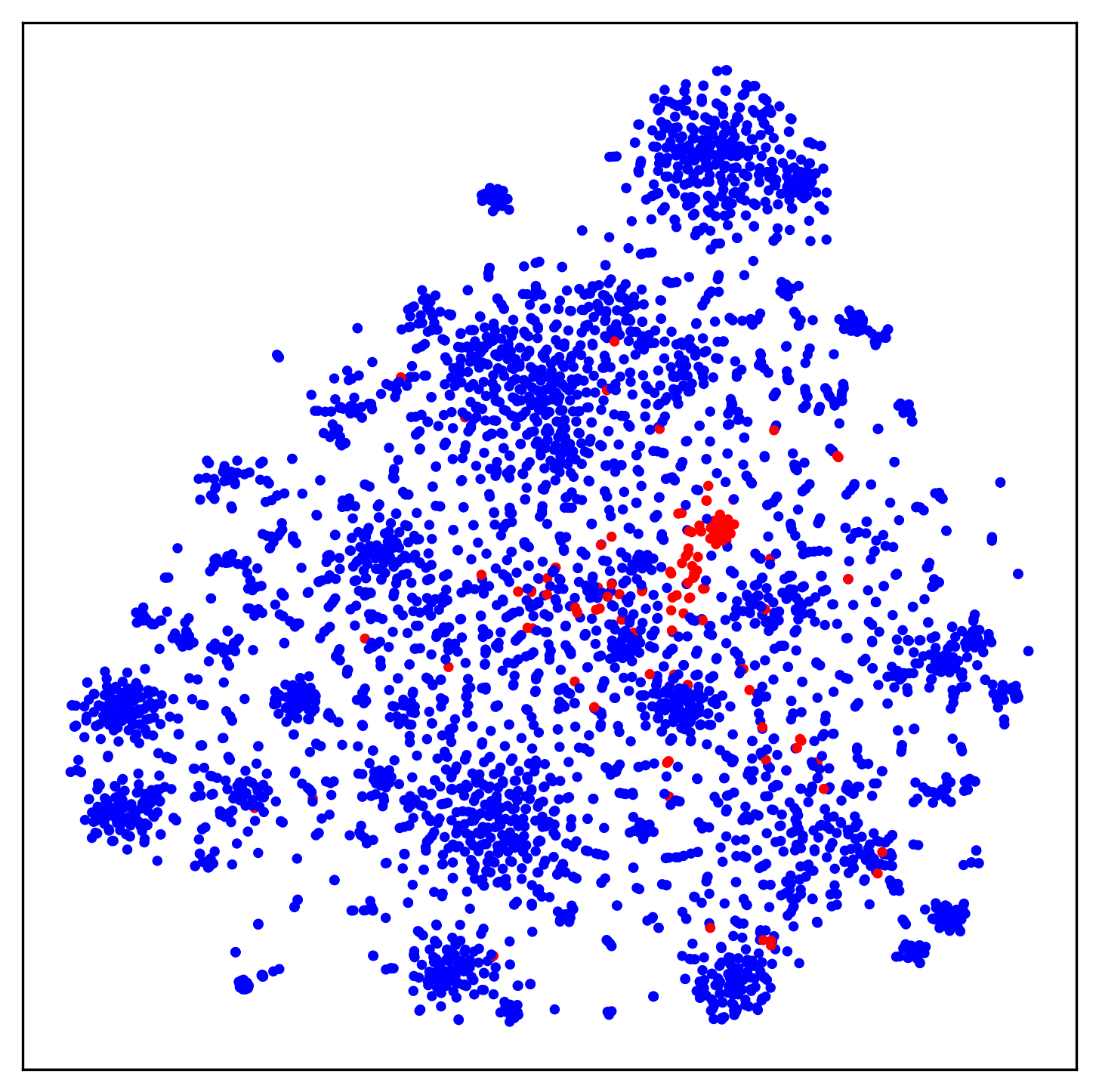} &
        \includegraphics[width=0.15\textwidth]{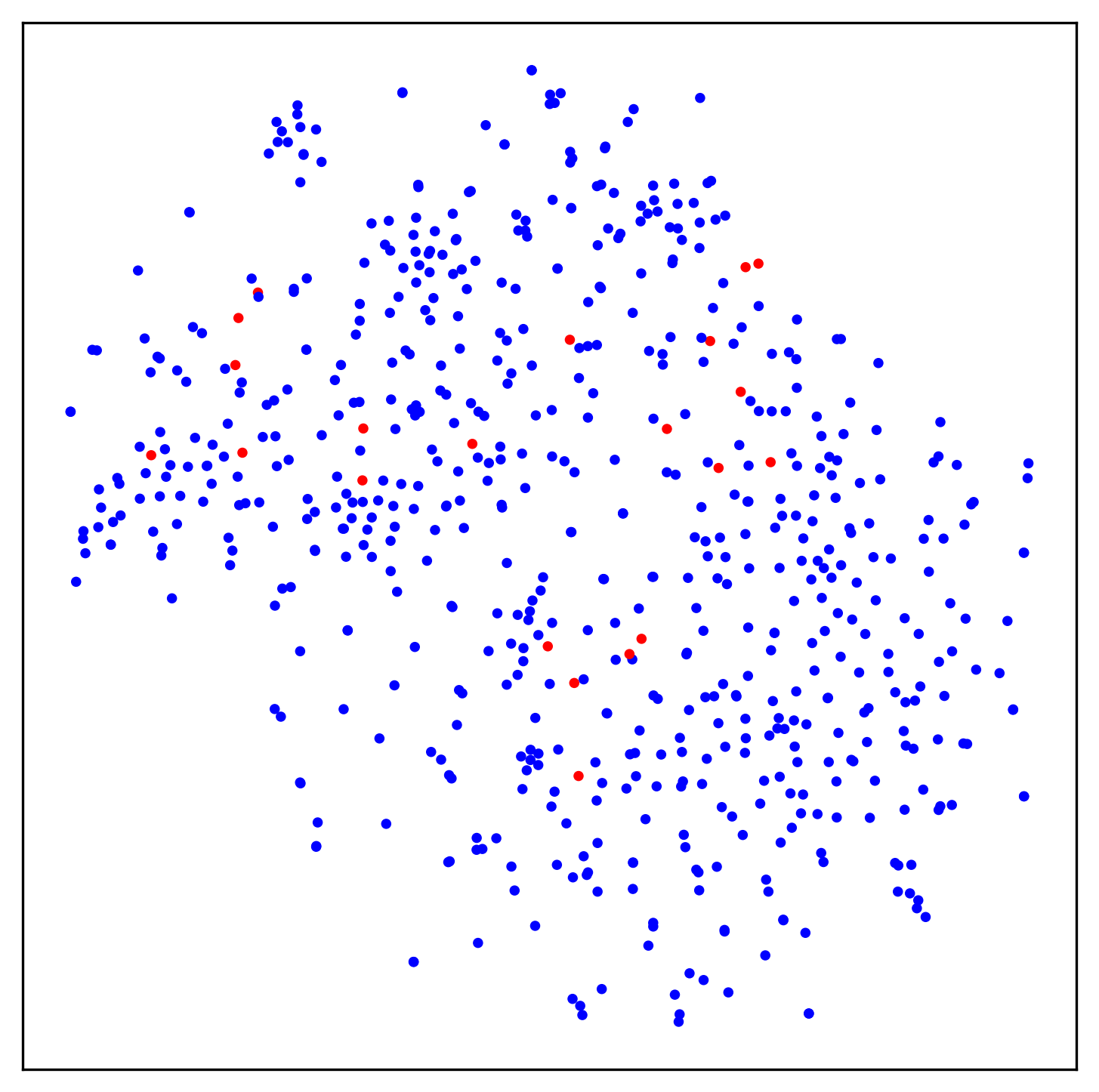}\\
        \midrule
		\textbf{Llama}       & \includegraphics[width=0.15\textwidth]{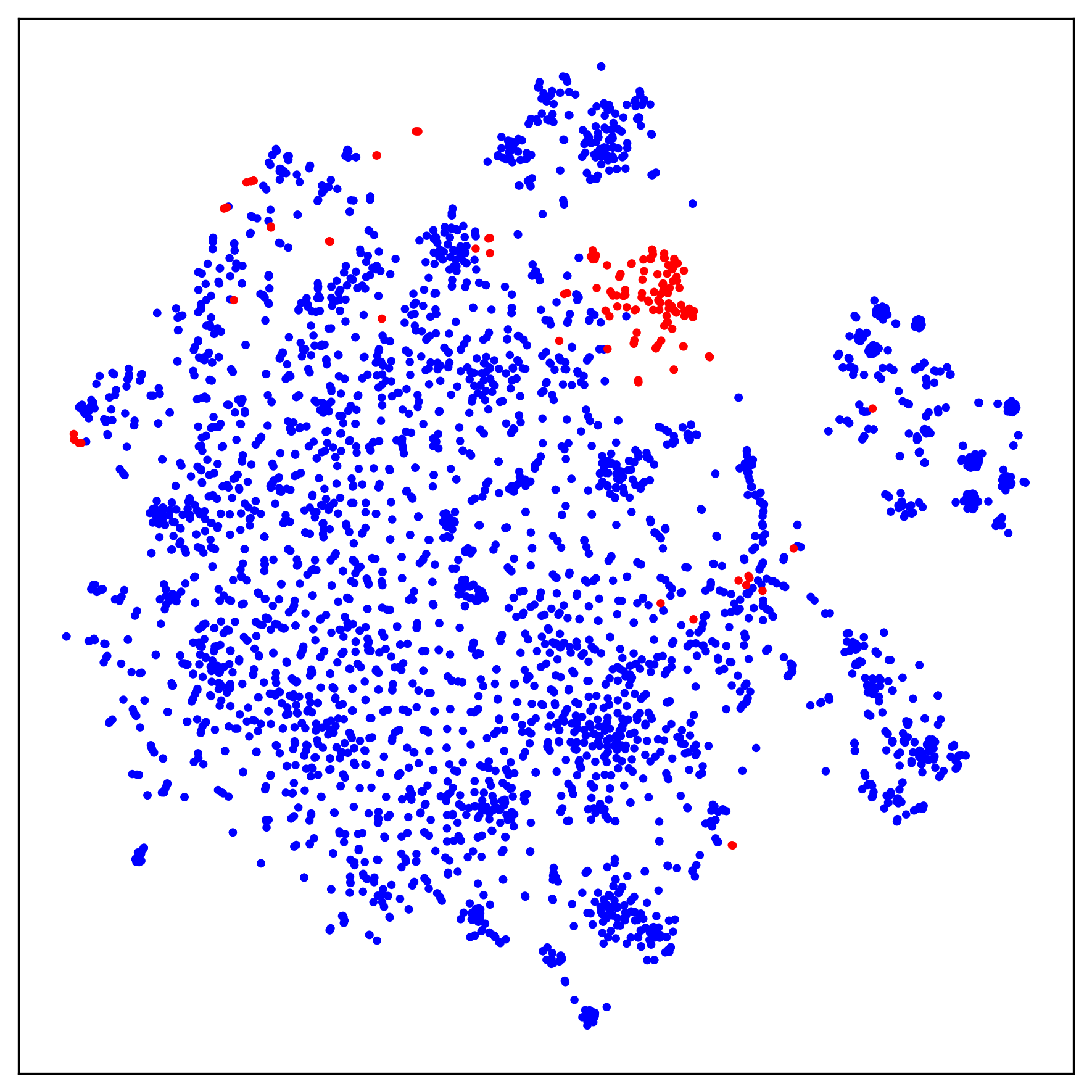} &
	\includegraphics[width=0.15\textwidth]{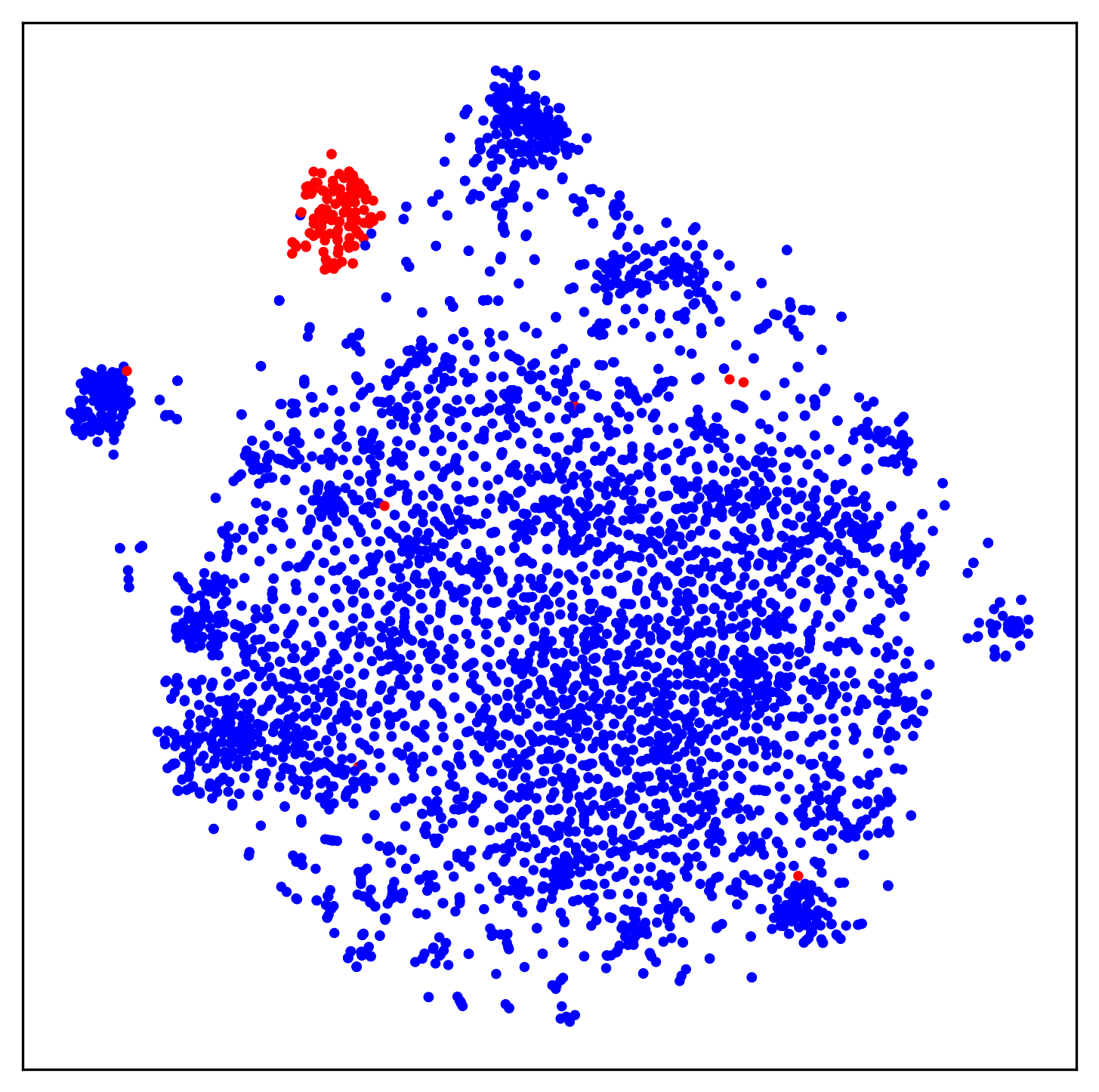} &
		\includegraphics[width=0.15\textwidth]{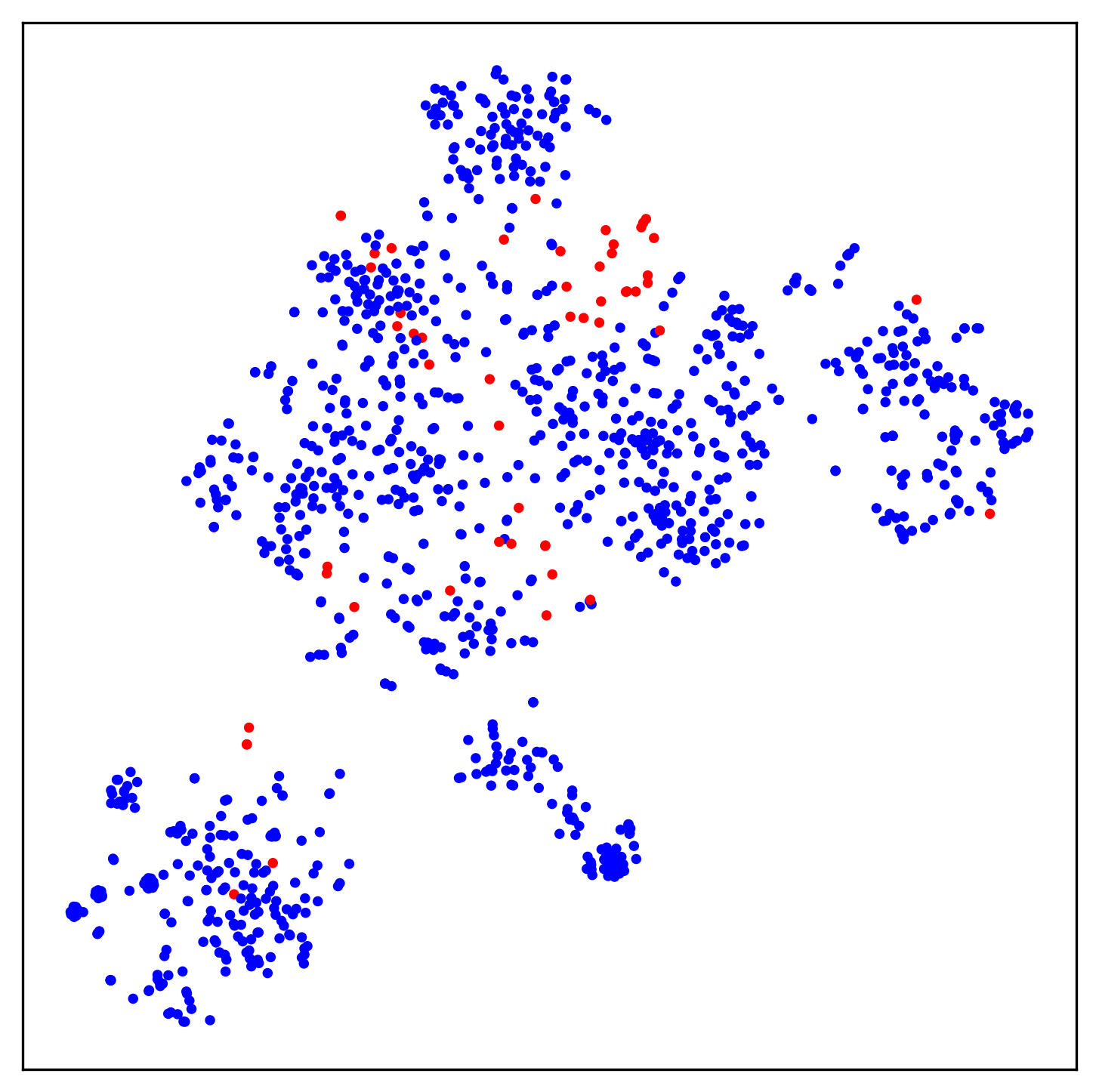} &
        \includegraphics[width=0.15\textwidth]{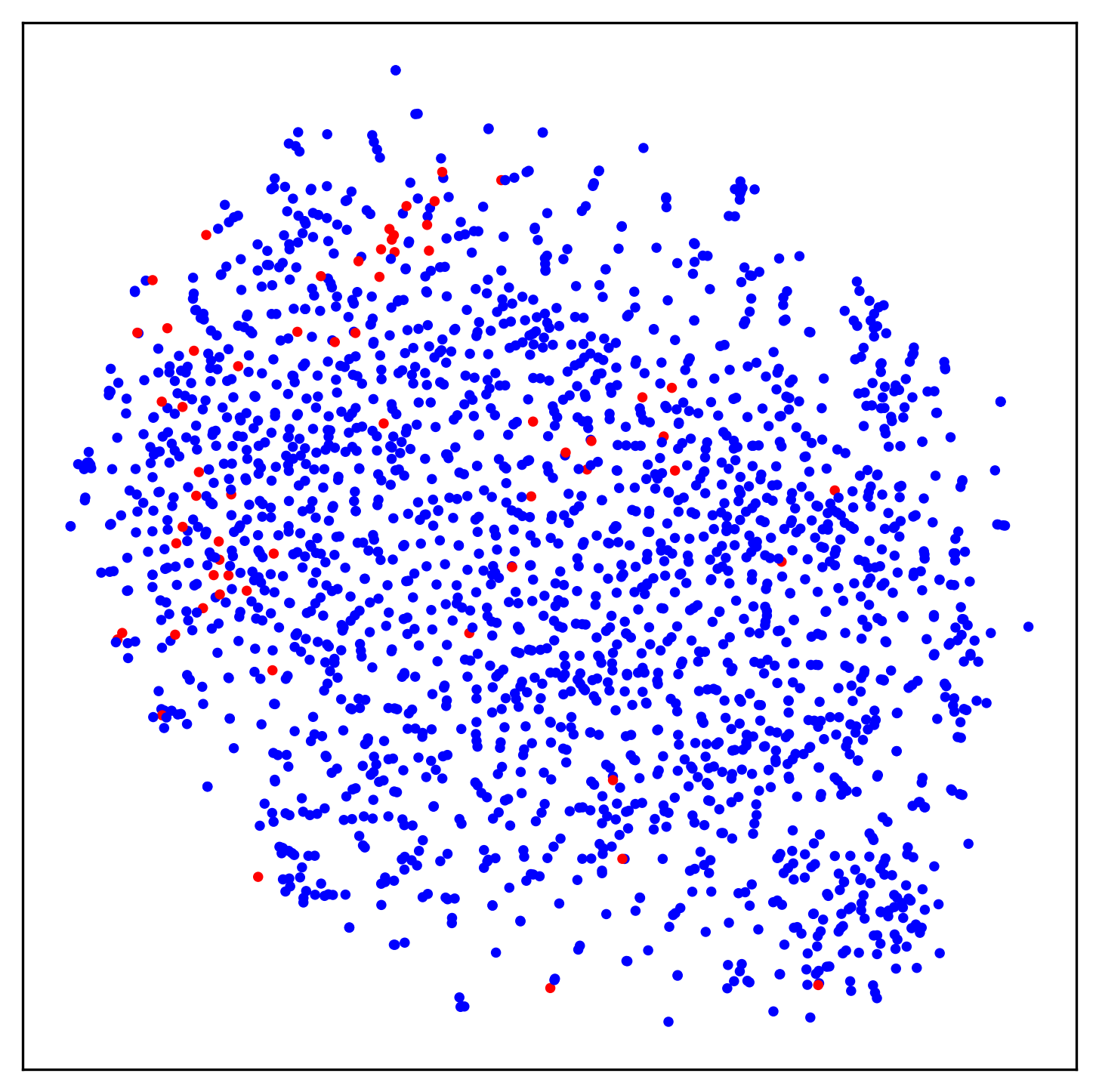} &
        \includegraphics[width=0.15\textwidth]{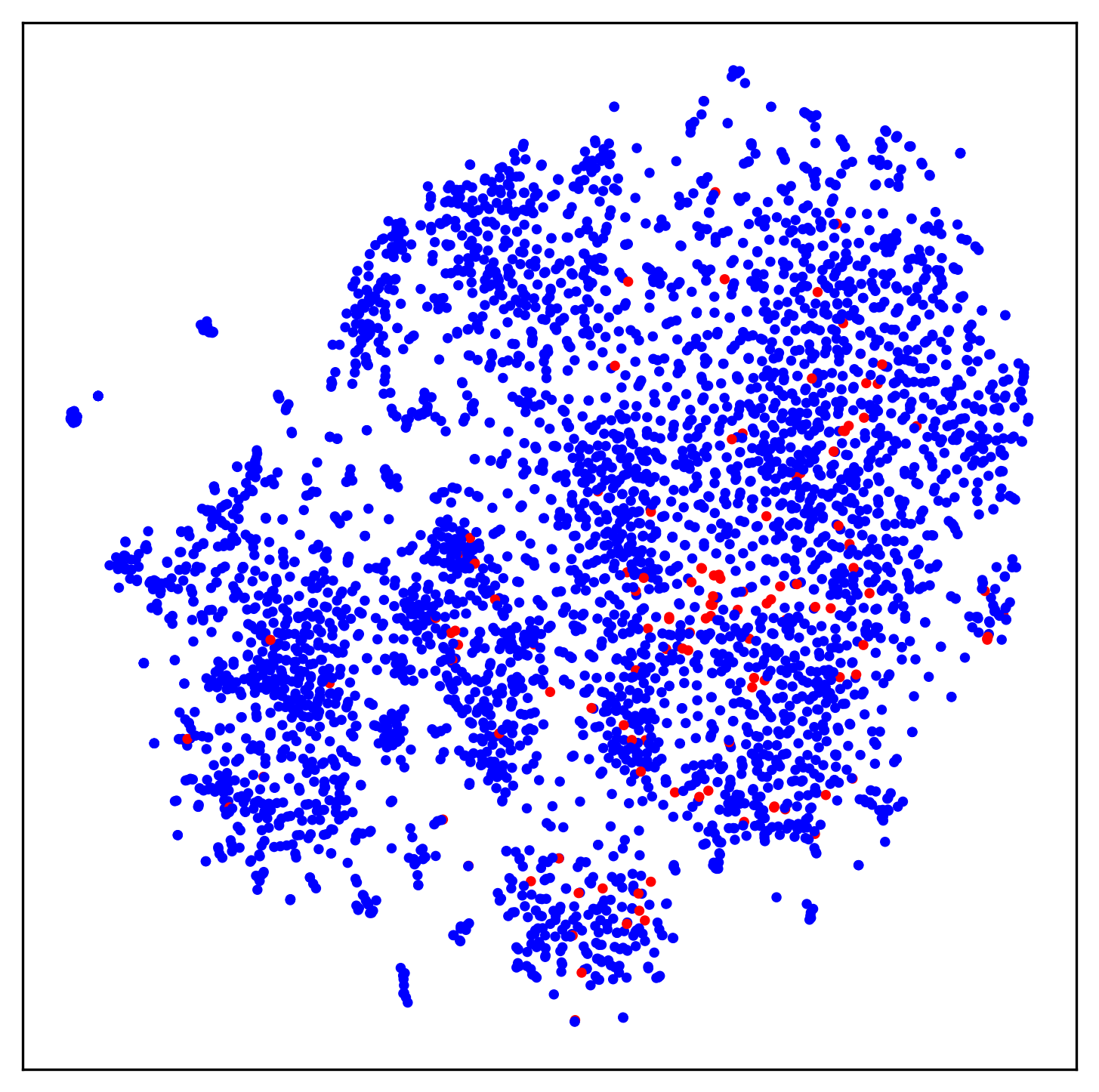} &
        \includegraphics[width=0.15\textwidth]{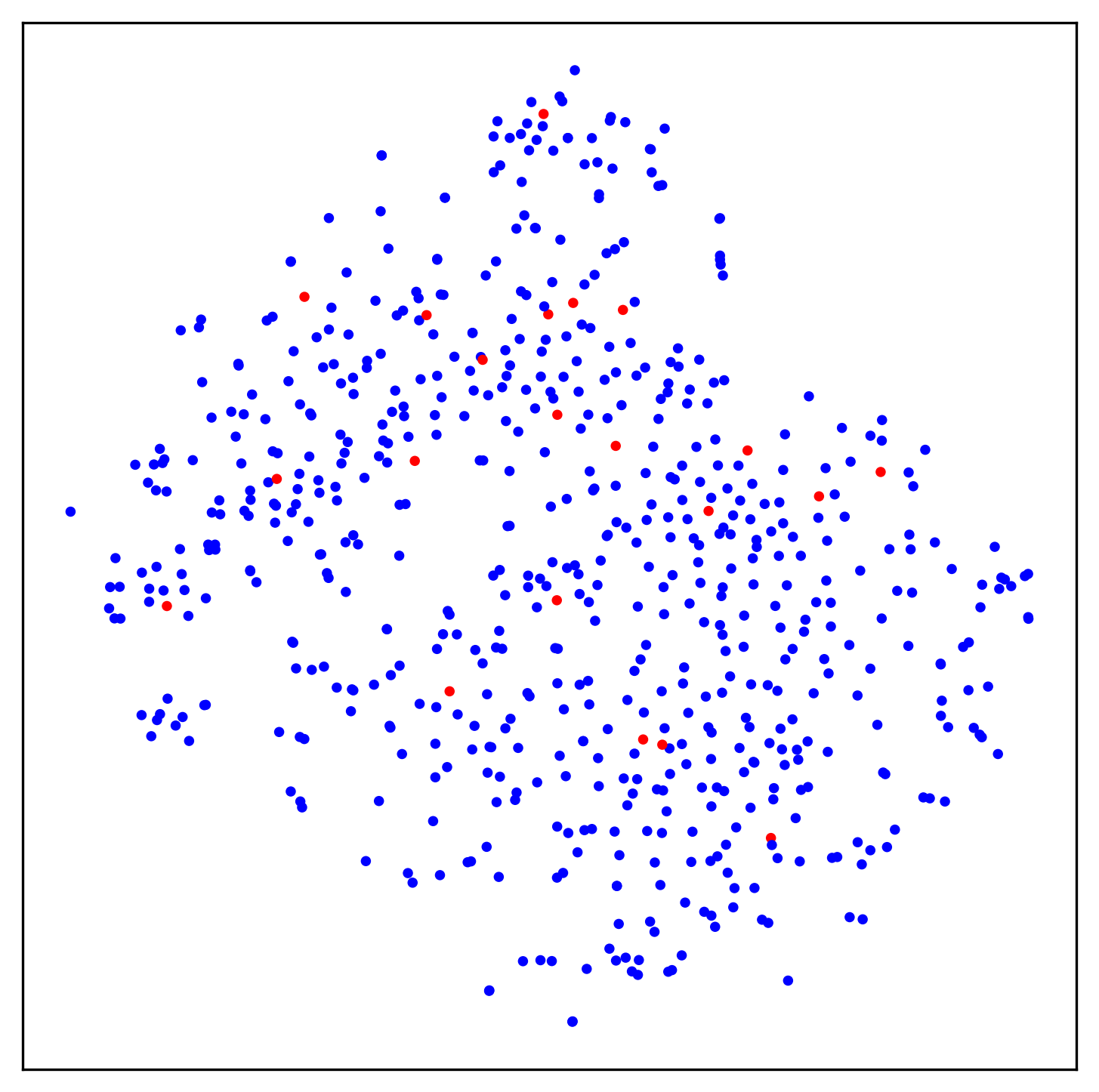}\\
        \midrule
		\textbf{stella}       & \includegraphics[width=0.15\textwidth]{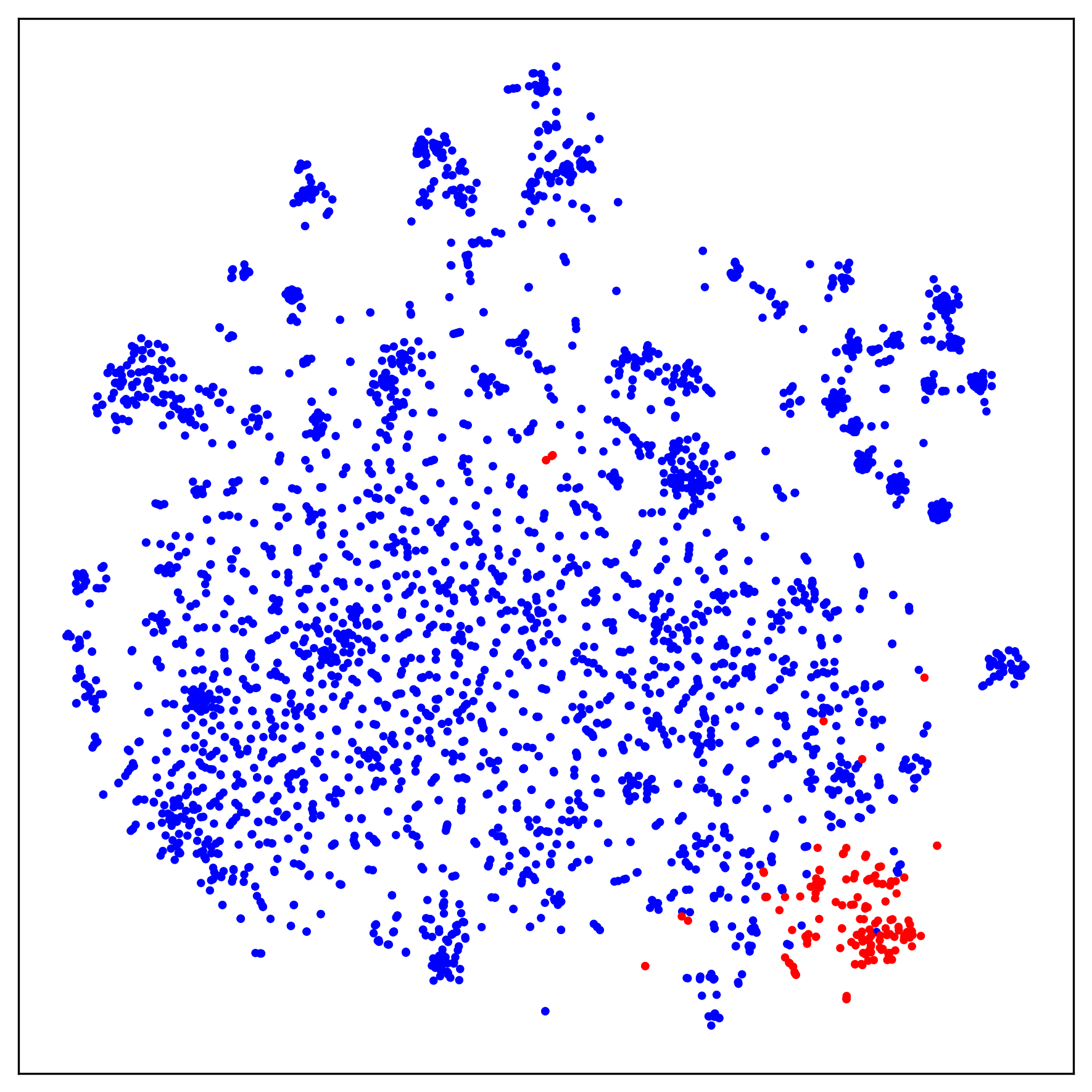} &
	\includegraphics[width=0.15\textwidth]{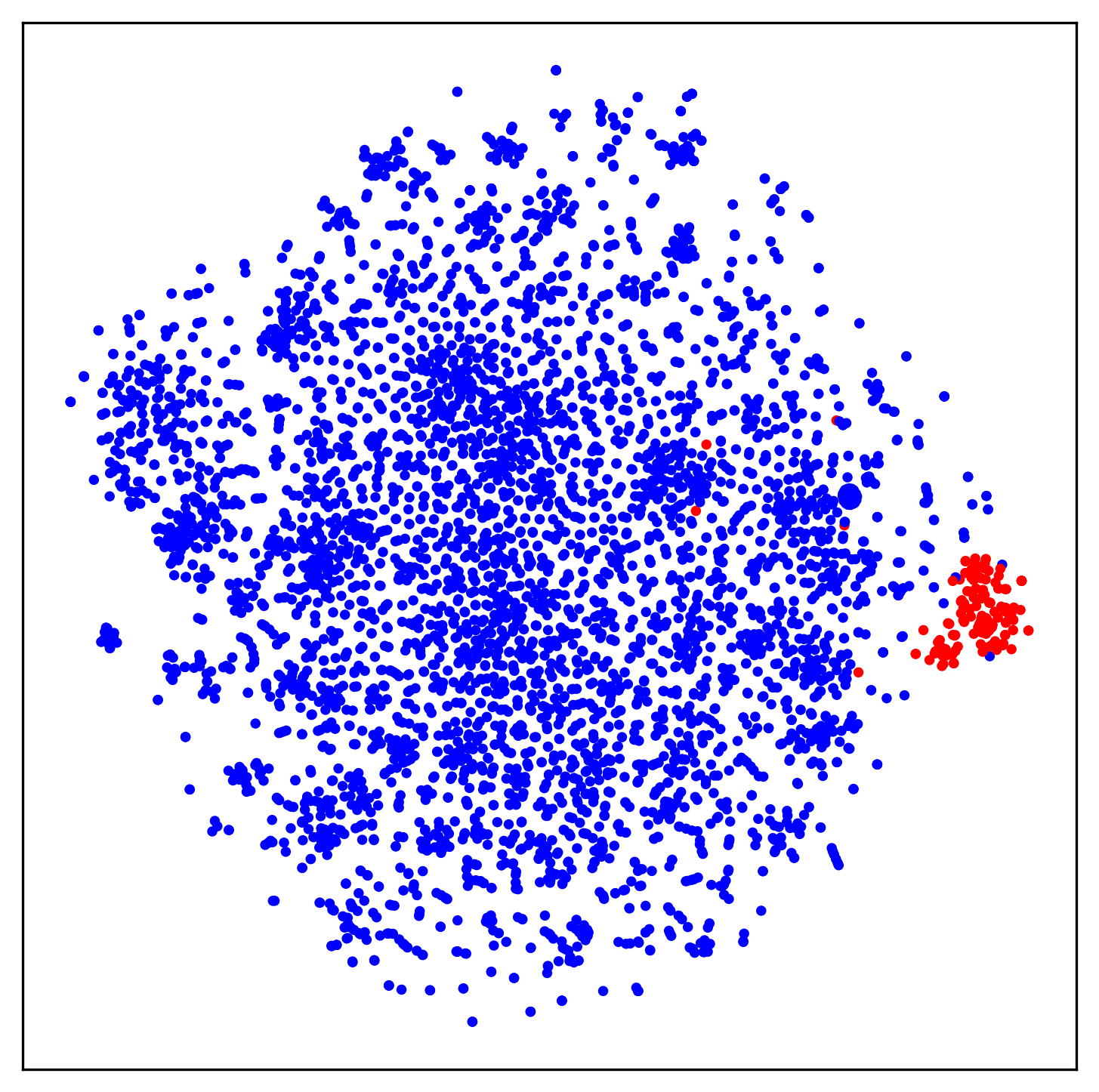} &
		\includegraphics[width=0.15\textwidth]{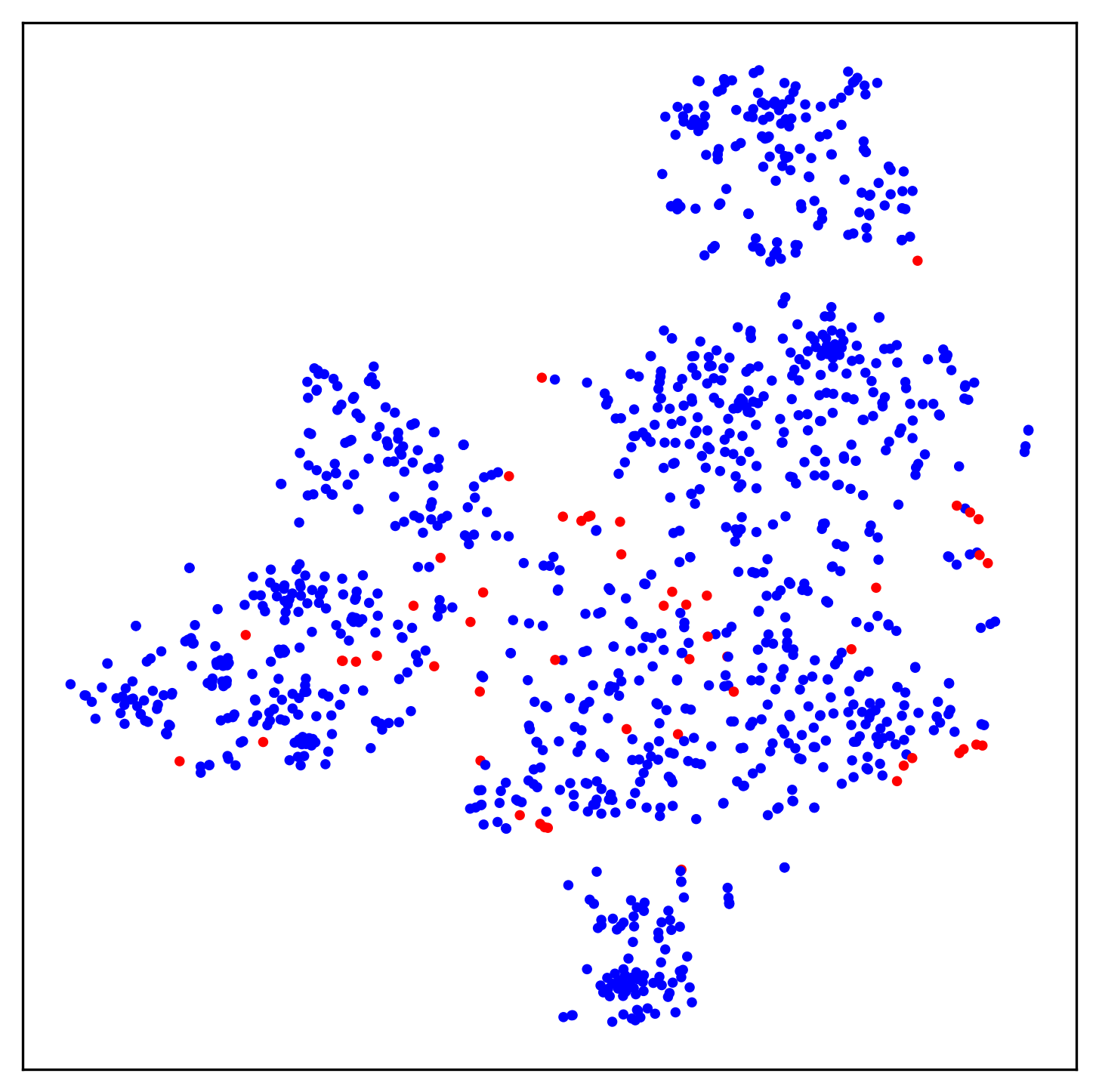} &
        \includegraphics[width=0.15\textwidth]{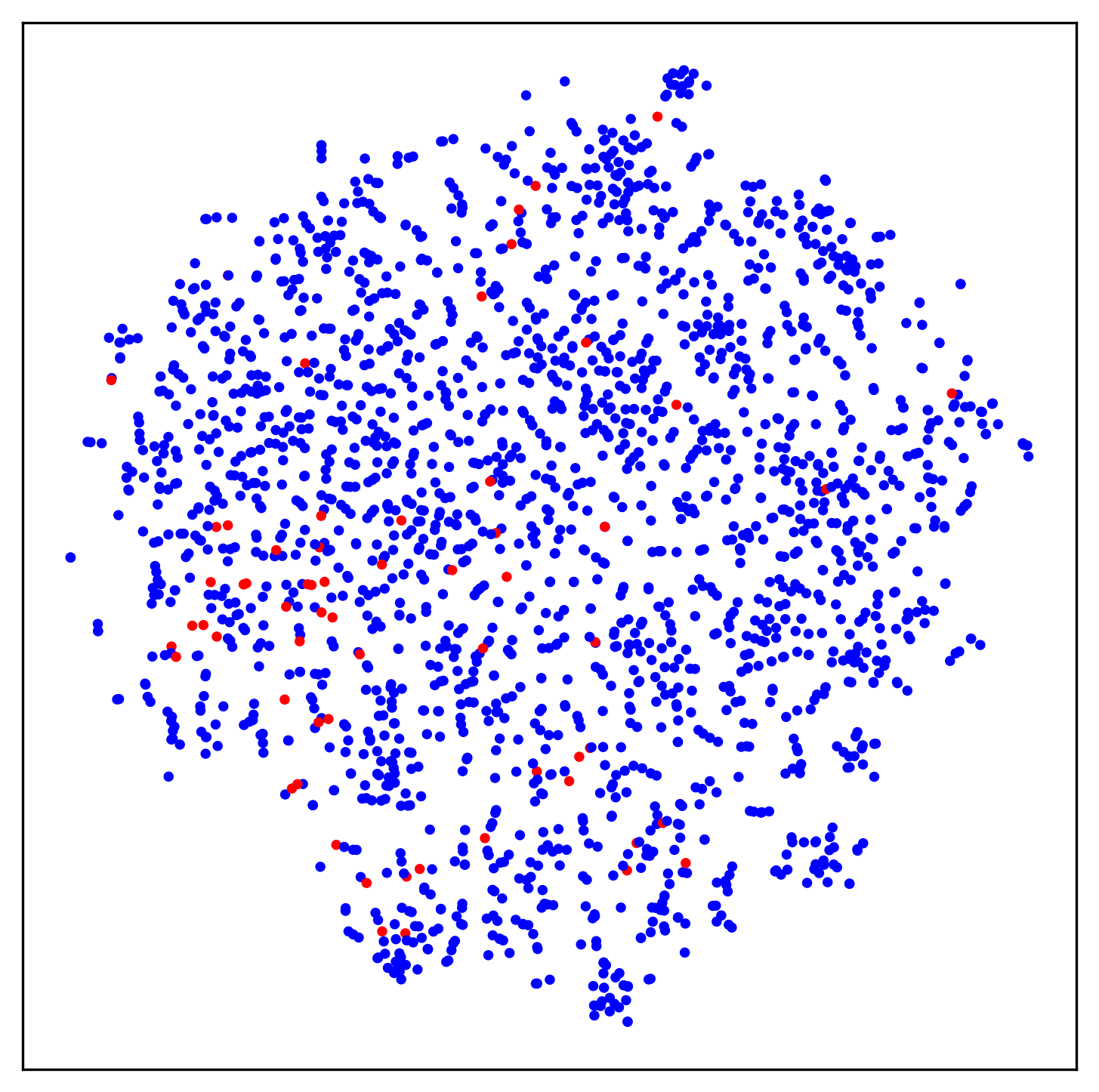} &
        \includegraphics[width=0.15\textwidth]{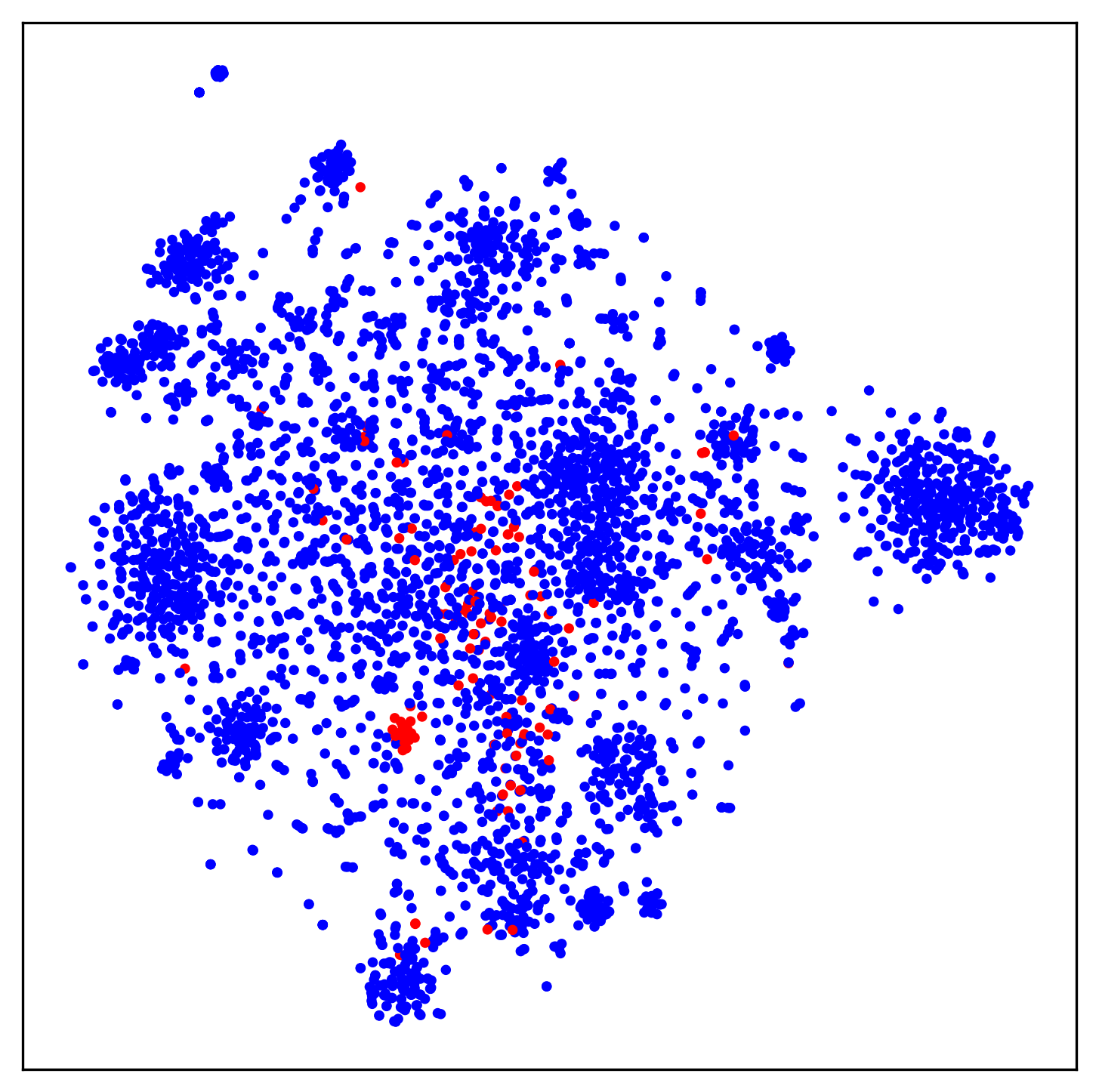} &
        \includegraphics[width=0.15\textwidth]{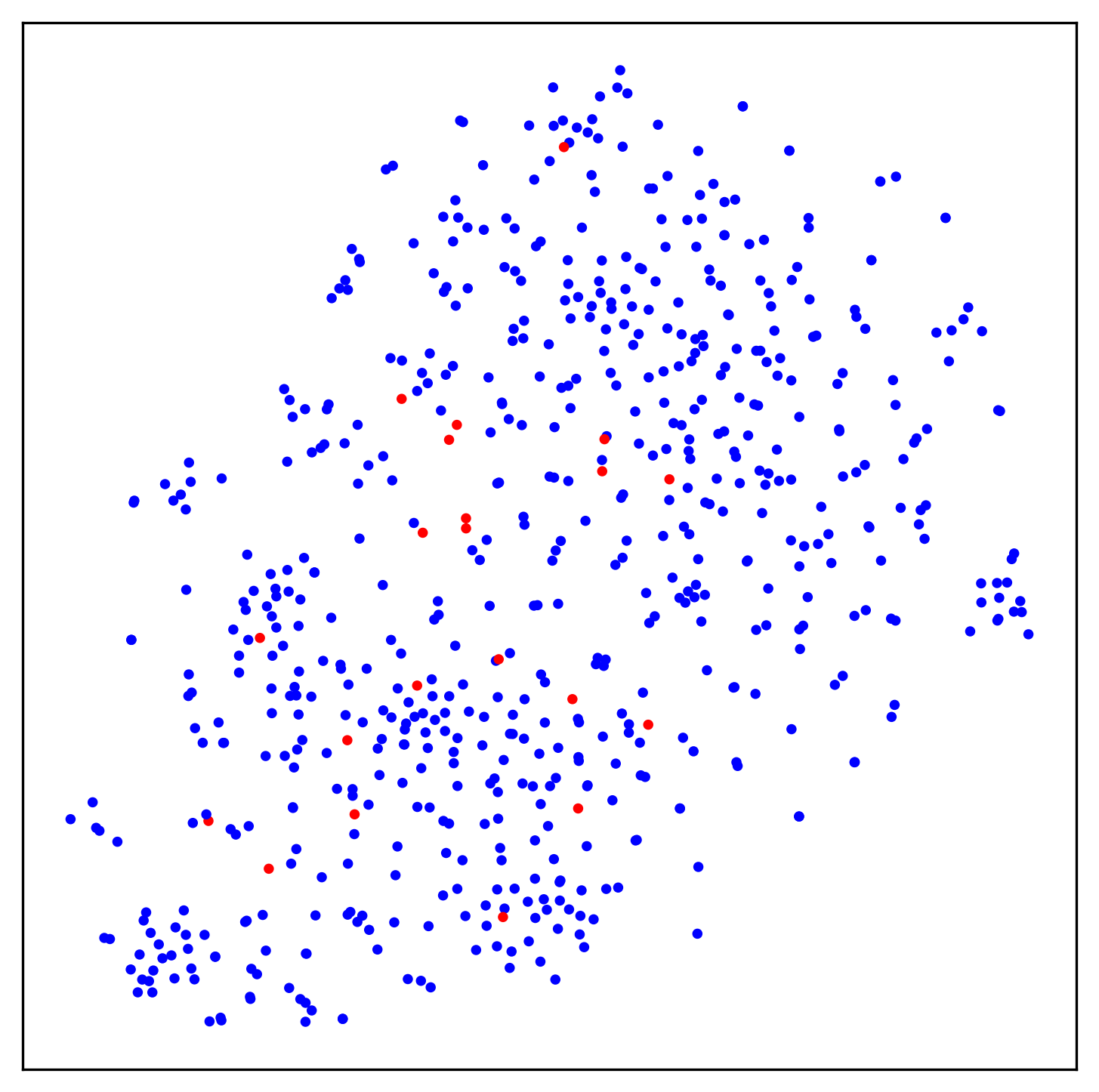}\\
        \midrule
		\textbf{Qwen}       & \includegraphics[width=0.15\textwidth]{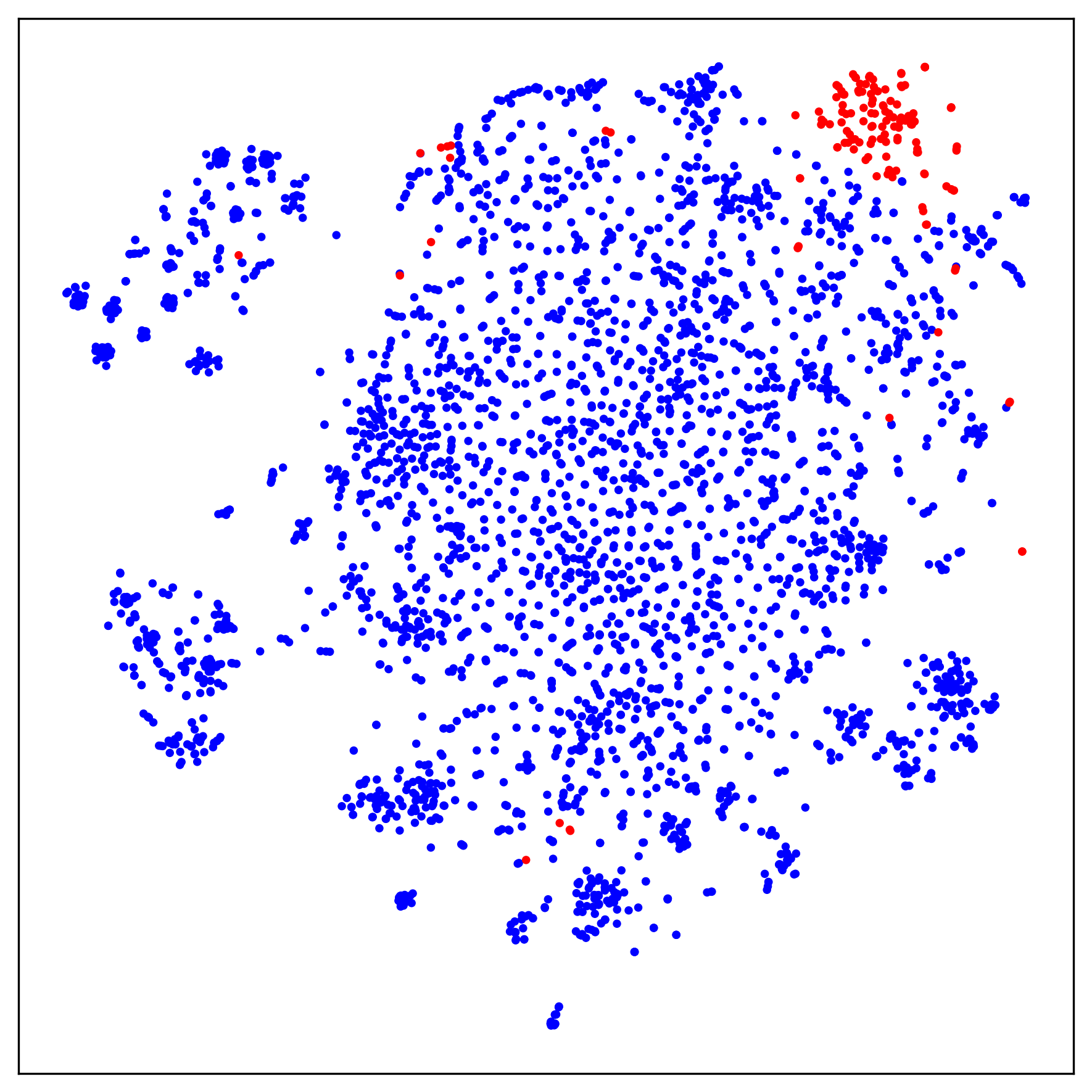} &
	\includegraphics[width=0.15\textwidth]{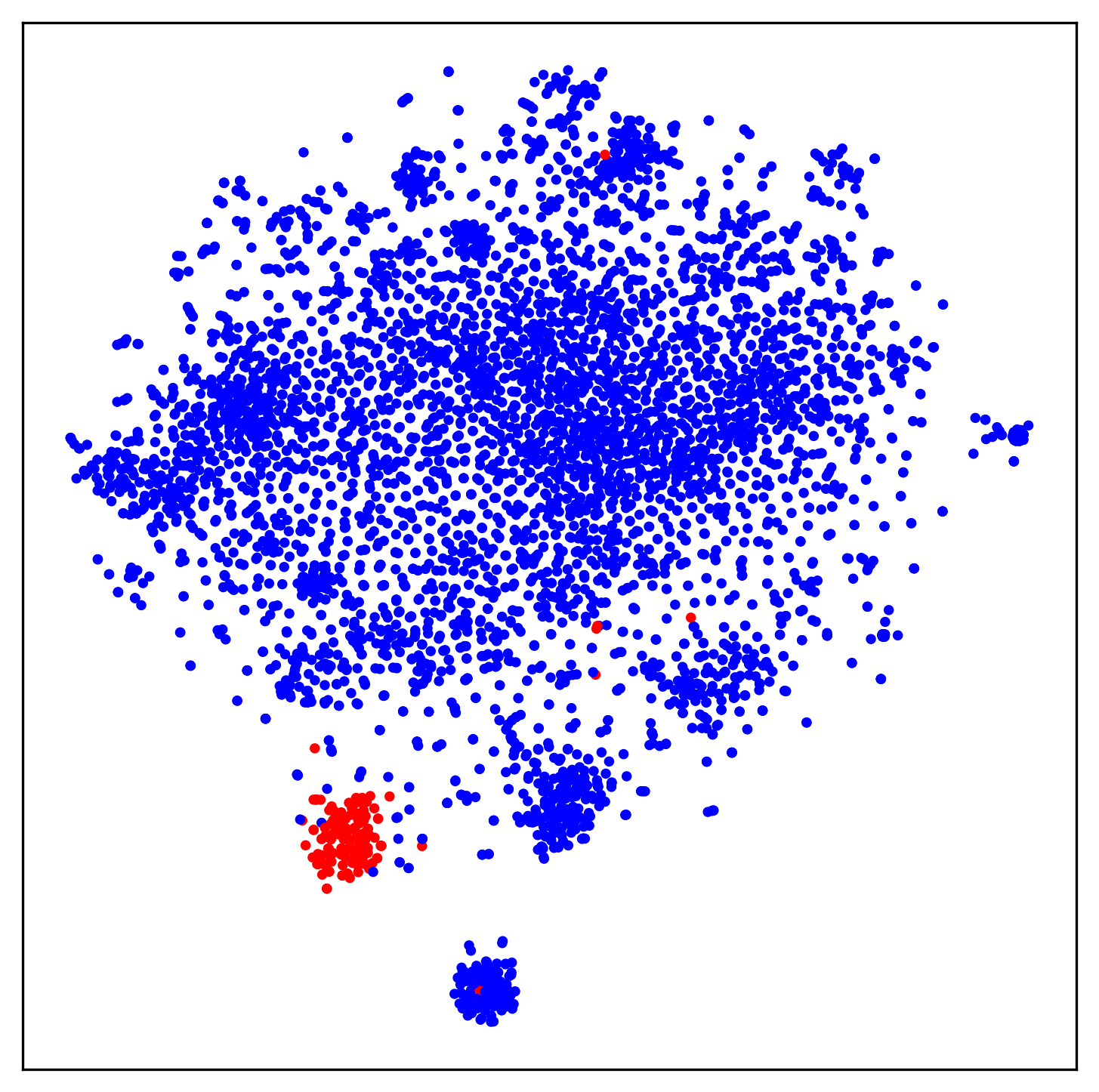} &
		\includegraphics[width=0.15\textwidth]{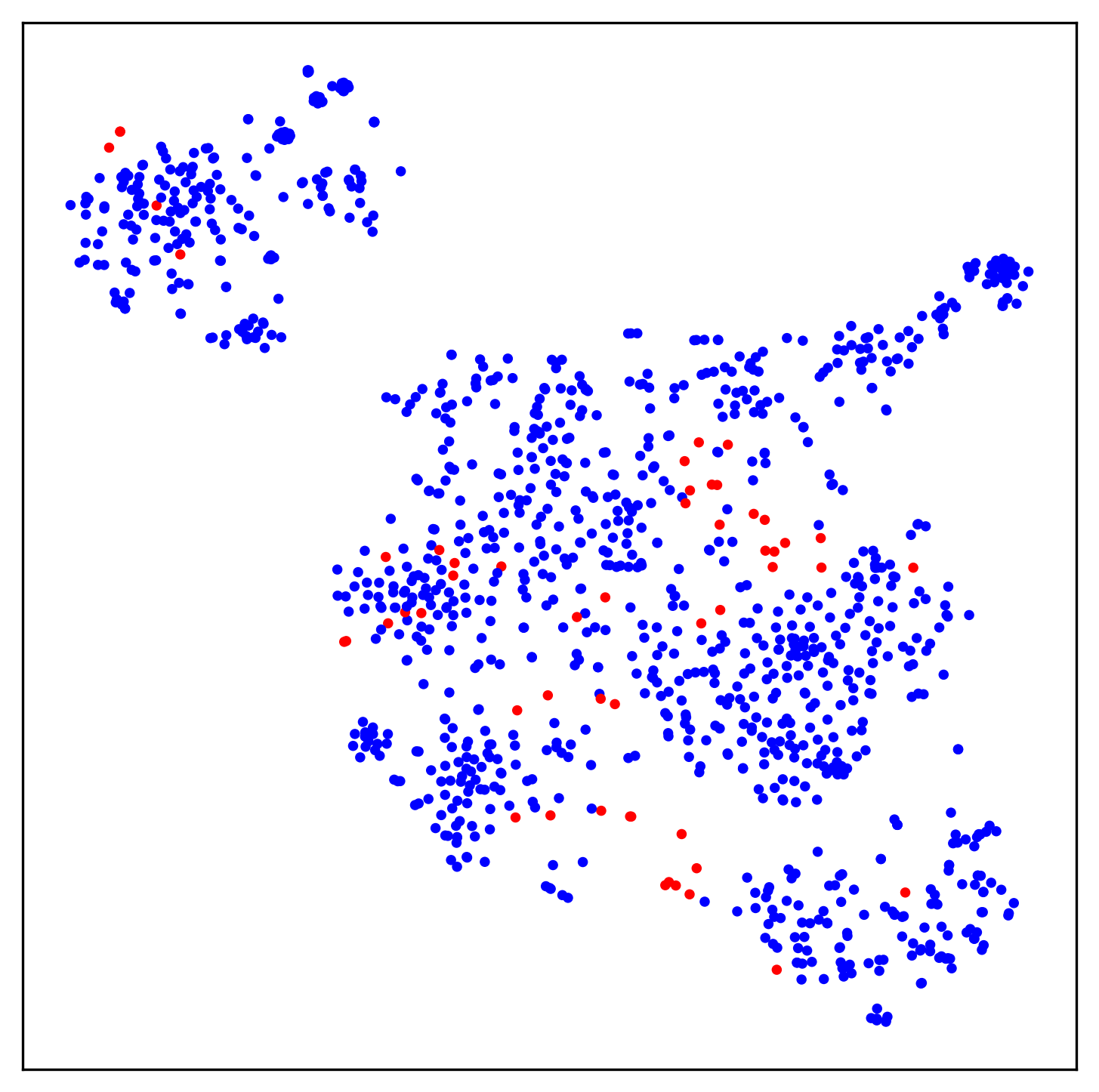} &
        \includegraphics[width=0.15\textwidth]{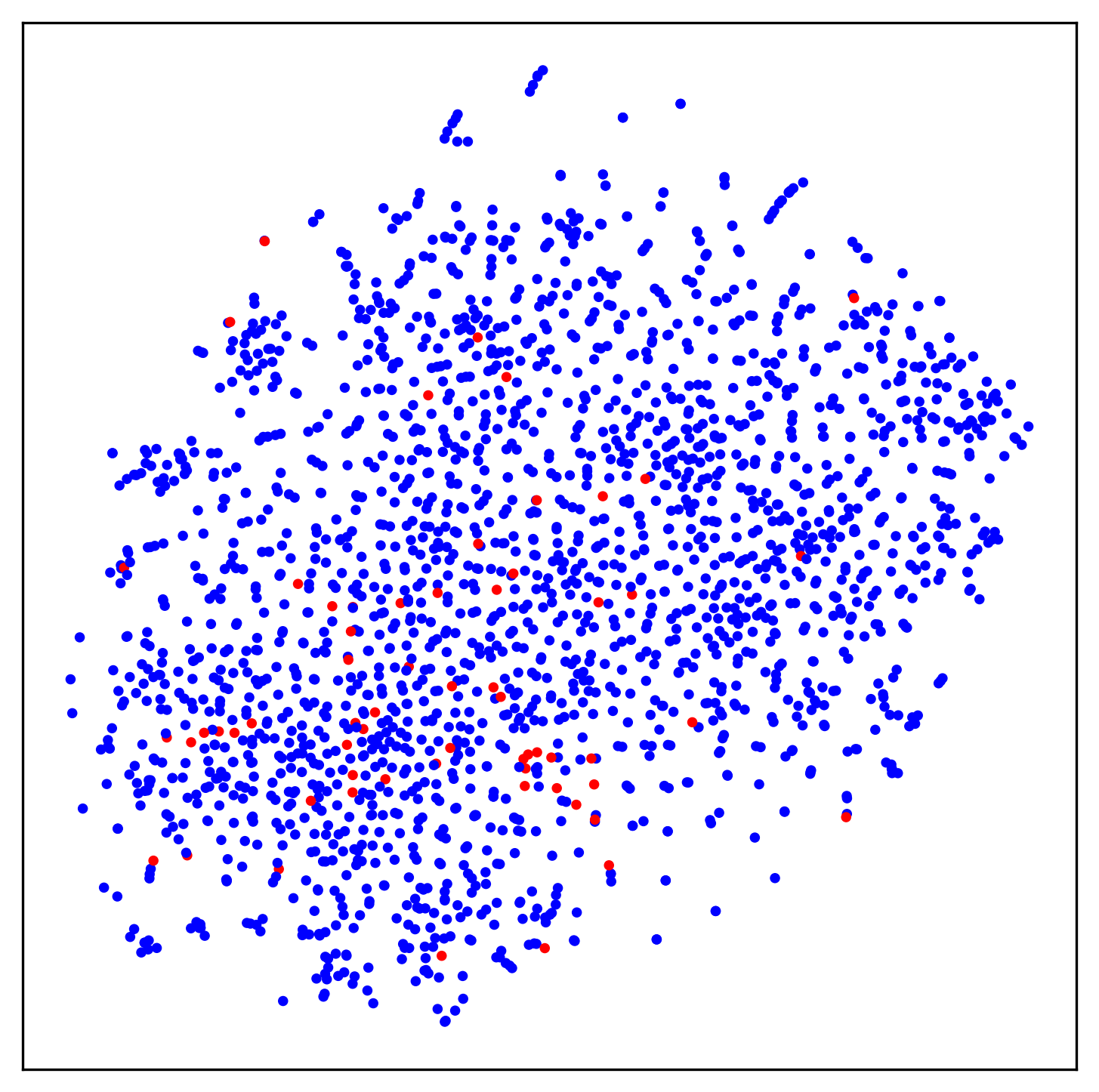} &
        \includegraphics[width=0.15\textwidth]{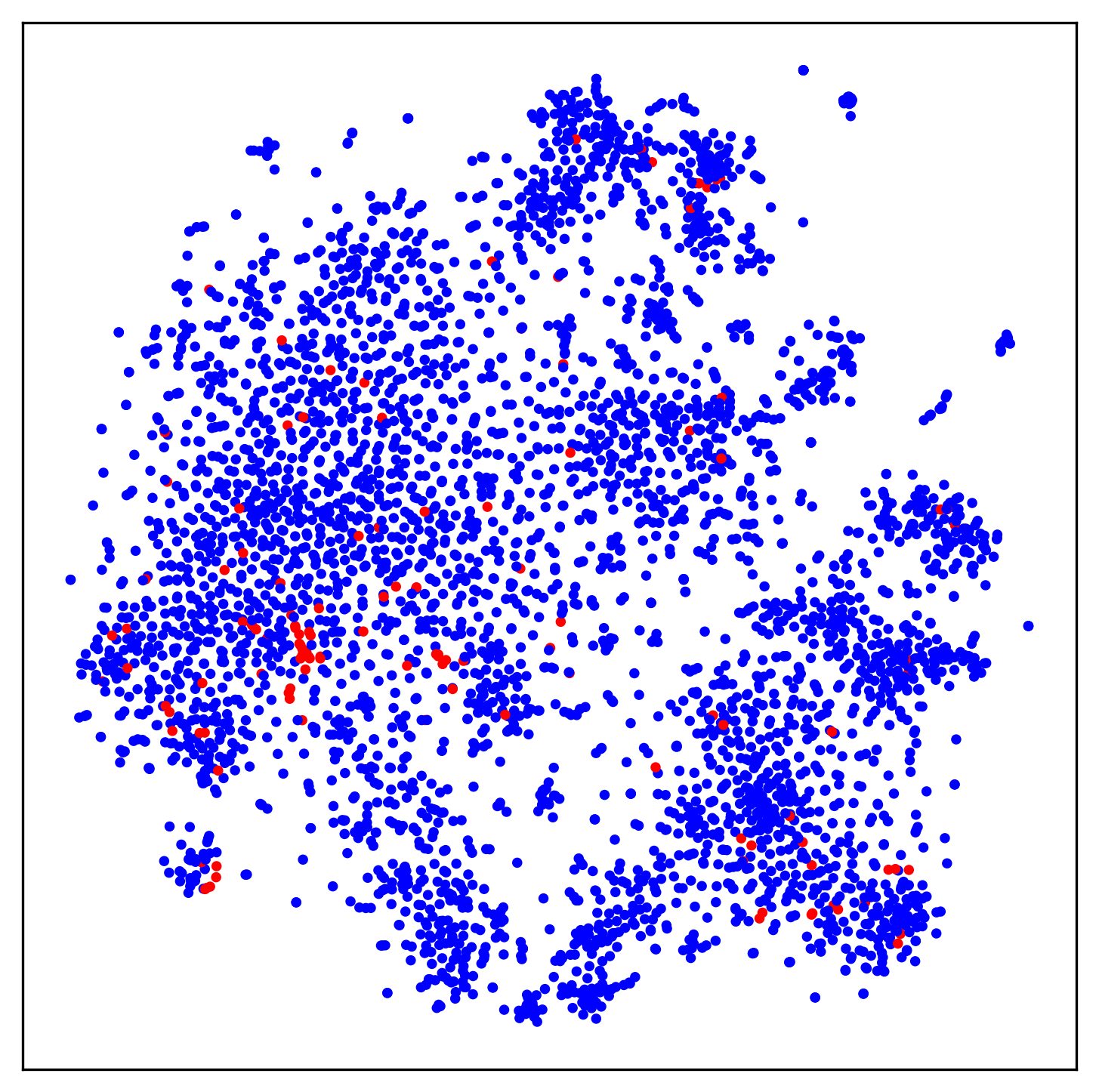} &
        \includegraphics[width=0.15\textwidth]{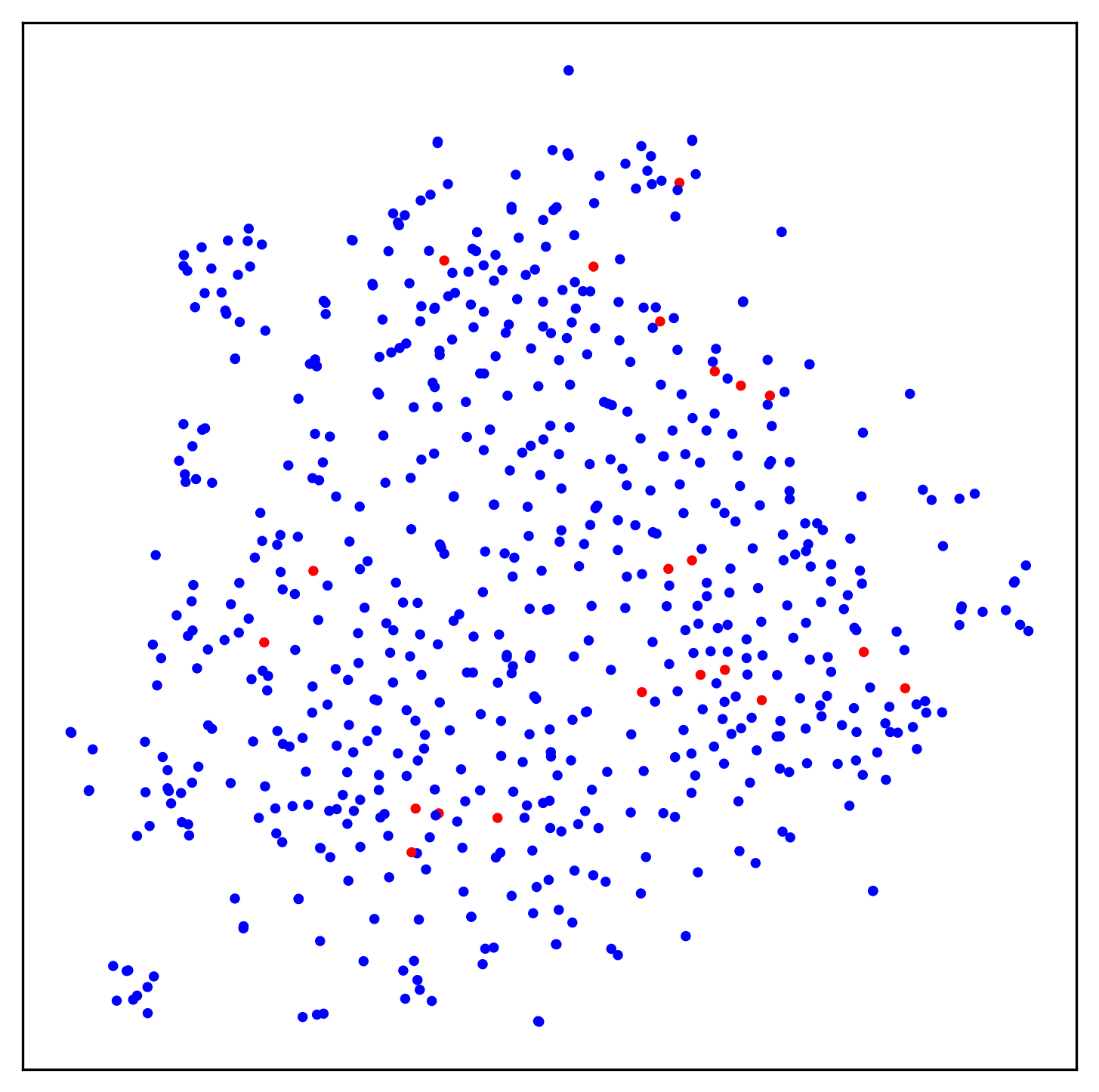}\\
		\bottomrule
	\end{tabular}}
	\label{tsne}
\end{table*}

\section{Embedding Analysis}

To better understand how different embedding models encode normal and anomalous instances, we visualize their embedding spaces using t-SNE projections across 6 datasets. 
Figure~\ref{tsne} presents the t-SNE plots for embeddings extracted from 8 embedding models, blue points represent normal instances, while red points denote anomalies.

\textbf{Separation of Normal and Anomalous Instances.} As defined in Section~\ref{pd}, anomalies should ideally exhibit significant deviation from normal instances in the embedding space. The extent to which embeddings separate anomalies from normal data is a crucial factor in determining their effectiveness for anomaly detection.

Most embedding models exhibits clear separation, particularly in the Email Spam dataset, where anomalous points form distinct regions away from the normal distribution. 
% Llama, stella and Qwen  effectively push anomalies to the periphery in datasets like SMS Spam, suggesting that they capture meaningful textual deviations.
BERT struggles with clear separation, with many anomalies still embedded within normal clusters. This indicates that these models may not encode sufficient discriminative features for anomaly detection tasks.

\textbf{Dataset-Specific Challenges.} The effectiveness of embedding-based anomaly detection varies significantly across datasets, highlighting the influence of domain characteristics:
\begin{itemize}
    \item Spam Detection (Email Spam, SMS Spam): most embedding models perform well, reflecting their ability to capture explicit spam patterns (e.g., domain-specific keywords, unusual syntax). In contrast, BERT shows more overlap between spam and normal messages, leading to weaker anomaly separation.
    \item Fake News Detection (COVID-Fake, LIAR2): The separation of anomalies is less pronounced across most embeddings, likely due to the subtle and nuanced nature of misinformation. This suggests that effective detection may require external knowledge or factual reasoning beyond what standard embeddings can provide.
    \item Offensive Language (Hate Speech, OLID): All embeddings perform poorly, with anomalies scattered among normal instances. This suggests that hate speech and offensive language often depend on implicit contextual cues rather than explicit linguistic differences, making them harder to distinguish using standard embeddings.
\end{itemize}

\textbf{Clustered Anomalies in Spam Detection.} For both Email Spam and SMS Spam datasets, the anomalies tend to form compact clusters rather than being scattered as isolated points. This behavior contrasts with other datasets, where anomalies are often more dispersed.

Unlike anomalies in misinformation or hate speech detection, which can manifest in subtle linguistic variations, spam messages tend to exhibit repetitive patterns, including URLs, phone numbers, irregular word spacing and excessive punctuation.
Since these patterns are highly distinct but internally consistent, embeddings may cluster them into a well-defined anomaly group rather than spreading them across the feature space.

\section{Experiments Environment}

The entire pipeline, including embedding extraction and anomaly detection, was implemented in Python 3.9. Experiments were executed on a computational setup equipped with a Ryzen 9 5900X 12-core CPU for data preprocessing and model orchestration, and an Nvidia RTX 3060 GPU with 12GB of memory for model inference and embedding generation.

\end{document}